\documentclass[10pt,journal,compsoc]{IEEEtran}

\usepackage{xr}
\usepackage{amsmath}
\usepackage{amssymb}
\usepackage{pifont}

\usepackage{subfig}
\usepackage{comment}
\usepackage{soul}
\usepackage{algorithmic}
\usepackage{hyperref}
\usepackage{booktabs}
\usepackage{threeparttable}
\usepackage{mathrsfs}
\usepackage{amsfonts}
\usepackage[ruled]{algorithm2e}

\usepackage{multirow}
\usepackage{graphicx}
\usepackage{color}
\usepackage{balance}
\usepackage{caption}
\usepackage{adjustbox}
\usepackage{soul}
\setstcolor{red}
\usepackage{array}
\usepackage{blindtext}
\usepackage{utfsym}
\usepackage{colortbl} 
\usepackage{tikz}

\newcommand*\emptycirc[1][1ex]{\tikz\draw[thick] (0,0) circle (#1);} 
\newcommand*\halfcirc[1][1ex]{%
  \begin{tikzpicture}
  \draw[fill] (0,0)-- (90:#1) arc (90:270:#1) -- cycle ;
  \draw[thick] (0,0) circle (#1);
  \end{tikzpicture}}
\newcommand*\fullcirc[1][1ex]{\tikz\fill (0,0) circle (#1);}
\newcommand{\ec}{\emptycirc[0.8ex]}
\newcommand{\hc}{\halfcirc[0.8ex]}
\newcommand{\fc}{\fullcirc[0.9ex]}

\usepackage[edges]{forest}
\definecolor{hidden-draw}{RGB}{0,0,0}
\definecolor{hidden-pink}{rgb}{0.98, 0.94, 0.75}
\definecolor{level0}{rgb}{0.67, 0.88, 0.69}
\definecolor{level1}{rgb}{0.98, 0.92, 0.84}
\definecolor{level2}{rgb}{0.8, 0.8, 1.0}
\definecolor{level3}{rgb}{1.0, 0.71, 0.76}
\definecolor{level4}{rgb}{0.49, 0.99, 0.0}

\newcommand{\definitionautorefname}{Definition}

\newcommand*{\shortautoref}[1]{%
  \begingroup
    \def\sectionautorefname{Sec.}%
    \def\subsectionautorefname{Sec.}%
    \def\figureautorefname{Fig.}%
    \def\tableautorefname{Tab.}%
    \def\equationautorefname{Eq.}%
    \def\subfigureautorefname{Fig.}%
    \def\definitionautorefname{Def.}%
    \autoref{#1}%
  \endgroup
}

\definecolor{lawngreen}{rgb}{0.49, 0.99, 0.0}
\definecolor{limegreen}{rgb}{0.2, 0.8, 0.2}
\definecolor{pink}{rgb}{1, 0, 0.5}
\definecolor{airforceblue}{rgb}{0.36, 0.54, 0.66}

\newcommand{\revision}{\textcolor{black}}

\hypersetup{
    linkbordercolor=airforceblue,
    urlbordercolor=pink,
    citebordercolor=green
}

\newtheorem{definition}{Definition}
\usepackage{makecell}
\ifCLASSOPTIONcompsoc
  \usepackage[nocompress]{cite}
\else
  \usepackage{cite}
\fi

\ifCLASSINFOpdf

\else

\fi

\begin{document}
\title{A Survey on Graph Neural Networks for Time Series: Forecasting, Classification, Imputation, and Anomaly Detection}

\author{
Ming Jin, 
Huan Yee Koh, 
Qingsong Wen,
Daniele Zambon, 
Cesare Alippi,~\IEEEmembership{Fellow,~IEEE},\\
Geoffrey I. Webb,~\IEEEmembership{Fellow,~IEEE}, 
Irwin King,~\IEEEmembership{Fellow,~IEEE},
Shirui Pan,~\IEEEmembership{Senior Member,~IEEE}

\IEEEcompsocitemizethanks{
\IEEEcompsocthanksitem Ming Jin and Shirui Pan are with the School of Information and Communication Technology, Griffith University, Queensland, Australia. 
E-mail: mingjinedu@gmail.com, s.pan@griffith.edu.au;
\IEEEcompsocthanksitem Huan Yee Koh and Geoff Webb are with the Department of Data Science and AI, Monash University, Melbourne, Australia. E-mail: \{huan.koh, geoff.webb\}@monash.edu;
\IEEEcompsocthanksitem Qingsong Wen is with Squirrel AI, Seattle, WA, USA. E-mail: qingsongedu@gmail.com;
\IEEEcompsocthanksitem Daniele Zambon and Cesare Alippi are with Swiss AI Lab IDSIA, Università della Svizzera italiana, Lugano, Switzerland. Cesare is also with Politecnico di Milano, Milano, Italy. E-mail: \{daniele.zambon, cesare.alippi\}@usi.ch;
\IEEEcompsocthanksitem Irwin King is with the Department of Computer Science \& Engineering, The Chinese University of Hong Kong. E-mail: king@cse.cuhk.edu.hk.
\IEEEcompsocthanksitem Corresponding author: Shirui Pan.}
\thanks{M. Jin and H. Y. Koh contributed equally to this work.}
\thanks{Version date:  \today}
}

\markboth{Journal of \LaTeX\ Class Files,~Vol.~14, No.~8, August~2021}%
{Shell \MakeLowercase{\textit{et al.}}: Bare Demo of IEEEtran.cls for Computer Society Journals}

\IEEEtitleabstractindextext{
\begin{abstract}
Time series are the primary data type used to record dynamic system measurements and generated in great volume by both physical sensors and online processes (virtual sensors). Time series analytics is therefore crucial to unlocking the wealth of information implicit in available data. With the recent advancements in graph neural networks (GNNs), there has been a surge in GNN-based approaches for time series analysis. These approaches can explicitly model inter-temporal and inter-variable relationships, which traditional and other deep neural network-based methods struggle to do. In this survey, we provide a comprehensive review of graph neural networks for time series analysis (GNN4TS), encompassing four fundamental dimensions: forecasting, classification, anomaly detection, and imputation. Our aim is to guide designers and practitioners to understand, build applications, and advance research of GNN4TS. At first, we provide a comprehensive task-oriented taxonomy of GNN4TS. Then, we present and discuss representative research works and introduce mainstream applications of GNN4TS. A comprehensive discussion of potential future research directions completes the survey. This survey, for the first time, brings together a vast array of knowledge on GNN-based time series research, highlighting foundations, practical applications, and opportunities of graph neural networks for time series analysis.
\end{abstract}

\centering
\vspace{2mm}
\small\textbf{Project Page:} \textcolor{pink}{https://github.com/KimMeen/Awesome-GNN4TS}
\vspace{2mm}

}

\maketitle

\IEEEdisplaynontitleabstractindextext

\IEEEpeerreviewmaketitle

\section{Introduction}\label{sec:introduction}

\IEEEPARstart{T}he advent of advanced sensing and data stream processing technologies has led to an explosion of time series data~\cite{wen2022robust,esling2012time,zhang2023selfsupervised}. 
The analysis of time series not only provides insights into past trends but also facilitates a multitude of tasks such as forecasting~\cite{lim2021time}, classification~\cite{ismail2019deep}, anomaly detection~\cite{blazquez2021review}, and data imputation~\cite{fang2020time}. This lays the groundwork for time series modeling paradigms that leverage on historical data to understand current and future possibilities. 
Time series analytics have become increasingly crucial in various fields, including but not limited to cloud computing, transportation, energy, finance, social networks, and the Internet-of-Things~\cite{zhang2011data,zhou2023robust,cook2019anomaly}.

\begin{figure}[t]
    \centering
    \includegraphics[width=0.9\linewidth]{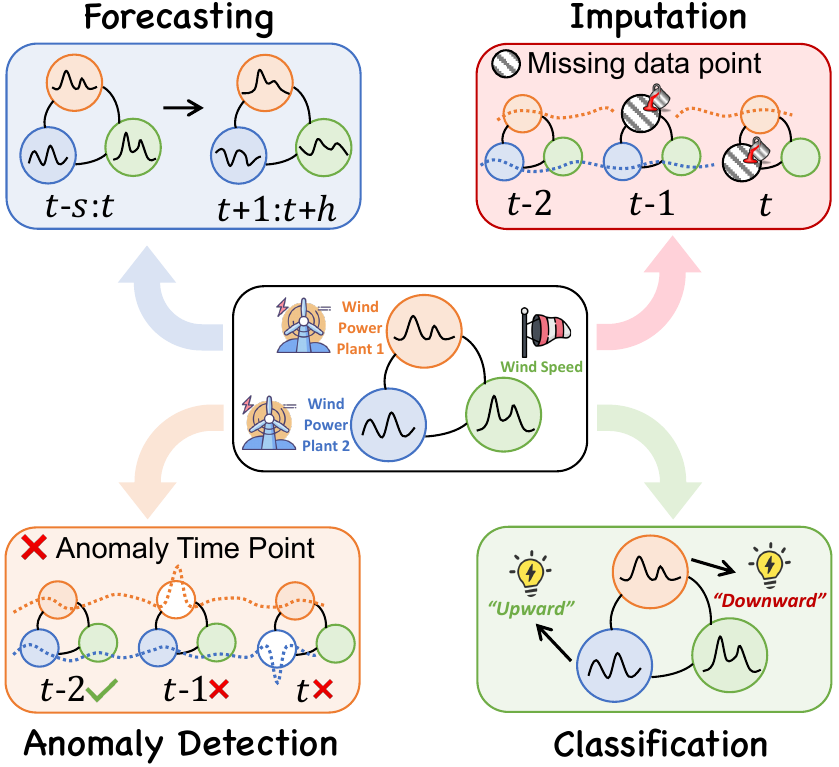}
    \caption{Graph neural networks for time series analysis (GNN4TS). In this example of a wind farm, different analytical tasks can be categorized into time series forecasting, classification, anomaly detection, and imputation.}
    \label{fig:GNN4TS-Intro}
    \vspace{-3mm}
\end{figure}

\begin{table*}[t]
\centering  
\renewcommand\arraystretch{1.3}
\caption{Comparison between our survey and other related surveys.}
\begin{threeparttable}
\begin{tabular}{p{0.18\textwidth} >{\centering\arraybackslash}p{0.06\textwidth} >{\centering\arraybackslash}p{0.06\textwidth} >{\centering\arraybackslash}p{0.06\textwidth} >{\centering\arraybackslash}p{0.09\textwidth} >{\centering\arraybackslash}p{0.09\textwidth} >{\centering\arraybackslash}p{0.09\textwidth} >{\centering\arraybackslash}p{0.09\textwidth}}
\toprule
\multicolumn{1}{c}{\multirow{2}{*}{\textbf{Survey}}} & \multicolumn{1}{c}{\multirow{2}{*}{\textbf{\revision{Year}}}} & \multicolumn{2}{c}{\textbf{Domain}} & \multicolumn{4}{c}{\textbf{Scope}}                                     \\ \cline{3-4} \cline{5-8} 
\multicolumn{1}{c}{} & \multicolumn{1}{c}{}                       & \textbf{Specific}        & \textbf{General}        & \textbf{Forecasting} & \textbf{Classification} & \textbf{Anomaly\ \ Detection} & \textbf{Imputation} \\ \hline
Wang et al.~\cite{wang2020deep} & \revision{2019}             &                                                  &  \textcolor{limegreen}{\usym{2714}}           & \fc             & \fc            & \fc               & \ec           \\
Ye et al.~\cite{ye2020build} & \revision{2020}                & \textcolor{limegreen}{\usym{2714}}               &                                               & \fc             & \hc            & \ec               & \ec           \\
Jiang and Luo~\cite{jiang2022graph} & \revision{2021}         & \textcolor{limegreen}{\usym{2714}}               &                                               & \fc             & \ec            & \ec               & \ec            \\
Bui et al.~\cite{bui2022spatial} & \revision{2021}            & \textcolor{limegreen}{\usym{2714}}               &                                               & \fc             & \ec            & \ec               & \ec            \\
Jin et al.~\cite{jin2023spatio} & \revision{2023}             &  \textcolor{limegreen}{\usym{2714}}              &                                               & \fc             & \ec            & \ec               & \ec            \\
Al Sahili and Awad~\cite{sahili2023spatio} & \revision{2023}  &                                                  & \textcolor{limegreen}{\usym{2714}}            & \hc             & \ec            & \ec               & \ec            \\
Rahmani et al.~\cite{rahmani2023graph} & \revision{2023}  &                                                  & \textcolor{limegreen}{\usym{2714}}            & \fc             & \ec            & \ec               & \hc            \\ \hline
\textbf{Our Survey} & \revision{2024}                         &                                                  &  \textcolor{limegreen}{\usym{2714}}           & \fc             & \fc            & \fc               & \fc            \\ \bottomrule
\end{tabular}
\begin{tablenotes}
    \footnotesize \item[*] Specifically, \ec \ represents ``Not Covered'', \hc \ signifies ``Partially Covered'', and \fc \ corresponds to ``Fully Covered''.
\end{tablenotes}
\end{threeparttable}
\label{table:compare_survey}
\end{table*}

Many time series involve complex interactions across time (such as lags in propagation of effects) and variables (such as the relationship among the variables representing neighboring traffic sensors). By treating time points or variables as nodes and their relationships as edges, a model structured in the manner of a network or graph can effectively solve the task at hand by exploiting both data and relational information. Indeed, much time series data is spatial-temporal in nature, with different variables in the series capturing information about different locations -- space -- too, meaning it encapsulates not only time information but also spatial relationships~\cite{wang2020deep}. This is particularly evident in scenarios such as urban traffic networks, population migration, and global weather forecasting. In these instances, a localized change, such as a traffic accident at an intersection, an epidemic outbreak in a suburb, or extreme weather in a specific area, can propagate and influence neighboring regions. This spatial-temporal characteristic is a common feature of many dynamic systems, including the wind farm in \shortautoref{fig:GNN4TS-Intro}, where the underlying time series displays a range of correlations and heterogeneities~\cite{jin2023spatio, cini2023taming}. Traditional analytic tools, such as support vector regression (SVR)~\cite{cao2003support}, gradient boosting decision tree (GBDT)~\cite{xia2017traffic}, vector autoregressive (VAR)~\cite{biller2003modeling}, and autoregressive integrated moving average (ARIMA)~\cite{box1970distribution}, struggle to handle complex time series relations (e.g., nonlinearities and inter-variable relationships), resulting in less accurate prediction results~\cite{jin2022multivariate}. The advent of deep learning technologies has led to the development of different neural networks based on convolutional neural networks (CNN)~\cite{zhao2017convolutional, borovykh2017conditional}, recurrent neural networks (RNN)~\cite{connor1994recurrent}, and transformers~\cite{wen2022transformers}, which have shown significant advantages in modeling real-world time series data. However, one of the biggest limitations of the above methods is that they do not explicitly model the spatial relations existing between time series in non-Euclidean space~\cite{jin2023spatio}, which limits their expressiveness~\cite{jin2023powerful}.

In recent years, graph neural networks (GNNs) have emerged as a powerful tool for learning non-Euclidean data representations~\cite{wu2020comprehensive,yang2021discrete,yang2022hyperbolic,koh2023psichic}, paving the way for modeling real-world time series data. This enables the capture of diverse and intricate relationships, both inter-variable (connections between different variables within a multivariate series) and inter-temporal (dependencies between different points in time). Considering the complex spatial-temporal dependencies inherent in real-world scenarios, a line of studies has integrated GNNs with various temporal modeling frameworks to capture both spatial and temporal dynamics and demonstrate promising results~\cite{jin2023spatio, sahili2023spatio, jiang2022graph, bui2022spatial, ye2020build}. While early research efforts were primarily concentrated on various forecasting scenarios~\cite{jin2023spatio, jiang2022graph, bui2022spatial}, recent advancements in time series analysis utilizing GNNs have demonstrated promising outcomes in other mainstream tasks. These include classification~\cite{zhang2022graphguided, wang2023irregularly}, anomaly detection~\cite{zhao2020multivariate, deng2021graph}, and imputation~\cite{cini2022filling,liu2023pristi}. In \shortautoref{fig:GNN4TS-Intro}, we provide an overview of graph neural networks for time series analysis (GNN4TS). \\

\noindent\textbf{Related Surveys.} Despite the growing body of research performing various time series analytic tasks with GNNs, existing surveys tend to focus on specific perspectives within a restricted scope. For instance, the survey by Wang et al.~\cite{wang2020deep} offers a review of deep learning techniques for spatial-temporal data mining, but it does not specifically concentrate on GNN-based methods. The survey by Ye et al.~\cite{ye2020build} zeroes in on graph-based deep learning architectures in the traffic domain, primarily considering forecasting scenarios. A recent survey by Jin et al.\cite{jin2023spatio} offers an overview of GNNs for predictive learning in urban computing, but neither extends its coverage to other application domains nor thoroughly discusses other tasks related to time series analysis. 
Finally, we mention the work by Rahmani et al.~\cite{rahmani2023graph}, which expands the survey of GNNs to many intelligent transportation systems, but tasks other than forecasting remain overlooked. A detailed comparison between our survey and others is presented in \shortautoref{table:compare_survey}. \\

To fill the gap, this survey offers a comprehensive and up-to-date review of graph neural networks for time series analysis, encompassing the majority of tasks ranging from time series forecasting, classification, anomaly detection, and imputation. Specifically, we first provide two broad views to classify and discuss existing works from the task- and methodology-oriented perspectives. Then, we delve into six popular application sectors within the existing research of GNN4TS, and propose several potential future research directions. The key contributions of our survey are summarized as follows:

\begin{itemize}
   \item \textbf{First Comprehensive Survey.} To the best of our knowledge, this is the first comprehensive survey that reviews the recent advances in mainstream time series analysis tasks with graph neural networks. It covers a wide range of recent research and provides a broad view of the development of GNN4TS without restricting to specific tasks or domains.
   \item \textbf{Unified and Structured Taxonomy.} We present a unified framework to structurally categorize existing works from task- and methodology-oriented perspectives. 
   In the first classification, we offer an overview of tasks in time series analysis, covering different problem settings prevalent in GNN-based research; and in the second classification, we dissect GNN4TS in terms of spatial and temporal dependencies modeling and the overall model architecture.
   \item \textbf{Detailed and Current Overview.} We conduct a comprehensive review that not only covers the breadth of the field but also delves into the depth of individual studies with fine-grained classification and detailed discussion, providing readers with an up-to-date understanding of the state-of-the-art in GNN4TS.
    \item \textbf{Broadening Applications.} We discuss the expanding applications of GNN4TS across various sectors, highlighting its versatility and potential for future growth in diverse fields.
    \item \textbf{Future Research Directions.} We shed light on potential future research directions, offering insights and suggestions that could guide and inspire future research in the field of GNN4TS.
\end{itemize}

The remainder of this survey is organized as follows: \shortautoref{sec:definition} provides notations used throughout the paper. \shortautoref{sec:categorization} presents the taxonomy of GNN4TS from different perspectives. \shortautoref{sec:forecasting}, \shortautoref{sec:anomaly detection}, \shortautoref{sec:classification}, and \shortautoref{sec:imputation} review the four major tasks in the GNN4TS literature. \shortautoref{sec:application} surveys popular applications of GNN4TS across various fields, while \shortautoref{sec:prospect} examines open questions and potential future directions.
\section{Definition and Notation}\label{sec:definition}
Time series data comprises a sequence of observations gathered or recorded over a period of time. This data can be either \textit{regularly} or \textit{irregularly sampled}, with the latter also referred to as time series data with missing values. Within each of these cases, the data can be further classified into two primary types: \textit{univariate} and \textit{multivariate time series}. In the sequel, we employ bold uppercase letters (e.g., $\mathbf{X}$), bold lowercase letters (e.g., $\mathbf{x}$), and calligraphic letters (e.g., $\mathcal{V}$) to denote matrices, vectors, and sets, respectively.

\begin{definition}[Univariate Time Series]
A univariate time series is a sequence of scalar observations collected over time, which can be regularly or irregularly sampled. A regularly sampled univariate time series is defined as $\mathbf{X} = \{x_1, x_2, ..., x_T\} \in \mathbb{R}^{T}$, where $x_t \in \mathbb{R}$. For an irregularly sampled univariate time series, observations are collected at non-uniform time intervals, such as $\mathbf{X} = \{(t_1, x_1), (t_2, x_2), ..., (t_T, x_T)\} \in \mathbb{R}^{T}$, where time points are non-uniformly spaced.
\end{definition}

\begin{definition}[Multivariate Time Series]
A multivariate time series is a sequence of $N$-dimensional vector observations collected over time, i.e., $\mathbf{X} \in \mathbb{R}^{N \times T}$. A regularly sampled multivariate time series has vector observations collected at uniform time intervals, i.e., $\mathbf{x}_t \in \mathbb{R}^{N}$. In an irregularly sampled multivariate time series, there are possibly $N$ unaligned time series with respect to time steps, which implies only $0 \leq n \leq N$ observations available at each time step.
\end{definition}

\begin{figure}[t]
	\centering
	\includegraphics[width=0.45\textwidth]{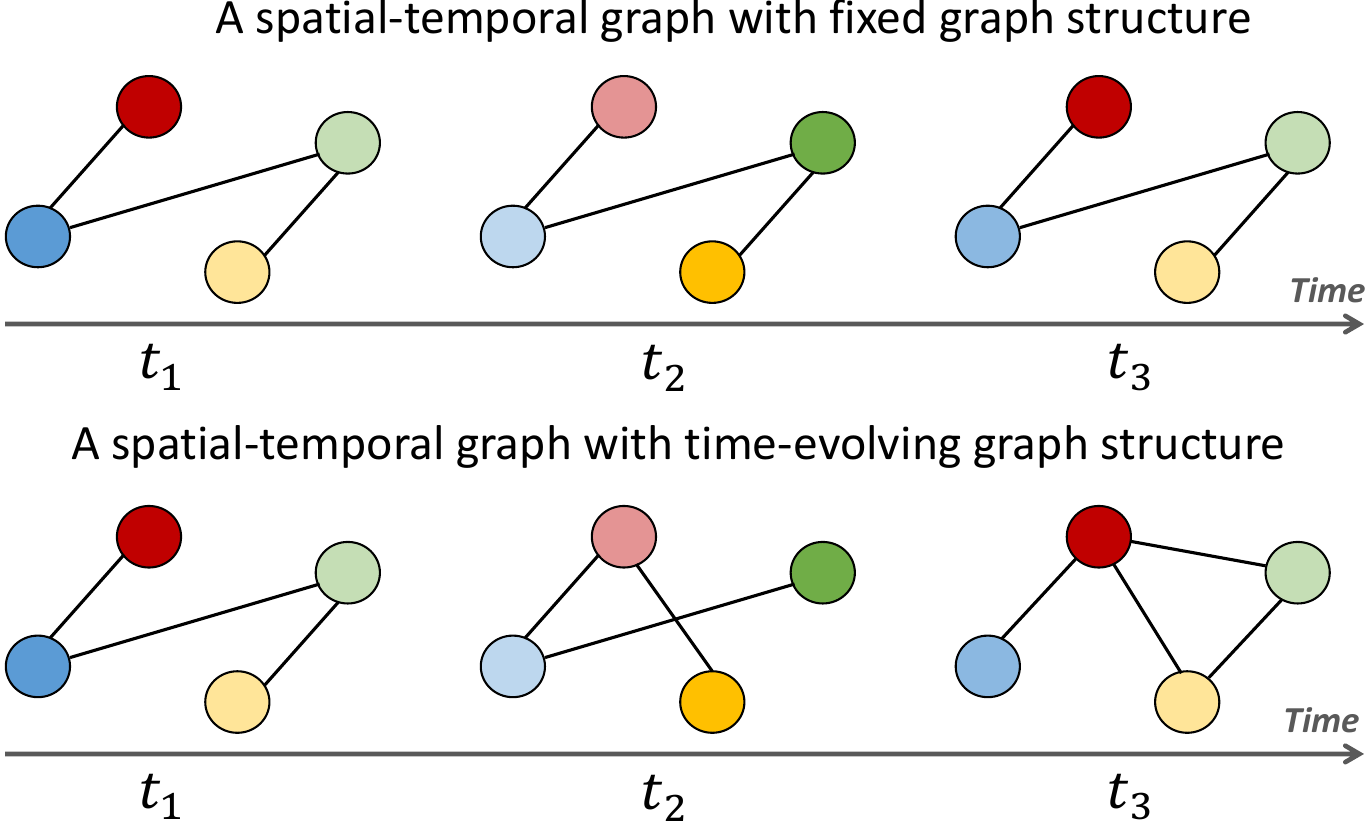}
	\caption{\revision{Examples of spatial-temporal graphs, where node colors represent distinct features. The top and bottom panels demonstrate spatio-temporal graphs with fixed and dynamic graph structures over time, respectively.}}
	\label{fig:example_spatial_temporal_graph}
\end{figure}

The majority of research based on GNNs focuses on modeling multivariate time series, as they can be naturally abstracted into \textit{spatial-temporal graphs}. This abstraction allows for an accurate characterization of dynamic inter-temporal and inter-variable dependencies. The former describes the relations between different time steps within each time series (e.g., the temporal dynamics of red nodes between $t_1$ and $t_3$ in \shortautoref{fig:example_spatial_temporal_graph}), while the latter captures dependencies between time series (e.g., the spatial relations between four nodes at each time step in \shortautoref{fig:example_spatial_temporal_graph}), such as the geographical information of the sensors generating the data for each variable. To illustrate this, we first define \textit{attributed graphs}.

\begin{definition}[Attributed Graph]
An attributed graph is a static graph that associates each node with a set of attributes, representing node features. Formally, an attributed graph is defined as $\mathcal{G} = (\mathbf{A}, \mathbf{X})$, which consists of a (weighted) adjacency matrix $\mathbf{A} \in \mathbb{R}^{N \times N}$ and a node-feature matrix $\mathbf{X} \in \mathbb{R}^{N \times D}$. The adjacency matrix represents the graph topology, which can be characterized by $\mathcal{V} = \{v_1, v_2, \dots, v_N\}$, the set of $N$ nodes, and $\mathcal{E} = \{e_{ij}:=(v_i, v_j) \in \mathcal{V} \times \mathcal{V} \mid \mathbf{A}_{ij} \neq 0\}$, the set of edges; $\mathbf{A}_{ij}$ is the $(i, j)$-th entry in the adjacency matrix $\mathbf{A}$. The feature matrix $\mathbf{X}$ contains the node attributes, where the $i$-th row $\mathbf{x}_i \in \mathbb{R}^{D}$ represents the $D$-dimensional feature vector of node $v_i$.
\end{definition}
In attributed graphs, multi-dimensional edge features can be considered too, however, this paper assumes only scalar weights encoded in the adjacency matrix to avoid overwhelming notations.

In light of this, a spatial-temporal graph can be described as a series of attributed graphs, which effectively represent (multivariate) time series data in conjunction with either evolving or fixed structural information over time.

\begin{definition}[Spatial-temporal Graph] \label{def:spatial-temporal graph}
A spatial-temporal graph can be interpreted as a \textit{discrete-time dynamic graph}~\cite{jin2022neural,yang2022hyperbolic}, i.e., $\mathcal{G} = \{\mathcal{G}_1, \mathcal{G}_2, \cdots, \mathcal{G}_{T}\}$, where $\mathcal{G}_t = (\mathbf{A}_t, \mathbf{X}_t)$ denotes an attributed graph at time $t$. 
$\mathbf{A}_t \in \mathbb{R}^{N \times N}$ and $\mathbf{X}_t \in \mathbb{R}^{N \times D}$ are corresponding adjacency and feature matrices.
$\mathbf{A}_t$ may either evolve over time or remain fixed, depending on specific settings. When abstracting time series data, we let $\mathbf{X}_t := \mathbf{x}_t \in \mathbb{R}^{N}$.
\end{definition}

We introduce graph neural networks as modern deep learning models to process graph-structured data. The core operation in typical GNNs, often referred to as \textit{graph convolution}, involves exchanging information across neighboring nodes. In the context of time series analysis, this operation enables us to explicitly rely on the inter-variable dependencies represented by the graph edges. Aware of the different nuances, we define GNNs in the spatial domain, which involves transforming the input signal with learnable functions along the dimension of $N$.

\begin{definition}[Graph Neural Network]\label{def: graph neural network}
Given an attributed graph $\mathcal{G} = (\mathbf{A}, \mathbf{X})$, we define $\mathbf{x}_i = \mathbf{X}[i,:] \in \mathbb{R}^{D}$ as the $D$-dimensional feature vector of node $v_i$. A GNN learns node representations through two primary functions~\cite{liu2022graph}: $\textsc{Aggregate}(\cdot)$ and $\textsc{Combine}(\cdot)$. The $\textsc{Aggregate}(\cdot)$ function computes and aggregates messages from neighboring nodes, while the $\textsc{Combine}(\cdot)$ function merges the aggregated and previous states to transform node embeddings. Formally, the $k$-th layer in a GNN is defined by the extended
\begin{equation}
    \begin{aligned}
        \mathbf{a_{i}^{(k)}}&=\textsc{Aggregate}^{(k)}\left(\left\{\mathbf{h_{j}^{(k-1)}}: v_j \in \mathcal{N}(v_i)\right\}\right), \\
        \mathbf{h_{i}^{(k)}}&=\textsc{Combine}^{(k)}\left(\mathbf{h_{i}^{(k-1)}}, \mathbf{a_{i}^{(k)}}\right),
    \end{aligned}
    \label{eq: graph convolution with message passing}
\end{equation}
or, more generally, aggregating messages 
computed from both sending and receiving nodes $v_j$ and $v_i$, respectively.
Here, $\mathbf{a_{i}^{(k)}}$ and $\mathbf{h_{i}^{(k)}}$ represent the aggregated message from neighbors and the transformed node embedding of node $v_i$ in the $k$-th layer, respectively. The input and output of a GNN are $\mathbf{h_{i}^{(0)}} := \mathbf{x}_i$ and $\mathbf{h_{i}^{(K)}} := \mathbf{h_{i}}$.
\end{definition}

The above formulation in \shortautoref{eq: graph convolution with message passing} is referred to as \textit{spatial GNNs}, as opposed to \textit{spectral GNNs} which defines convolution from the lens of spectral graph theory. We refer the reader to recent publication~\cite{jin2023powerful} for a deeper analysis of spectral versus spatial GNNs, and \cite{wu2020comprehensive} for a comprehensive review of GNNs.

To employ GNNs for time series analysis, it is implied that a graph structure must be provided. However, not all time series data have readily available graph structures and, in practice, two types of strategies are utilized to generate the missing graph structures from the data: \textit{heuristics} or \textit{learned} from data. \\

\noindent\textbf{Heuristic-based Graphs.} This group of methods extracts graph structures from data based on heuristics, such as:
\begin{itemize}

\item \textit{Spatial Proximity}: This approach defines the graph structure by considering the proximity between pairs of nodes based on, e.g., their geographical location. A typical example is the construction of the adjacency matrix $\mathbf{A}$ based on the shortest travel distance between nodes when the time series data have geospatial properties:
\begin{equation}
\mathbf{A}_{i,j} =
\begin{cases}
\frac{1}{d{ij}}, & \text{if}\ d_{ij} \neq 0, \\
0, & \text{otherwise},
\end{cases}
\label{eq:pairwise_proximity}
\end{equation}
where $d_{ij}$ denotes the shortest travel distance between node $i$ and node $j$. Some common kernel functions, e.g., Gaussian radial basis, can also be applied~\cite{jin2023spatio}.

\item \textit{Pairwise Connectivity}: In this approach, the graph structure is determined by the connectivity between pairs of nodes, like that determined by transportation networks. The adjacency matrix $\mathbf{A}$ is defined as:
\begin{equation}
\mathbf{A}_{i,j} =
\begin{cases}
1, & \text{if}\ v_i\ \text{and}\ v_j\ \text{are directly linked}, \\
0, & \text{otherwise}.
\end{cases}
\label{eq:pairwise_connectivity}
\end{equation}
Typical scenarios include edges representing roads, railways, or adjacent regions~\cite{geng2019spatiotemporal, he2020towards}. In such cases, the graph can be undirected or directed, resulting in symmetric and asymmetric adjacency matrices.

\item \textit{Pairwise Similarity}: This method constructs the graph by connecting nodes with similar attributes. A simple example is the construction of adjacency matrix $\mathbf{A}$ based on the cosine similarity between time series:
\begin{equation}
    \mathbf{A}_{i,j} = \frac{\mathbf{x}_i^\top\mathbf{x}_j}{\|\mathbf{x}_i\|\|\mathbf{x}_j\|},
\end{equation}
where $\|\cdot\|$ denotes the Euclidean norm. There are also several variants for creating similarity-based graphs, such as Pearson correlation coefficient (PCC)~\cite{zhang2020crowd} and dynamic time warping (DTW)~\cite{li2021spatial}.

\item \textit{Functional Dependence}: This approach defines the graph structure based on \revision{known} functional dependencies between pairs of nodes\revision{, such as direct causal relationships or dependence from common hidden factors. For instance, adjacency matrix $\mathbf{A}$ can be constructed based on Granger causality~\cite{yi2012sparse} as}
\begin{equation}
    \mathbf{A}_{i,j} = \begin{cases}
        1, & \text{if node}\ j\ \text{Granger-causes}\\
        & \text{node}\ i\ \text{at a significance level}\ \alpha, \\
        0, & \text{otherwise}.
    \end{cases}
\end{equation}
Other examples involve transfer entropy (TE)~\cite{wang2022mthetgnn} and directed phase lag index (DPLI)~\cite{grattarola2019change}.
\revision{While overlap with previous heuristics exists, functional relations typically represent or estimate actual node dependencies in the data-generating process.}
\end{itemize}

\noindent\textbf{Learning-based Graphs.} In contrast to heuristic-based methods, learning-based approaches aim to learn the graph structure directly from the data and end-to-end with the downstream task. \revision{Accordingly, graph $\mathbf{A}$ can be defined as a function  $\rho({}\cdot{})$ of some trainable model parameters $\Theta$ and, possibly, also of the time series observations $\mathbf{X}$, i.e., $\mathbf{A} = \rho(\Theta, \mathbf{X})$.
Embedding-based approaches (e.g., \cite{wu2019graph, bai2020adaptive}) define the presence of each edge by comparing learned embedding vectors of the associated nodes; as an example, consider $\mathbf{A}_{i,j} = \text{ReLU}(\Theta_i^\top \Theta_j)$ with $\Theta_i,\Theta_j$ node embedding of nodes $i,j$, respectively.
Another common example involves defining $\mathbf{A}$ through attention scores computed between node signals \cite{cao2020spectral}. 
Sparsification methods -- such as the ReLU activation above or top-k selection -- are often applied to discard edges with null or low weights, thereby reducing the computational load requested by dense GNN operations \cite{wu2020connecting}.
An alternative strategy is modeling $\mathbf{A}$ as a discrete random variable whose parametric distribution, say $p_\Theta(\mathbf{A}\mid\mathbf{X})$, is learned alongside the other model parameters \cite{shang2021discrete, cini2023sparse}. Learning-based approaches enable the data-driven discovery of less obvious graph structures that are tailored to solve the given task, potentially providing more effective relations than heuristic-based graphs.}
\begin{figure*}[t]
	\centering
	\includegraphics[width=0.74\textwidth]{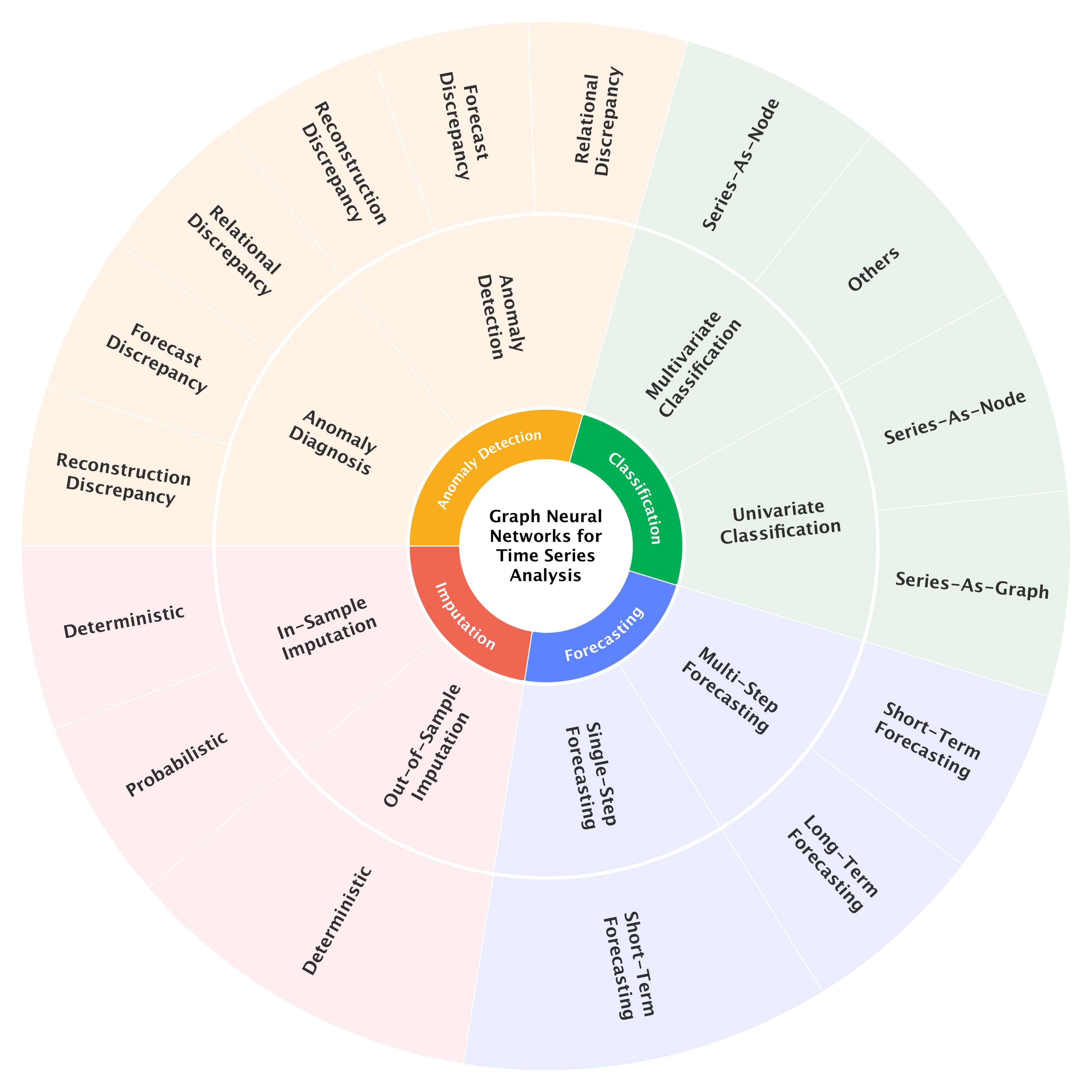}
        \caption{Task-oriented taxonomy of graph neural networks for time series analysis in the existing literature.}
	\label{fig:gnn4ts-task-taxonomy}
\end{figure*}

\section{Framework and Categorization}\label{sec:categorization}
In this section, we present a comprehensive task-oriented taxonomy for GNNs within the context of time series analysis (\shortautoref{sec:general taxonomy}). Subsequently, we investigate how to encode time series across various tasks by introducing a unified methodological framework for GNN architectures (\shortautoref{sec:framework-stgnn}). 
According to the framework, all architectures are composed of a similar graph-based processing module $f_\theta$ and a second module $p_\phi$ specialized in downstream tasks.

\subsection{Task-oriented Taxonomy}\label{sec:general taxonomy}

\begin{figure*}[t]
    \centering
    \subfloat[Graph neural networks for time series forecasting.]{\includegraphics[width=0.4\textwidth]{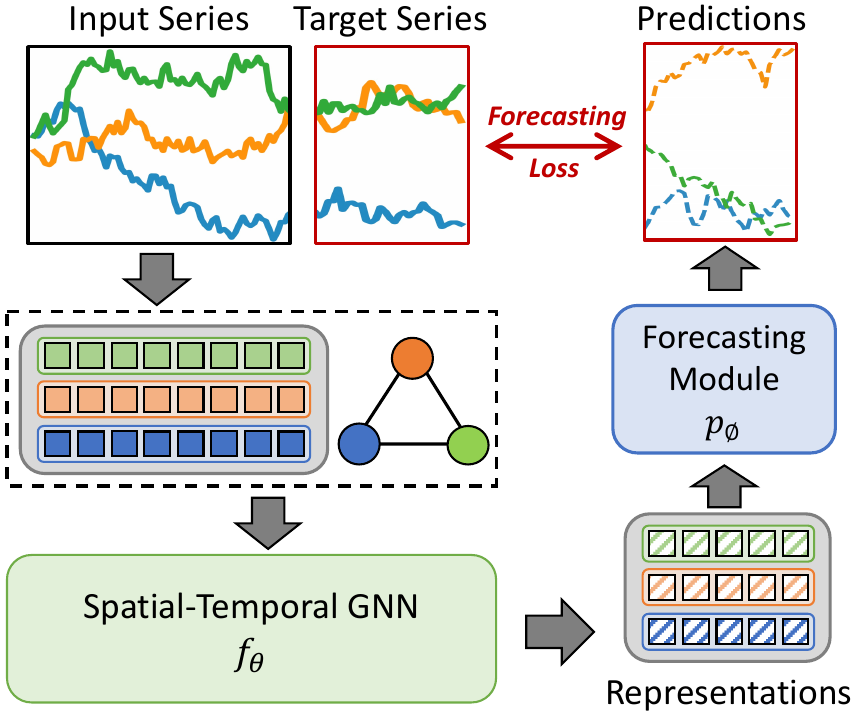}\label{fig:gnn4tsf}}
    \hspace{0.06\textwidth}
    \subfloat[Graph neural networks for time series anomaly detection.]{\includegraphics[width=0.4\textwidth]{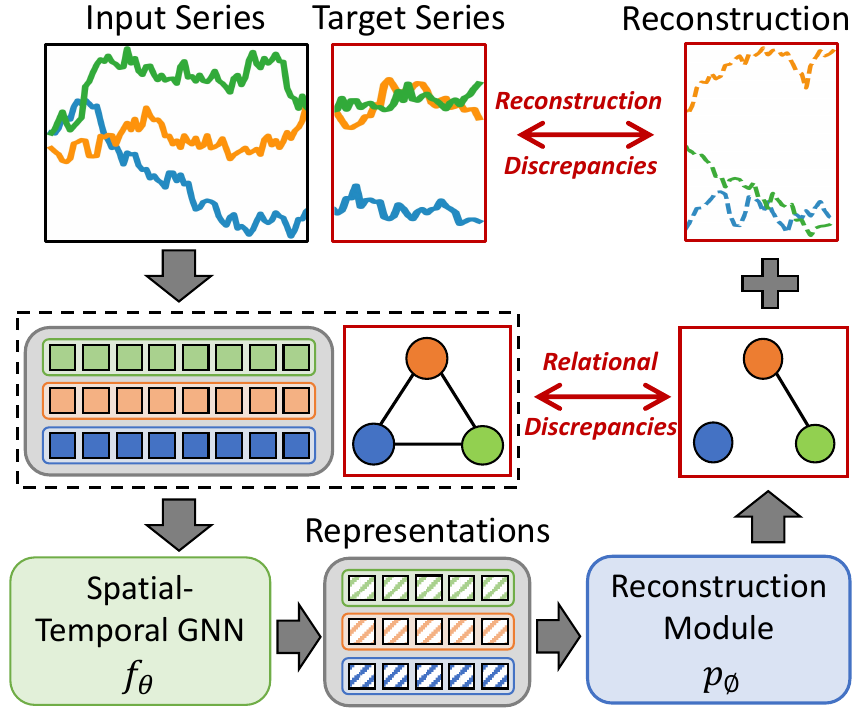}\label{fig:gnn4tsad}}
    \vspace{0.02\textwidth}
    \subfloat[Graph neural networks for time series imputation.]
    {\includegraphics[width=0.4\textwidth]{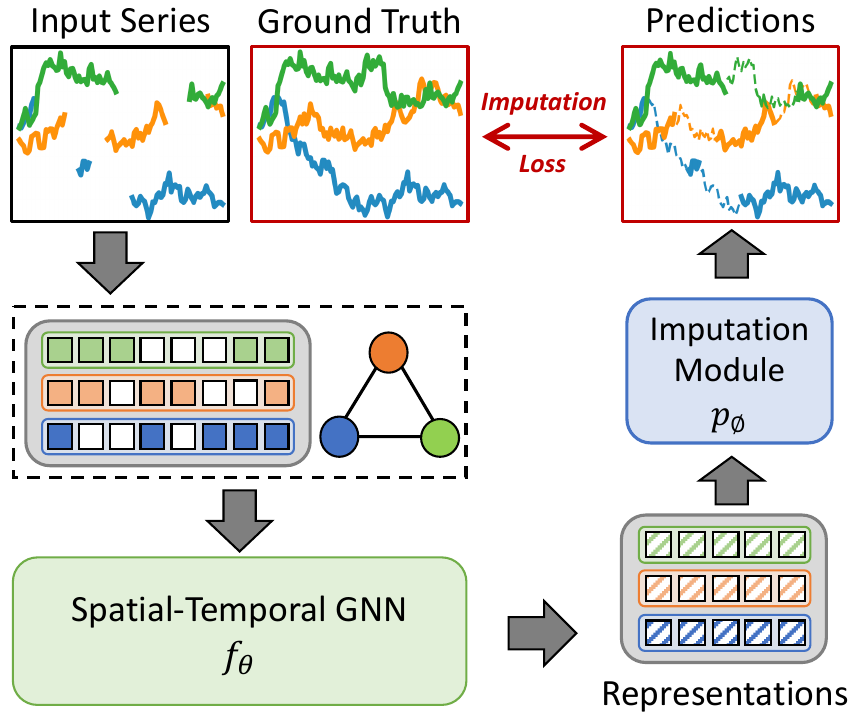}\label{fig:gnn4tsi}}
    \hspace{0.06\textwidth}
    \subfloat[Graph neural networks for time series classification: Formulates green series classification as a graph (top) or node (bottom) classification task.]{\includegraphics[width=0.4\textwidth]{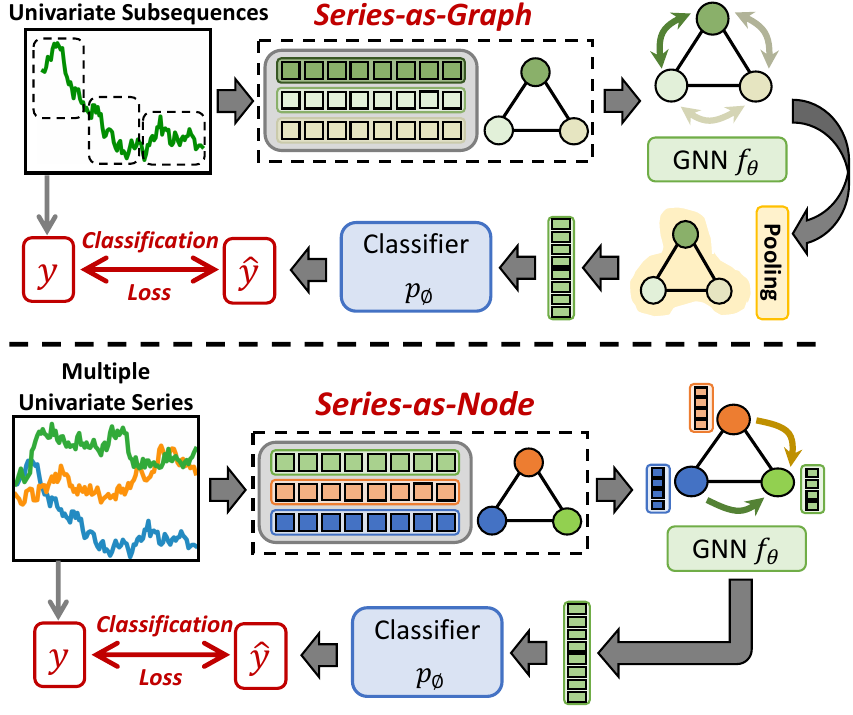}\label{fig:gnn4tsc}}
    \caption{Four categories of graph neural networks for time series analysis. For the sake of simplicity and illustrative purposes, we assume the graph structures are fixed in all subplots.}
    \label{fig:gnn4ts-taxonomy}
\end{figure*}

In \shortautoref{fig:gnn4ts-task-taxonomy}, we illustrate a task-oriented taxonomy of GNNs encompassing the primary tasks and mainstream modeling for time series analysis, and showcasing the potential of GNN4TS. This survey focuses on four categories: time series \textit{forecasting}, \textit{anomaly detection}, \textit{imputation}, and \textit{classification}. These tasks are performed on top of the time series representations learned by \textit{spatial-temporal graph neural networks} (STGNNs), which serve as the foundation for encoding time series data in existing literature across various tasks. We detail this in \shortautoref{sec:framework-stgnn}. \\

\noindent\textbf{Time Series Forecasting.} This task is centered around predicting future values of the time series based on historical observations, as depicted in \shortautoref{fig:gnn4tsf}. Depending on application needs, we categorize this task into two types: \textit{single-step-ahead forecasting} and \textit{multi-step-ahead forecasting}. The former is meant to predict single future observations of the time series once at a time, i.e., the target at time $t$ is $\mathbf Y := \mathbf X_{t+H}$ for some $H \in \mathbb N$ steps ahead, while the latter makes predictions for a time interval, e.g., $\mathbf Y := \mathbf X_{t+1:t+H}$. Parameterized solutions to both predictive cases can be derived by optimizing
\begin{equation}
\theta^*, \phi^* = \mathop{\arg\min}\limits_{{\theta, \phi}}\mathcal{L}_\textsc{f}\Big( p_{\phi}\big(f_{\theta}(\mathbf{X}_{t-T:t}, \mathbf{A}_{t-T:t})\big), \mathbf{Y} \Big),
\label{eq:forecasting formula}
\end{equation}
where $f_{\theta}(\cdot)$ and $p_{\phi}(\cdot)$ represent a spatial-temporal GNN and the predictor, respectively. Details regarding the $f_{\theta}(\cdot)$ architecture are given in \shortautoref{sec:framework-stgnn} while the predictor is, normally, a multi-layer perceptron. In the sequel, we denote by $\mathbf{X}_{t-T:t}$ and $\mathbf{A}_{t-T:t}$ a spatial-temporal graph $\mathcal{G} = \{\mathcal{G}_{t-T}, \mathcal{G}_{t-T+1}, \cdots, \mathcal{G}_{t}\}$ with length $T$. If the underlying graph structure is fixed, then $\mathbf{A}_{t} := \mathbf{A}$. $\mathcal{L}_\textsc{f}(\cdot)$ denotes the forecasting loss, which is typically a squared or absolute loss function, e.g., STGCN~\cite{yu2018spatio} and MTGNN~\cite{wu2020connecting}. Most existing works minimize the error between the forecasting and the ground truth $\mathbf{Y}$ through \shortautoref{eq:forecasting formula}; this process is known as \textit{deterministic} time series forecasting. Besides, we have \textit{probabilistic} time series forecasting methods, such as DiffSTG~\cite{wen2023diffstg}, that share the same objective \shortautoref{eq:forecasting formula} function though it is not directly optimized. Based on the size of the forecasting horizon $H$, we end up in either \textit{short-term} or \textit{long-term forecasting}. \\

\noindent\textbf{Time Series Anomaly Detection.} This task focuses on detecting irregularities and unexpected \revision{events} in time series data (\shortautoref{fig:gnn4tsad}). Detecting anomalies requires determining \emph{when} the anomalous event occurred, while \emph{diagnosing} them requests gaining insights about how and why the anomaly occurred. Due to the general difficulty of acquiring anomaly events, current research commonly treats anomaly detection as an unsupervised problem that involves the design of a model describing normal, non-anomalous data. The learned model is then used to detect anomalies by generating a high score whenever an anomaly event occurs. This model learning process mirrors the forecasting optimization, \shortautoref{eq:forecasting formula}, with $f_{\theta}(\cdot)$ and $p_{\phi}(\cdot)$ denote the spatial-temporal GNN and the predictor, respectively. In general, the spatial-temporal GNN and the predictor are trained on normal, non-anomalous data using either forecasting~\cite{deng2021graph,han2022learning} or reconstruction~\cite{zhao2020multivariate,dai2022graphaugmented} optimization approaches, with the aim of minimizing the discrepancy between the normal input and the forecast (or reconstructed) series. Nonetheless, when these models are put to use for detecting anomalies, they are expected to fail in minimizing this discrepancy upon receiving anomalous input. This inability to conform to the expected low-discrepancy model behavior during anomaly periods creates a detectable difference, facilitating the detection of anomalies. The threshold separating normal and anomalous data is a sensitive hyperparameter that should be set considering the rarity of anomalies and aligned with a desired false alarm rate~\cite{zambon2018concept}. Lastly, to diagnose the causes of anomalies, a common strategy involves calculating discrepancies for each channel node and consolidating these into a single anomaly score~\cite{zambon2023where}. This approach allows for the identification of the channel variables responsible for the anomaly events by calculating their respective contributions to the final score.
\\

\noindent\textbf{Time Series Imputation.} This task is centered around estimating and filling in missing or incomplete data points within a time series (\shortautoref{fig:gnn4tsi}). Current research in this domain can be broadly classified into two main approaches: \textit{in-sample imputation} and \textit{out-of-sample imputation}. In-sample imputation involves filling missing values in a given time series, while out-of-sample imputation pertains to inferring missing data not present in the training dataset. We formulate the learning objective as follows:
\begin{equation}
    \theta^*, \phi^* = \mathop{\arg\min}\limits_{{\theta, \phi}}\mathcal{L}_\textsc{i}\Big( p_{\phi}\big(f_{\theta}(\Tilde{\mathbf{X}}_{t-T:t}, \mathbf{A}_{t-T:t})\big), \mathbf{X}_{t-T:t} \Big),
\label{eq:imputation formula}
\end{equation}
where $f_{\theta}(\cdot)$ and $p_{\phi}(\cdot)$ denote the spatial-temporal GNN and imputation module to be learned, respectively. The imputation module can e.g., be a multi-layer perceptron. In this task, $\Tilde{\mathbf{X}}_{t-T:t}$ represents input time series data with missing values (reference time series), while $\mathbf{X}_{t-T:t}$ denotes the same time series without missing values. As it is impossible to access the reference time series during training, a surrogate optimization objective is considered, such as generating synthetic missing values~\cite{cini2022filling}. In \shortautoref{eq:imputation formula}, $\mathcal{L}_\textsc{i}(\cdot)$ refers to the imputation loss, which can be, for instance, an absolute or a squared error, similar to forecasting tasks. For in-sample imputation, the model is trained and evaluated on $\Tilde{\mathbf{X}}_{t-T:t}$ and $\mathbf{X}_{t-T:t}$. Instead, for out-of-sample imputation, the model is trained and evaluated on disjoint sequences, e.g., trained on $\Tilde{\mathbf{X}}_{t-T:t}$ but evaluated on $\mathbf{X}_{t:t+H}$, where the missing values in $\Tilde{\mathbf{X}}_{t:t+H}$ will be estimated. Similar to time series forecasting and anomaly detection, the imputation process can be either \textit{deterministic} or \textit{probabilistic}. The former predicts the missing values directly (e.g., GRIN~\cite{cini2022filling}), while the latter estimates the missing values from data distributions (e.g., PriSTI~\cite{liu2023pristi}).\\

\tikzstyle{my-box}=[
    rectangle,
    draw=hidden-draw,
    rounded corners,
    text opacity=1,
    minimum height=1.5em,
    minimum width=5em,
    inner sep=2pt,
    align=center,
    fill opacity=.5,
    line width=0.8pt,
]

\tikzstyle{leaf}=[my-box, minimum height=1.5em,
    fill=hidden-pink!80, text=black, align=left, font=\normalsize,
    inner xsep=2pt,
    inner ysep=4pt,
    line width=0.8pt,
]

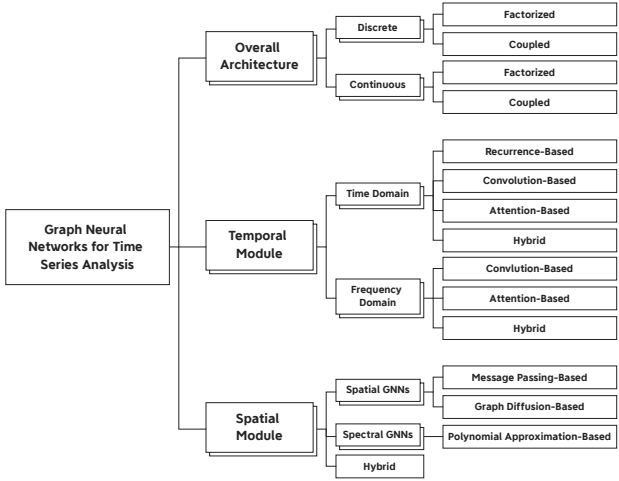
\begin{figure}[t]
    \centering
    \resizebox{0.48\textwidth}{!}
    {
        \begin{forest}
            forked edges,
            for tree={
                fill=level0!80,
                grow=east,
                reversed=true,
                anchor=base west,
                parent anchor=east,
                child anchor=west,
                base=left,
                font=\large,
                rectangle,
                draw=hidden-draw,
                rounded corners,
                align=left,
                minimum width=4em,
                edge+={darkgray, line width=1pt},
                s sep=3pt,
                inner xsep=2pt,
                inner ysep=3pt,
                line width=0.8pt,
                ver/.style={rotate=90, child anchor=north, parent anchor=south, anchor=center},
            },
            where level=1{text width=8.2em,font=\normalsize,fill=level1!80,}{},
            where level=2{text width=6.5em,font=\normalsize,fill=level2!80,}{},
            where level=3{text width=9.5em,font=\normalsize,fill=level3!60,}{},
            [GNNs for Time \\Series Analysis
                [Overall \\Architecture
                    [Discrete
                        [Factorized
                            [\textbf{Forecasting}: \cite{yu2018spatio}{,}\cite{geng2019spatiotemporal}{,}\cite{cao2020spectral}{,}\cite{li2021spatial}{,}\cite{liu2022multivariate}{,}\cite{shao2022pre}, leaf, text width=18.8em]
                            [\textbf{Others}: \cite{zheng2023correlation}{,}\cite{zha2022towards}{,}\cite{liu2023todynet}{,}\cite{ye2021spatial}{,}\cite{wu2021spatial}{,}\cite{mariscalearning}{,}\cite{liu2023pristi}, leaf, text width=19em]
                        ]
                        [Coupled
                            [\textbf{Forecasting}: \cite{lidiffusion}{,}\cite{chen2020multi}{,}\cite{zheng2020gman}{,}\cite{shang2021discrete}{,}\cite{chen2021z}{,}\cite{yu2022regularized}, leaf, text width=18.8em]
                            [\textbf{Others}: \cite{deng2021graph}{,}\cite{han2022learning}{,}\cite{zhang2022graphguided}{,}\cite{wu2021inductive}{,}\cite{cini2022filling}, leaf, text width=15em]
                        ]
                    ]
                    [Continuous
                        [Factorized
                            [\textbf{Forecasting and other tasks}: \cite{fang2021spatial}{,}\cite{dai2022graphaugmented}, leaf, text width=17em]
                        ]
                        [Coupled
                            [\textbf{Forecasting}: \cite{jin2022multivariate}{,}\cite{choi2022graph}, leaf, text width=10em]
                        ]
                    ]
                ]
                [Temporal Module
                    [Time Domain
                        [Recurrence-based
                            [\textbf{Forecasting}: \cite{lidiffusion}{,}\cite{pan2019urban}{,}\cite{geng2019spatiotemporal}{,}\cite{chen2020multi}{,}\cite{bai2020adaptive}{,}\cite{cini2023scalable}, leaf, text width=18.7em]
                            [\textbf{Others}: \cite{hu2021time}{,}\cite{chen2022deep}{,}\cite{han2022learning}{,}\cite{zhou2022hybrid}{,}\cite{cini2022filling}{,}\cite{liang2022memory}, leaf, text width=17em]
                        ]
                        [Convolution-based
                            [\textbf{Forecasting}: \cite{yu2018spatio}{,}\cite{wu2019graph}{,}\cite{wu2020connecting}{,}\cite{chen2021z}{,}\cite{li2021spatial}{,}\cite{jin2022multivariate}, leaf, text width=19em]
                            [\textbf{Others}: \cite{chen2023DyGraphAD}{,}\cite{zheng2023correlation}{,}\cite{zha2022towards}{,}\cite{liu2023todynet}{,}\cite{ye2021spatial}{,}\cite{wu2021spatial}, leaf, text width=17em]
                        ]
                        [Attention-based
                            [\textbf{Forecasting}: \cite{zheng2020gman}{,}\cite{yu2020spatio}{,}\cite{liu2022multivariate}{,}\cite{shao2022pre}, leaf, text width=15em]
                            [\textbf{Others}: \cite{zhao2020multivariate}{,}\cite{wu2021event2graph}{,}\cite{zhang2022graphguided}{,}\cite{mariscalearning}{,}\cite{wu2022multi}{,}\cite{liu2023pristi}, leaf, text width=16.5em]
                        ]
                        [Hybrid
                            [\textbf{Forecasting}: \cite{guo2019attention}{,}\cite{wang2020traffic}{,}\cite{lan2022dstagnn}, leaf, text width=12em]
                            [\textbf{Anomaly Detection}: \cite{chen2021learning}{,}\cite{zhang2022grelen}, leaf, text width=14em]
                        ]
                    ]
                    [Frequency \\Domain
                        [Convolution-based
                            [\textbf{Forecasting}: \cite{cao2020spectral}{,}\cite{xia2023deciphering}, leaf, text width=10em]
                        ]
                        [Hybrid
                            [\textbf{Forecasting}: \cite{jin2023powerful}, leaf, text width=8em]
                        ]
                    ]
                ]
                [Spatial Module
                    [Spatial GNNs
                        [Message \\passing-based
                            [\textbf{Forecasting}: \cite{pan2019urban}{,}\cite{song2020spatial}{,}\cite{bai2020adaptive}{,}\cite{fang2021spatial}{,}\cite{li2021spatial}{,}\cite{cui2021metro}, leaf, text width=19em]
                            [\textbf{Others}: \cite{zhao2020multivariate}{,}\cite{deng2021graph}{,}\cite{chen2022deep}{,}\cite{zhang2022graphguided}{,}\cite{liu2023todynet}{,}\cite{ye2021spatial}, leaf, text width=17em]
                        ]
                        [Graph \\diffusion-based
                            [\textbf{Forecasting}: \cite{lidiffusion}{,}\cite{wu2019graph}{,}\cite{shang2021discrete}{,}\cite{cirstea2019graph}, leaf, text width=14.5em]
                            [\textbf{Imputation}: \cite{wu2021inductive}, leaf, text width=8em]
                        ]
                    ]
                    [Spectral GNNs
                        [Polynomial \\approximation-based
                            [\textbf{Forecasting}: \cite{yu2018spatio}{,}\cite{guo2019attention}{,}\cite{cao2020spectral}{,}\cite{lan2022dstagnn}{,}\cite{yu2022regularized}{,}\cite{jin2023powerful}, leaf, text width=18.6em]
                        ]
                    ]
                    [Hybrid
                            [\textbf{Forecasting}: \cite{zhang2020spatio}, leaf, text width=8em]
                    ]
                ]
            ]
        \end{forest}
    }    
\caption{\revision{Methodology-oriented taxonomy of graph neural networks for time series analysis.}}
\label{fig:gnn4ts-na-taxonomy}
\end{figure}    

\noindent\textbf{Time Series Classification.} This task aims to assign a categorical label to a given time series based on its underlying patterns or characteristics. Rather than capturing patterns within a time series data sample, the essence of time series classification resides in discerning differentiating patterns that help separate samples based on their class labels. The optimization problem can be expressed as:
\begin{equation}
    \theta^*, \phi^* = \mathop{\arg\min}\limits_{{\theta, \phi}}\mathcal{L}_\textsc{c}\Big( p_{\phi}\big(f_{\theta}(\mathbf{X}, \mathbf{A})\big), \mathbf{Y} \Big),
\label{eq:classification formula}
\end{equation}
where $f_{\theta}(\cdot)$ and $p_{\phi}(\cdot)$ denote, e.g., a GNN and a classifier to be learned, respectively. Using univariate time series classification as an example \shortautoref{fig:gnn4tsc}, the task can be formulated as either a graph or node classification task. In the case of graph classification \revision{(\textit{Series-as-Graph})} \cite{cheng2021time2graph+}, each series is transformed into a graph, and the graph will be the input of a GNN to generate a classification output. This can be achieved by dividing a series into multiple subsequences with a window size, $W$, serving as graph nodes, $\mathbf{X} \in \mathbb{R}^{N \times W}$, and an adjacency matrix, $\mathbf{A}$, describing the relationships between subsequences. A simple GNN, $f_{\theta}(\cdot)$, then employs graph convolution and pooling to obtain a condensed graph feature to be exploited by a classifier $p_{\phi}(\cdot)$ which assigns a class label to the graph. Alternatively, the node classification formulation \revision{(\textit{Series-as-Node})}, treats each series as a node in a dataset graph. \revision{Series-as-Node} constructs an adjacency matrix representing the relationships between multiple distinct series in a given dataset \cite{zha2022towards}. With several series of length $T$ stacked into a matrix $\mathbf{X} \in \mathbb{R}^{N \times T}$ as node features and $\mathbf{A}$ representing pairwise relationships, the GNN operation, $f_{\theta}(\cdot)$, aims at leveraging the relationships across different series for accurate node series classification \cite{xi2023lb}. In all cases, $\mathbf{Y}$ is typically a one-hot encoded vector representing the categorical label of a univariate or multivariate time series.

\subsection{Unified Methodological Framework}\label{sec:framework-stgnn}

In \shortautoref{fig:gnn4ts-na-taxonomy}, we present a unified methodological framework of STGNNs mentioned in \shortautoref{sec:general taxonomy} for time series analysis. Specifically, our framework serves as the basis for encoding time series data in the existing literature across various downstream tasks (\shortautoref{fig:gnn4ts-task-taxonomy}). As an extension, STGNNs incorporate spatial information by considering the relationships between nodes in the graph and temporal information by taking into account the evolution of node attributes over time. Similar to \cite{jin2023spatio}, we systematically categorize STGNNs from three perspectives: \textit{spatial module}, \textit{temporal module}, and \textit{overall model architecture}.

\textbf{Spatial Module.} To model dependencies between time series over time, STGNNs employ the design principles of GNNs on static graphs. These can be further categorized into three types: \textit{spectral GNNs}, \textit{spatial GNNs}, and a combination of both (i.e., \textit{hybrid}) \cite{wu2020comprehensive}. Spectral GNNs are based on spectral graph theory and use the graph shift operator (like the graph Laplacian) to capture node relationships in the graph frequency domain~\cite{shuman2013emerging,sandryhaila2013discrete,jin2023powerful}. Differently, spatial GNNs simplify spectral GNNs by directly designing filters that are localized to each node's neighborhood. 
\revision{The approaches in this category can be broadly classified into two types: \textit{graph diffusion-based} and \textit{message passing-based}. Notably, graph transformers~\cite{ying2021transformers} represent a specialized extension of message passing neural networks~\cite{cai2023connection}, further expanding the capabilities of this paradigm.}
Hybrid approaches combine both spectral and spatial methodologies to capitalize on the strengths of each method.

\textbf{Temporal Module.} To account for temporal dependencies in time series, STGNNs incorporate temporal modules that work in tandem with spatial modules to model intricate spatial-temporal patterns. Temporal dependencies can be represented in either the \textit{time} or \textit{frequency} domains. The approaches in the first category encompass \textit{recurrence-based} (e.g., RNNs~\cite{connor1994recurrent}), \textit{convolution-based} (e.g., TCNs~\cite{wan2019multivariate}), \textit{attention-based} (e.g., transformers~\cite{wen2022transformers}), and a combination of these (i.e., \textit{hybrid}). For the second category, analogous techniques are employed, followed by orthogonal space projections~\cite{jin2023powerful}, such as the Fourier transform.

\textbf{Model Architecture.} To integrate the two modules, existing STGNNs are either \textit{discrete} or \textit{continuous} in terms of their overall neural architectures. Both types can be further subdivided into two subcategories: \textit{factorized} and \textit{coupled}. With typical factorized STGNN model architectures, the temporal processing is performed either before or after the spatial processing, whether in a discrete (e.g., STGCN~\cite{yu2018spatio}) or continuous manner (e.g., STGODE~\cite{fang2021spatial}). Conversely, the coupled model architecture refers to instances where spatial and temporal modules are interleaved, such as DCRNN~\cite{lidiffusion} (discrete) and MTGODE~\cite{jin2022multivariate} (continuous). Other authors refer to very related categories as time-then-space and time-and-space \cite{gao2022equivalence}. \\

\begin{figure}[t]
    \centering
    \includegraphics[width=0.49\textwidth]{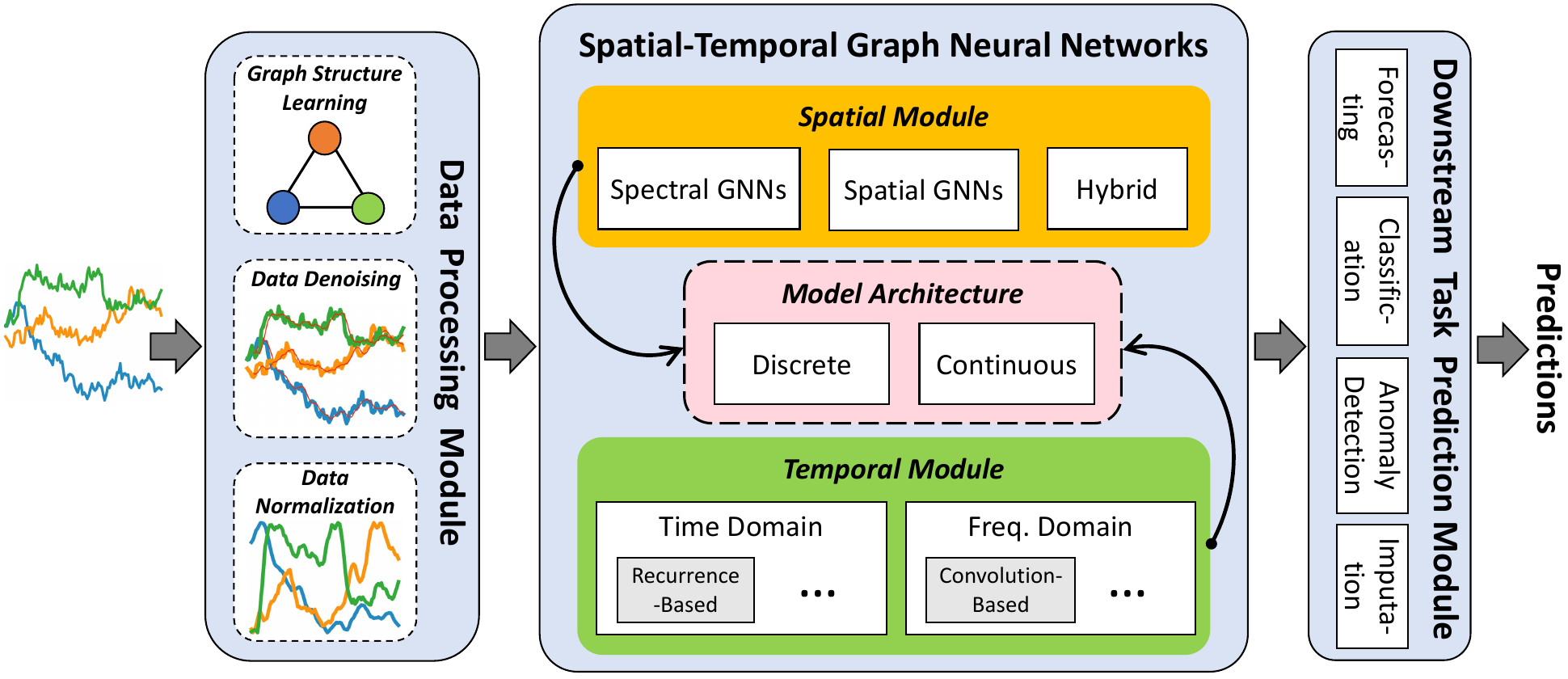}
    \caption{General pipeline for time series analysis using graph neural networks.}
    \label{fig:gnn4ts-pipeline}
    \vspace{-2mm}
\end{figure}

\noindent\textbf{General Pipeline.} In \shortautoref{fig:gnn4ts-pipeline}, we showcase a general pipeline that shows how STGNNs can be integrated into time series analysis. Given a time series dataset, we first process it using the \textit{data processing module}, which performs essential data cleaning and normalization tasks, including the extraction of time series topology (i.e., graph structures). Subsequently, STGNNs are utilized to obtain time series representations, which can then be passed to different handlers (i.e., \textit{downstream task prediction module}) to execute various analytical tasks, such as forecasting and anomaly detection.
\section{GNNs for Time Series Forecasting}\label{sec:forecasting}

\begin{table*}[thp]
\caption{Summary of representative graph neural networks for time series forecasting. \textit{Task notation}: The first letter, ``M'' or ``S'', indicates multi-step or single-step forecasting, and the second letter, ``S'' or ``L'', denotes short-term or long-term forecasting. \textit{Architecture notation}: ``D'' and ``C'' represent ``Discrete'' and ``Continuous''; ``C'' and ``F'' stand for ``Coupled'' and ``Factorized''. \textit{Temporal module notation}: ``T'' and ``F'' signify ``Time'' and ``Frequency'' domains; ``R'', ``C'', ``A'', and ``H'' correspond to ``Recurrence'', ``Convolution'', ``Attention'', and ``Hybrid''. \textit{Input graph notation}: ``R'' indicates that a pre-calculated graph structure (with a certain graph heuristic) is a required input of the model, ``NR'' that such graph is not required (not a model's input), while ``O'' signifies that the model can optionally exploit given input graphs. \textit{Notation of learned graph relations}: ``S'' and ``D'' indicate ``Static'' and ``Dynamic''. \textit{Notation of adopted graph heuristics}: ``SP'', ``PC'', ``PS'', and ``FD'' denote ``Spatial Proximity'', ``Pairwise Connectivity'', ``Pairwise Similarity'', and ``Functional Dependency''. The ``Missing Values'' column indicates whether corresponding methods can handle missing values in input time series.
}
\label{tab:forecasting summary}
\centering
\begin{adjustbox}{width=2\columnwidth,center}
\begin{tabular}{lcccccccccc}
\toprule
\thead{\textbf{Approach}} & \thead{\textbf{Year}} & \thead{\textbf{Venue}} & \thead{\textbf{Task}} & \thead{\textbf{Architecture}} & \thead{\textbf{Spatial} \\ \textbf{Module}} & \thead{\textbf{Temporal} \\ \textbf{Module}} & \thead{\textbf{Missing} \\ \textbf{Values}} &\thead{\textbf{Input} \\ \textbf{Graph}} & \thead{\textbf{Learned} \\ \textbf{Relations}} & \thead{\textbf{Graph} \\ \textbf{Heuristics}} \\ \midrule
\rowcolor{lightgray!20}DCRNN~\cite{lidiffusion} & 2018 & ICLR & M-S & D-C & Spatial GNN & T-R & No & R & - & SP \\
STGCN~\cite{yu2018spatio} & 2018 & IJCAI & M-S & D-F & Spectral GNN & T-C & No & R & - & SP \\
\rowcolor{lightgray!20}ST-MetaNet~\cite{pan2019urban} & 2019 & KDD & M-S & D-F & Spatial GNN & T-R & No & R & - & SP, PC \\
NGAR~\cite{zambon2019autoregressive} & 2019 & IEEE IJCNN & S-S & D-F & Spatial GNN & T-R & No & R & - & - \\
\rowcolor{lightgray!20}ASTGCN~\cite{guo2019attention} & 2019 & AAAI & M-S & D-F & Spectral GNN & T-H & No & R & - & SP, PC \\
ST-MGCN~\cite{geng2019spatiotemporal} & 2019 & AAAI & S-S & D-F & Spectral GNN & T-R & No & R & - & SP, PC, PS \\
\rowcolor{lightgray!20}Graph WaveNet~\cite{wu2019graph} & 2019 & IJCAI & M-S & D-F & Spatial GNN & T-C & No & O & S & SP \\
MRA-BGCN~\cite{chen2020multi} & 2020 & AAAI & M-S & D-C & Spatial GNN & T-R & No & R & - & SP \\
\rowcolor{lightgray!20}MTGNN~\cite{wu2020connecting} & 2020 & KDD & S-S, M-S & D-F & Spatial GNN & T-C & No & NR & S & - \\
STGNN*~\cite{wang2020traffic} & 2020 & WWW & M-S & D-C & Spatial GNN & T-H & No & R & - & SP \\
\rowcolor{lightgray!20}GMAN~\cite{zheng2020gman} & 2020 & AAAI & M-S & D-C & Spatial GNN & T-A & No & R & - & SP \\
SLCNN~\cite{zhang2020spatio} & 2020 & AAAI & M-S & D-F & Hybrid & T-C & No & NR & S & - \\
\rowcolor{lightgray!20}STSGCN~\cite{song2020spatial} & 2020 & AAAI & M-S & D-C & Spatial GNN & T & No & R & - & PC \\
StemGNN~\cite{cao2020spectral} & 2020 & NeurIPS & M-S & D-F & Spectral GNN & F-C & No & NR & S & - \\
\rowcolor{lightgray!20}AGCRN~\cite{bai2020adaptive} & 2020 & NeurIPS & M-S & D-C & Spatial GNN & T-R & No & NR & S & - \\
LSGCN~\cite{huang2020lsgcn} & 2020 & IJCAI & M-S & D-F & Spectral GNN & T-C & No & R & - & SP \\
\rowcolor{lightgray!20}STAR~\cite{yu2020spatio} & 2020 & ECCV & M-S & D-F & Spatial GNN & T-A & No & R & - & PC \\
GTS~\cite{shang2021discrete} & 2021 & ICLR & M-S & D-C & Spatial GNN & T-R & No & NR & S & - \\
\rowcolor{lightgray!20}GEN~\cite{paassen2021graph} & 2021 & ICLR & S-S & D-F & Spatial GNN & T-R & No & R & - & - \\
Z-GCNETs~\cite{chen2021z} & 2021 & ICML & M-S & D-C & Spatial GNN & T-C & No & NR & S & - \\
\rowcolor{lightgray!20}STGODE~\cite{fang2021spatial} & 2021 & KDD & M-S & C-F & Spatial GNN & T-C & No & R & - & SP, PS \\
STFGNN~\cite{li2021spatial} & 2021 & AAAI & M-S & D-F & Spatial GNN & T-C & No & R & - & SP, PS \\
\rowcolor{lightgray!20}DSTAGNN~\cite{lan2022dstagnn} & 2022 & ICML & M-S & D-F & Spectral GNN & T-H & No & R & - & PC, PS \\
TPGNN~\cite{liu2022multivariate} & 2022 & NeurIPS & S-S, M-S & D-F & Spatial GNN & T-A & No & NR & D & - \\
\rowcolor{lightgray!20}MTGODE~\cite{jin2022multivariate} & 2022 & IEEE TKDE & S-S, M-S & C-C & Spatial GNN & T-C & No & NR & S & - \\
STG-NCDE~\cite{choi2022graph} & 2022 & AAAI & M-S & C-C & Spatial GNN & T-C & Yes & NR & S & - \\
\rowcolor{lightgray!20}STEP~\cite{shao2022pre} & 2022 & KDD & M-S & D-F & Spatial GNN & T-A & No & NR & S & - \\
Chauhan et al.~\cite{chauhan2022multi} & 2022 & KDD & M-S & - & - & - & Yes & O & S & SP \\
\rowcolor{lightgray!20}RGSL~\cite{yu2022regularized} & 2022 & IJCAI & M-S & D-C & Spectral GNN & T-R & No & R & S & SP, PC \\
FOGS~\cite{rao2022fogs} & 2022 & IJCAI & M-S & - & - & - & No & NR & S & - \\
\rowcolor{lightgray!20}METRO~\cite{cui2021metro} & 2022 & VLDB & M-S & D-C & Spatial GNN & T & No & NR & D & - \\
SGP~\cite{cini2023scalable} & 2023 & AAAI & M-S & D-F & Spatial GNN & T-R & No & R & - & SP, PS \\
\rowcolor{lightgray!20}HiGP~\cite{cini2023graphbased} & 2023 & arXiv & M-S & D-F & Spatial GNN & T-R & No & R & S & SP, PS \\
Jin et al.~\cite{jin2023powerful} & 2023 & arXiv & M-S, M-L & D-F & Spectral GNN & F-H & No & NR & S & - \\
\rowcolor{lightgray!20}CaST~\cite{xia2023deciphering} & 2023 & NeurIPS & M-S & D-F & Spectral GNN & T\&F-C & No & O & S & PC \\
\revision{GPT-ST~\cite{li2024gpt}} & \revision{2023} & \revision{NeurIPS} & \revision{M-S} & \revision{D-F} & \revision{Spatial GNN} & \revision{T-C} & \revision{No} & \revision{NR} & \revision{S} & \revision{-} \\
\bottomrule
\end{tabular}
\end{adjustbox}
\vspace{-0.3cm}
\end{table*}

Time series forecasting aims to predict future time series values based on historical observations. While deep learning models have demonstrated considerable success in forecasting time series by capturing nonlinear temporal and spatial patterns more effectively than the linear counterpart~\cite{jin2022multivariate}, many of these approaches, such as LSTNet~\cite{lai2018modeling} and TPA-LSTM~\cite{shih2019temporal}, overlook and implicitly model the rich underlying dynamic spatial correlations between time series. Recently, graph neural network (GNN)-based methods have shown great potential in explicitly and effectively modeling spatial and temporal dependencies in multivariate time series data, leading to enhanced forecasting performance. 

GNN-based forecasting models can be categorized and examined from multiple perspectives. In terms of forecasting tasks, while many models focus on \textit{multi-step forecasting} (i.e., predicting multiple consecutive steps ahead based on historical observations), a minority also discuss \textit{single-step forecasting} (i.e., predicting the next or one arbitrary step ahead). From a methodological standpoint, these models can be dissected from three aspects: (1) modeling spatial (i.e., inter-variable) dependencies, (2) modeling inter-temporal dependencies, and (3) the fusion of spatial and temporal modules for time series forecasting. A summary of representative works is given in \shortautoref{tab:forecasting summary}.

\subsection{Modeling Inter-variable Dependencies}
Spatial dependencies, or inter-variable relationships, play a pivotal role in affecting a model's forecasting capability~\cite{jin2023powerful}. When presented with time series data and corresponding graph structures that delineate the strength of interconnections between time series, current studies typically employ (1) \textit{spectral GNNs}, (2) \textit{spatial GNNs}, or (3) a \textit{hybrid} of both to model these spatial dependencies. At a high level, these methods all draw upon the principles of graph signal processing (as detailed in \shortautoref{def: graph neural network} and subsequent discussion). Considering input variables $\mathbf{X}_{t}$ and $\mathbf{A}_{t}$ at a given time $t$, the goal here is to devise an effective GNN-based model $\textsc{Spatial}(\cdot)$ to adeptly capture salient patterns from different time series. This can be expressed as $\Hat{\mathbf{X}}_{t} = \textsc{Spatial}(\mathbf{X}_{t}, \mathbf{A}_{t})$, where $\Hat{\mathbf{X}}_{t}$ collects all time series representations at time $t$ with spatial dependencies embedded.

\textbf{Spectral GNN-based Approaches.} 
Early GNN-based forecasting models predominantly utilized ChebConv~\cite{defferrard2016convolutional} to approximate graph convolution with Chebyshev polynomials, thereby modeling inter-variable dependencies. For instance, STGCN~\cite{yu2018spatio} intersects temporal convolution~\cite{lea2017temporal} and ChebConv layers to capture both spatial and temporal patterns. StemGNN~\cite{cao2020spectral} further proposes spectral-temporal graph neural networks to extract rich time series patterns by leveraging ChebConv and frequency-domain convolution neural networks. Other relevant research has largely followed suit, employing ChebConv to model spatial time series dependencies, while introducing novel modifications. These include attention mechanisms~\cite{guo2019attention, huang2020lsgcn}, multi-graph construction~\cite{geng2019spatiotemporal, yu2022regularized}, and combinations of the two~\cite{lan2022dstagnn}.

\textbf{Spatial GNN-based Approaches.} 
Inspired by the recent success of spatial GNNs~\cite{wu2020comprehensive}, another line of research has been modeling inter-variable dependencies using message passing~\cite{gilmer2017neural} or graph diffusion~\cite{gasteiger2019diffusion}. From the graph perspective, these methods are certain simplifications compared to those based on spectral GNNs, where strong local homophilies are emphasised~\cite{jin2023powerful, zheng2023finding}. Early methods such as DCRNN~\cite{lidiffusion} and Graph WaveNet~\cite{wu2019graph} incorporated graph diffusion layers into GRU~\cite{chung2014empirical} or temporal convolution to model time series data. In contrast, STGCN$(1^{st})$ (a second version of STGCN~\cite{yu2018spatio}) and ST-MetaNet~\cite{pan2019urban} modeled spatial dependencies with GCN~\cite{kipf2016semi} and GAT~\cite{velivckovic2017graph} to aggregate information from adjacent time series. To enhance learning capabilities, several improvements have been proposed. For example, STSGCN~\cite{song2020spatial} proposed spatial-temporal synchronous graph convolution, extending GCN to model spatial and temporal dependencies on localized spatial-temporal graphs. MTGODE~\cite{jin2022multivariate} and TPGNN~\cite{liu2022multivariate} proposed continuous graph propagation and graph propagation based on temporal polynomial coefficients. \revision{Additionally, recent approaches based on graph transformer~\cite{ying2021transformers} or hypergraphs~\cite{feng2019hypergraph}, such as \cite{feng2022adaptive} and \cite{li2024gpt}, can capture longer-range spatial dependencies due to their global receptive field, making them a separate branch of enhanced methods.}

\textbf{Hybrid Approaches.} 
Some hybrid models also exist, integrating both spectral and spatial GNNs. For instance, SLCNN~\cite{zhang2020spatio} employs ChebConv and localized message passing as global and local convolutions to capture spatial relations at multiple granularities. Conversely, Auto-STGNN~\cite{wang2023auto} integrates neural architecture search to identify high-performance GNN-based forecasting models.

More details about modeling inter-variable dependencies in GNNs for time series forecasting are in \emph{Appendix A.1}.

\subsection{Modeling Inter-temporal Dependencies}
The modeling of temporal dependencies within time series represents another important element in various GNN-based forecasting methods. These dependencies (i.e., temporal patterns) are capable of being modeled in the \textit{time} or/and \textit{frequency} domains. A summary of representative methods, along with their temporal module classifications, is presented in \shortautoref{tab:forecasting summary}. Given a univariate time series $\mathbf{X}_n$ with length $T$, the primary goal here is to learn an effective temporal model, referred to as $\textsc{Temporal}(\cdot)$. This model is expected to accurately capture the dependencies between data points within $\mathbf{X}_n$, such that $\Hat{\mathbf{X}}_n = \textsc{Temporal}(\mathbf{X}_n)$, where $\Hat{\mathbf{X}}_n$ symbolizes the representation of time series $\mathbf{X}_n$. In the construction of $\textsc{Temporal}(\cdot)$, both the time and frequency domains can be exploited within \textit{convolutional} and \textit{attentive} mechanisms. \textit{Recurrent} models can also be employed for modeling in the time domain specifically. Additionally, \textit{hybrid} models exist in both domains, integrating different methodologies such as attention and convolution neural networks.

\textbf{Recurrent Models.}  
Several early methodologies rely on recurrent models for understanding inter-temporal dependencies in the time domain. For instance, DCRNN~\cite{lidiffusion} integrates graph diffusion with gated recurrent units (GRU)~\cite{chung2014empirical} to model the spatial-temporal dependencies in traffic data. On a different note, AGCRN~\cite{bai2020adaptive} merges GRU with a factorized variant of GCN~\cite{kipf2016semi} and a graph structure learning module. More recent studies, such as GTS~\cite{shang2021discrete} and RGSL~\cite{yu2022regularized}, share similar designs but primarily emphasize different graph structure learning mechanisms.

\textbf{Convolution Models.} 
Convolutional neural networks (CNNs), on the other hand, provide a more efficient perspective for modeling inter-temporal dependencies, with the bulk of existing studies in the time domain. An instance of this is STGCN~\cite{yu2018spatio}, which introduces temporal gated convolution that integrates 1-D convolution with gated linear units (GLU) to facilitate tractable model training. Other representative examples include MTGNN\cite{wu2020connecting} and MTGODE\cite{jin2022multivariate}. An alternative strand of methodologies, including StemGNN\cite{cao2020spectral} and TGC\cite{jin2023powerful}, focuses on modeling temporal clues in the frequency domain.

\textbf{Attention Models.} 
Recently, a growing number of methodologies is turning towards attention mechanisms, e.g., the self-attention in transformer~\cite{vaswani2017attention}, to embed temporal correlations. For instance, GMAN~\cite{zheng2020gman} attentively aggregates historical information by considering both spatial and temporal features. ST-GRAT~\cite{park2020st} mirrors the transformer's architecture to embed historical observations in conjunction with its proposed spatial attention mechanism. Recent strides such as STEP~\cite{shao2022pre} similarly employ transformer layers to model the temporal dependencies within each univariate time series.

\textbf{Hybrid Models.} 
Hybrid models also find application in modeling inter-temporal dependencies. For example, ASTGCN~\cite{guo2019attention} and DSTAGNN~\cite{lan2022dstagnn} concurrently employ temporal attention and convolution in learning temporal correlations. STGNN*~\cite{wang2020traffic} amalgamates both GRU and transformer to capture local and global temporal dependencies. In the frequency domain, the nonlinear variant of TGC~\cite{jin2023powerful} captures temporal relations through the combination of spectral attention and convolution models.

Complete discussion of modeling inter-temporal dependencies is in \emph{Appendix A.2}.

\subsection{Forecasting Architectural Fusion}
Given the spatial and temporal modules discussed, denoted as $\textsc{Spatial}(\cdot)$ and $\textsc{Temporal}(\cdot)$, four categories of neural architectural fusion have been identified as effective means to capture spatial-temporal dependencies within time series data: (1) \textit{discrete factorized}, (2) \textit{discrete coupled}, (3) \textit{continuous factorized}, and (4) \textit{continuous coupled}. In discrete factorized models, spatial and temporal dependencies are usually learned and processed independently. Discrete coupled models, on the other hand, explicitly or implicitly incorporate spatial and temporal modules into a singular process when modeling spatial-temporal dependencies. Different from discrete models, some methods abstract the underlying modeling processes with neural differential equations, which we categorize as continuous models. Specifically, continuous factorized models involve distinct processes, either partially or entirely continuous (e.g., \cite{fang2021spatial}), to model spatial and temporal dependencies. In contrast, continuous coupled models employ a single continuous process to accomplish this task, such as \cite{jin2022multivariate} and \cite{choi2022graph}.

\textbf{Discrete Architectures.} Numerous existing GNN-based time series forecasting methods are models processing discrete data. For instance, factorized approaches like STGCN~\cite{yu2018spatio} employ a sandwich structure of graph and temporal gated convolution layers as its fundamental building block. Subsequent works, such as DGCNN~\cite{diao2019dynamic} and HGCN~\cite{guo2021hierarchical}, retain this model architecture while introducing enhancements such as dynamic graph structure estimation and hierarchical graph generation. In the realm of discrete coupled models, early works such as DCRNN~\cite{lidiffusion} and Cirstea et al.\cite{cirstea2019graph} straightforwardly incorporate graph diffusion or attention models into recurrent units. Subsequent works, such as MRA-BGCN~\cite{chen2020multi} and RGSL~\cite{yu2022regularized}, are based on similar concepts but with varying implementations. There are also some studies integrating spatial and temporal convolution or attention operations into a single module. An example is GMAN~\cite{zheng2020gman}, which proposes a building block that integrates the spatial and temporal attention mechanisms in a gated manner.

\textbf{Continuous Architectures.} To date, only a handful of methods fall into this category. For factorized methods, STGODE~\cite{fang2021spatial} proposes to depict the graph propagation as a continuous process with a neural ordinary differential equation (NODE)~\cite{chen2018neural}. This approach allows for the effective characterization of long-range spatial-temporal dependencies in conjunction with dilated convolutions along the time axis. For coupled methods, MTGODE~\cite{jin2022multivariate} generalizes both spatial and temporal modeling processes found in most related works into a single unified process that integrates two NODEs. STG-NCDE~\cite{choi2022graph} shares a similar idea but operates under the framework of neural controlled differential equations (NCDEs)~\cite{kidger2020neural}. 

Refer to \emph{Appendix A.3} for more details about the architecture fusion in GNNs for time series forecasting.

\section{GNNs for Time Series Anomaly Detection}\label{sec:anomaly detection}

\begin{table*}[thp]
\caption{Summary of representative graph neural networks for time series anomaly detection. \textit{Strategy notation}: ``CL'', ``FC'', ``RC'', and ``RL'' indicate ``Class'', ``Forecast'', ``Reconstruction'', and ``Relational Discrepancies'', respectively. The remaining notations are shared with \autoref{tab:forecasting summary}.}
\label{tab:anomaly summary}
\centering
\begin{adjustbox}{width=2\columnwidth,center}
\begin{tabular}{lccccccccc}
\toprule
\thead{\textbf{Approach}} & \thead{\textbf{Year}} & \thead{\textbf{Venue}} & \thead{\textbf{Strategy}} & \thead{\textbf{Spatial} \\ \textbf{Module}} & \thead{\textbf{Temporal} \\ \textbf{Module}} & \thead{\textbf{Missing} \\ \textbf{Values}} & \thead{\textbf{Input} \\ \textbf{Graph}} & \thead{\textbf{Learned} \\ \textbf{Relations}} & \thead{\textbf{Graph} \\ \textbf{Heuristics}} \\ \midrule
\rowcolor{lightgray!20}CCM-CDT \cite{grattarola2019change} &2019& IEEE TNNLS&RC&Spatial GNN&T-R&No& R & - & PC, FD \\
MTAD-GAT \cite{zhao2020multivariate} &2020& IEEE ICDM&FC+RC&Spatial GNN&T-A& No & NR&-&-\\ 
\rowcolor{lightgray!20}GDN\cite{deng2021graph}&2021&AAAI&FC&Spatial GNN&-&No&NR&S&-\\
GTA\cite{chen2021learning}&2021&IEEE IoT&FC&Spatial GNN&T-H&No&NR&S&-\\
\rowcolor{lightgray!20}EvoNet\cite{hu2021time} & 2021 & WSDM & CL & Spatial GNN & T-R & No & R&-&PS\\
Event2Graph\cite{wu2021event2graph} & 2021 & arXiv & RL & Spatial GNN & T-A & No & R&-&PS\\
\rowcolor{lightgray!20}GANF\cite{dai2022graphaugmented} & 2022 & ICLR & RC+RL & Spatial GNN & T-R & No &NR&S&- \\
Grelen\cite{zhang2022grelen}&2022 & IJCAI &RC+RL&Spatial GNN&T-H&No&NR&D&-\\ 
\rowcolor{lightgray!20}VGCRN\cite{chen2022deep} & 2022 & ICML & FC+RC&Spatial GNN&T-R&No&NR&S&- \\ 
FuSAGNet\cite{han2022learning} & 2022 & KDD & FC+RC&Spatial GNN & T-R&No&NR&S&-\\ 
\rowcolor{lightgray!20}GTAD\cite{guan2022gtad} & 2022 & Entropy & FC+RC & Spatial GNN & T-C & No & NR&-&-\\
HgAD\cite{srinivas2022hypergraph}&2022 & IEEE BigData & FC&Spatial GNN&-&No&NR&S&\\
\rowcolor{lightgray!20}HAD-MDGAT\cite{zhou2022hybrid}& 2022 & IEEE Access&FC+RC&Spatial GNN&T-A&No&NR&-&-\\
STGAN\cite{zhou2022hybrid}&2022&IEEE TNNLS&RC&Spatial GNN&T-R&No&R&-&SP\\
\rowcolor{lightgray!20}GIF \cite{zambon2022graph} &2022& IEEE IJCNN&RC&Spatial GNN&-&No& R & - & SP, PC, FD \\ 
DyGraphAD\cite{chen2023DyGraphAD}&2023&arXiv&FC+RL&Spatial GNN&T-C&No&R&-&PS\\
\rowcolor{lightgray!20}GraphSAD\cite{chen2023time}&2023&arXiv&CL&Spatial GNN&T-C&No&R&-&PS, PC \\
CST-GL\cite{zheng2023correlation}&2023&IEEE TNNLS&FC&Spatial GNN&T-C&No&NR&S&-\\ 
 \bottomrule
\end{tabular}
\end{adjustbox}
\vspace{-0.3cm}
\end{table*}

Time series anomaly detection identifies data observations that deviate from the nominal data-generating process~\cite{hawkins1980identification}. Anomalies, defined as such deviations, contrast with normal data. Terms like novelty and outlier are often used interchangeably with anomaly~\cite{pimentel2014review}. Deviations can be single observations (points) or a series of observations (subsequences)~\cite{darban2022deep}. However, unlike normal time series data, anomalies are hard to characterize because they are rare, making them difficult to collect and label, and it is usually impossible to establish all potential anomalies, limiting supervised learning. Therefore, unsupervised detection techniques are widely explored as practical solutions.

Traditionally, distance-based~\cite{keogh2005hot} and distributional techniques~\cite{ting2021isolation} have been widely used for detecting irregularities in time series data. Recently, deep learning has driven significant advancements, particularly with recurrent models employing reconstruction~\cite{park2018multimodal} and forecasting~\cite{hundman2018detecting} strategies. These models use forecast and reconstruction errors to measure discrepancies between expected and actual signals, as a model trained on normal data is more likely to fail with anomalous data. However, recurrent models~\cite{su2019robust} often lack explicit modeling of pairwise interdependence among variables, limiting their effectiveness in detecting complex anomalies~\cite{zhao2020multivariate, xu2021anomaly}. Recently, GNNs have shown promise in capturing temporal and spatial dependencies among variables, addressing this gap~\cite{deng2021graph,han2022learning}.

\subsection{General \revision{Approach to} Anomaly Detection}
Treating anomaly detection as an unsupervised task relies on models to learn a general concept of what normality is for a given dataset~\cite{jin2021anemone, zheng2021generative}.
To achieve this, deep learning architectures deploy a bifurcated modular framework, constituted by a backbone module and a scoring module~\cite{garg2021evaluation}. Firstly, a backbone model, $\textsc{Backbone}(\cdot)$, is trained to fit given training data, assumed to be nominal, or to contain very few anomalies. Then, a scoring module, $\textsc{Scorer} (\mathbf{X}, \mathbf{\hat{X}})$, produces a score used to identify the presence of anomalies by comparing the output $\mathbf{\hat{X}}=\textsc{Backbone}(\mathbf{X})$ of the backbone module with the observed time series data $\mathbf{X}$. The score is intended as a measure of the discrepancy between the expected signals under normal and anomalous circumstances. Furthermore, it is also important for a model to diagnose anomaly events by pinpointing the responsible variables. Consequently, a scoring function typically computes the discrepancy for each individual channel first, before consolidating these discrepancies across all channels into a single anomaly value.

To provide a simple illustration of the entire process, the backbone can be a GNN forecaster that makes a one-step-ahead forecast for the scorer. The scorer then computes the anomaly score as the sum of the absolute forecast error for each channel variable, represented as $\sum_i^N |x_t^i - \hat{x}_t^i|$ across $N$ channel variables. Since the final score is computed based on the summation of channel errors, an operator can determine the root cause variables by computing the contribution of each variable to the summed error.

Advancements in the anomaly detection and diagnosis field have led to the proposal of more comprehensive backbone and scoring modules~\cite{garg2021evaluation,darban2022deep}, primarily driven by the adoption of GNN methodologies~\cite{zhao2020multivariate,deng2021graph,han2022learning}.

\subsection{Discrepancy Frameworks for Anomaly Detection}
Most of anomaly detection models follow the same backbone-scorer architecture. However, the way the backbone module is trained to learn data structure from nominal data and the implementation of the scoring module differentiate these methods into three categories: (1) \emph{reconstruction}, (2) \emph{forecast}, and (3) \emph{relational discrepancy} frameworks. 

\textbf{Reconstruction Discrepancy.}
Reconstruction discrepancy frameworks operate on the assumption that reconstruction error is low during normal periods and high during anomalies. They are designed to replicate their inputs as outputs, similar to autoencoders~\cite{goodfellow2016deep}. The backbone is expected to effectively model and reconstruct the training data distribution but not out-of-sample data. To achieve this, these frameworks often include constraints and regularization, such as enforcing a low-dimensional embedded code~\cite{ranzato2007sparse} or using variational objectives~\cite{kingma2019introduction}. Once the data structure is learned, the model should approximate the input well during normal periods but struggle during anomalies. The $\textsc{Scorer}(\cdot)$ then computes a discrepancy score from the reconstructed outputs to identify anomalous events.
Although deep reconstruction models generally follow these principles for detecting anomalies, a key distinction between GNNs and other architectural types rests in the backbone reconstructor, $\textsc{Backbone}(\cdot)$, which is characterized by its STGNN implementation. For example, MTAD-GAT~\cite{zhao2020multivariate} and LSTM-VAE~\cite{park2018multimodal} both use a variational objective~\cite{kingma2019introduction}, but MTAD-GAT employs a graph attention network as a spatial-temporal encoder to learn inter-variable and inter-temporal dependencies. Other similar works include VGCRN~\cite{chen2022deep} and FuSAGNet~\cite{han2022learning}. Another research direction in this category of methods focused on graph-level embeddings to represent the input graph data as vectors to enable the application of well-established and sophisticated detection methods designed for multivariate time series~\cite{zambon2018concept,zambon2019change}.

\textbf{Forecast Discrepancy.}
Forecast discrepancy frameworks rely on the assumption that forecast error should be low during normal periods, but high during anomalous periods. Here, the backbone module is substituted with a GNN forecaster that is trained to predict a one-step-ahead forecast. During deployment, the forecaster makes a one-step-ahead prediction, and the forecasts are given to the scorer. The scorer compares the forecasts against the real observed signals to compute discrepancies such as absolute error~\cite{deng2021graph} or mean-squared error~\cite{chen2021learning}. Importantly, it is generally assumed that a forecasting-based model will exhibit erratic behavior during anomaly periods when the input data deviates from the normal patterns, resulting in a significant forecasting discrepancy. GDN~\cite{deng2021graph} is a representative work, consisting of a graph structure module that learns the underlying topology and a graph attention network that encodes input series representations for one-step-ahead forecasts. Then, its scorer computes the forecast discrepancy as the maximum absolute forecast error among the channel variables to indicate whether an anomaly event has occurred. Other similar approaches include GTA~\cite{chen2021learning} and CST-GL~\cite{zheng2023correlation}.

\revision{While forecast discrepancy frameworks can be similar to reconstruction discrepancy frameworks, they rely on fundamentally different principles and implementation strategies. Reconstruction discrepancy frameworks project current input onto a latent space, attempting to reconstruct it and identifying anomalies when reconstruction fails. This method uses current input data during both training and inference, focusing on the model's ability to reconstruct training data but not out-of-sample data. Conversely, forecast discrepancy frameworks use historical data to predict current values and identify anomalies by comparing predictions with actual observations. Reconstruction focuses on replicating current inputs, while forecasting emphasizes predicting future data points and detecting anomalies through prediction errors. Advanced forecast discrepancy-based  methods, such as GST-Pro~\cite{zheng2024graph}, can even predict anomalies without using actual observations, allowing anomaly prediction at future timestamps.}

\textbf{Relational Discrepancy.}
Relational discrepancy frameworks rely on the assumption that the relationship between variables should exhibit significant shifts from normal to anomalous periods. The logical evolution of using STGNN involves leveraging its capability to learn graph structures for both anomaly detection and diagnosis. In this context, the backbone serves as a graph learning module that constructs the hidden evolving relationship between variables. The scorer, on the other hand, is a function that evaluates changes in these relationships and assigns an anomaly or discrepancy score accordingly. 
GReLeN~\cite{zhang2022grelen} pioneered the use of learned dynamic graphs to detect anomalies through relational discrepancies. Its reconstruction module dynamically constructs graph structures based on input time series data at each time point. These structures are then used by a scorer to compute changes in the in-degree and out-degree values of channel nodes. In contrast, DyGraphAD employs a forecasting approach~\cite{chen2023DyGraphAD}. It divides a multivariate series into subsequences, converts these into dynamically evolving graphs, and trains the network to predict one-step-ahead graph structures. The scorer in DyGraphAD computes the forecast error in the graph structure as the relational discrepancy for anomaly detection.

\textbf{Hybrid and Other Discrepancies.}
Different discrepancy-based frameworks offer unique advantages for detecting various types of anomalies. As demonstrated in GDN~\cite{deng2021graph}, the relational discrepancy framework can uncover spatial anomalies that are concealed within the relational patterns between different channels. In contrast, forecast discrepancy frameworks excel at identifying temporal anomalies like sudden spikes or seasonal inconsistencies. A comprehensive solution would leverage the strengths of STGNNs by combining multiple discrepancy measures for anomaly detection. For example, MTAD-GAT~\cite{zhao2020multivariate} uses both reconstruction and forecast discrepancies, while DyGraphAD~\cite{chen2023DyGraphAD} combines forecast and relational discrepancies. Additionally, incorporating prior knowledge about anomalous behaviors can enhance detection. For instance, GraphSAD~\cite{chen2023time} creates pseudo labels for six distinct anomaly types on training data, transforming unsupervised anomaly detection into a standard classification task, with class discrepancy as the anomaly indicator.

Our detailed discussion about the GNNs for time series anomaly detection can be found in \emph{Appendix B}.
\section{GNNs for Time Series Classification}\label{sec:classification}

\begin{table*}[htbp]
\caption{Summary of graph neural networks for time series classification. \textit{Task notation}: ``U'' and ``M'' refer to univariate and multivariate time series classification tasks. \textit{Conversion} represents the transformation of a time series classification task into a graph-level task as either graph or node classification task, \revision{represented as ``Series-as-Graph'' and ``Series-as-Node'', respectively.} The remaining notations are shared with \autoref{tab:forecasting summary}.}
\label{tab:analysis summary}
\centering
\begin{adjustbox}{width=2\columnwidth,center}
\begin{tabular}{lcccccccccc}
\toprule
\thead{\textbf{Approach}} & \thead{\textbf{Year}} & \thead{\textbf{Venue}} & \thead{\textbf{Task}} & \thead{\textbf{Conversion}} & \thead{\textbf{Spatial} \\ \textbf{Module}} & \thead{\textbf{Temporal} \\ \textbf{Module}} & \thead{\textbf{Missing} \\ \textbf{Values}} & \thead{\textbf{Input} \\ \textbf{Graph}} & \thead{\textbf{Learned} \\ \textbf{Relations}} & \thead{\textbf{Graph} \\ \textbf{Heuristics}} \\ \midrule
\rowcolor{lightgray!20}MTPool\cite{duan2022multivariate} & 2021 & NN & M & - & Spatial GNN & T-C  & No & NR&S&-\\
Time2Graph+\cite{cheng2021time2graph+} & 2021 & IEEE TKDE & U & \revision{Series-as-Graph} & Spatial GNN & - & No & R&-&PS\\
\rowcolor{lightgray!20}RainDrop\cite{zhang2022graphguided} & 2022 & ICLR & M & - & Spatial GNN & T-A & Yes & NR&S&-\\
SimTSC\cite{zha2022towards} & 2022 & SDM & U+M & \revision{Series-as-Node} & Spatial GNN & T-C & No & R&-&PS\\
\rowcolor{lightgray!20}LB-SimTSC\cite{xi2023lb} & 2023 & arXiv & U+M & \revision{Series-as-Node} & Spatial GNN & T-C & No & R&-&PS\\
TodyNet\cite{liu2023todynet} & 2023 & arXiv & M & - & Spatial GNN & T-C & No & NR&D&-\\
\rowcolor{lightgray!20}\revision{EC-GCN\cite{diao2023ec}} & \revision{2023} & \revision{Comput. Netw.} & \revision{U} & \revision{Series-as-Graph} & \revision{Spatial GNN} & \revision{T-C} & \revision{No} & \revision{R} & \revision{D} & \revision{PS} \\
\revision{MTS2Graph\cite{younis2024mts2graph}} & \revision{2024} & \revision{Pattern Recognit.} & \revision{M} & \revision{Series-as-Graph} & \revision{Spatial GNN} & \revision{T-C} & \revision{No} & \revision{NR} & \revision{-} &- \\
\bottomrule
\end{tabular}
\end{adjustbox}
\vspace{-0.3cm}
\end{table*}

Time series classification seeks to assign a categorical label to a given time series based on its underlying patterns or characteristics. As outlined in a recent survey~\cite{foumani2023deep}, early literature primarily focused on distance-based approaches~\cite{lines2015time,herrmann2021amercing} and ensembling methods~\cite{lines2018time,middlehurst2021hive}. However, despite their strong performance, both strategies struggle with scalability for high-dimensional or large datasets~\cite{dempster2020rocket,tan2022multirocket}. To address these limitations, researchers are exploring deep learning techniques to enhance the performance and scalability of time series classification. A comprehensive discussion can be found in the latest survey by Foumani et al.~\cite{foumani2023deep}, though it does not cover the application of GNNs in this field. By transforming time series data into graph representations, one can leverage the powerful capabilities of GNNs to capture both local and global patterns. Furthermore, GNNs are capable of mapping the intricate relationships among different time series data samples within a particular dataset. In the following discussion, we provide a fresh GNN perspective on the univariate and multivariate time series classification problem.

\subsection{Univariate Time Series Classification}
Time series classification inherently differs from other time series analyses by focusing on discerning patterns that distinguish samples based on class labels, rather than capturing patterns within the data. Unlike forecasting future points or detecting real-time anomalies, it aims to identify divergent patterns \emph{across series}. We explore two novel graph-based approaches for univariate time series classification: \revision{\emph{Series-as-Graph} and \emph{Series-as-Node}.}

\textbf{\revision{Series-as-Graph.}} This approach transforms a univariate time series into a graph to identify unique patterns that enable accurate classification using a GNN. Firstly, each series is broken down into subsequences as nodes, and the nodes are connected with edges illustrating their relationships. Following this, a GNN is applied to make graph classification. This procedure is represented in the upper block of \shortautoref{fig:gnn4tsc}. \revision{Series-as-Graph} was first presented in Time2Graph~\cite{cheng2020time2graph}, which was later developed further with the incorporation of GNN, as Time2Graph+~\cite{cheng2021time2graph+}. The Time2Graph+ modeling process can be described as a two-step process: first, a time series is transformed into a shapelet graph, and second, a GNN is utilized to model the relations between shapelets along with a graph pooling operation to derive the global representation of the time series. This representation is then fed into a classifier to assign class labels to the time series. \revision{Similarly, EC-GCN~\cite{diao2023ec} constructs graphs from encrypted traffic for classification. The Series-as-Graph approach can also be extended to multivariate time-series classification task~\cite{younis2024mts2graph}.} While we use \revision{Series-as-Graph} to describe the formulation of a time series classification task as a graph classification task, this should not be confused with Series2Graph~\cite{boniol13series2graph}, which is for anomaly detection tasks.

\textbf{\revision{Series-as-Node.}} As capturing differentiating class patterns across different series data samples is important, leveraging relationships across the different series data samples in a given dataset can be beneficial for classifying a time series. To achieve this, one can take the \revision{Series-as-Node} approach, where each series sample is seen as a separate node. These series nodes are connected with edges that represent the relationships between them, creating a large graph that provides a complete view of the entire dataset.
This procedure is depicted in the lower block of \shortautoref{fig:gnn4tsc}, which essentially formulates a time series classification task as a node classification task. \revision{Series-as-Node} was originally presented in SimTSC~\cite{zha2022towards}. In this work, series nodes are connected using edges, which are defined by their pairwise DTW distance, to construct a graph. LB-SimTSC~\cite{xi2023lb} further extends on SimTSC to improve the DTW preprocessing efficiency by employing the widely-used lower bound for DTW~\cite{keogh2005exact}. This allows for a time complexity of $O(L)$ rather than $O(L^2)$, dramatically reducing computation time.

More details about GNN-based univariate time series classification are in \emph{Appendix C.1}.

\subsection{Multivariate Time Series Classification}
In essence, multivariate time series classification maintains fundamental similarities with its univariate counterpart. However, it introduces an additional layer of complexity: the necessity to capture intricate inter-variable dependencies. For example, given the interconnectedness of brain regions, analyzing a single node in isolation may not fully capture the comprehensive neural dynamics~\cite{tang2022selfsupervised}. By modeling inter-variable dependencies, we can understand the relationships between different nodes, thereby offering a more holistic view of brain activity. This facilitates the differentiation of intricate patterns that can classify patients with and without specific neurological conditions. Since the relationships between the variables, or inter-variable dependencies, can be naturally thought of as a network graph, GNNs are ideally suitable as illustrated before in \shortautoref{sec:forecasting}. The primary aim here is to effectively distill the complexity of high-dimensional series data into a more comprehensible, yet equally expressive, representation that enables differentiation of time series into their representative classes~\cite{duan2022multivariate,liu2023todynet, zhang2022graphguided}. More discussion is in \textit{Appendix C.2}.
\section{GNNs for Time Series Imputation}\label{sec:imputation}

\begin{table*}[htbp]
\caption{Summary of graph neural networks for time series imputation. \textit{Task notation}: ``Out-of-sample'', ``In-sample'', and ``Both'' refer to the types of imputation problems addressed by the approach. \textit{Type} represents the imputation method as either deterministic or probabilistic. \textit{Inductiveness} indicates if the method can generalize to unseen nodes. The remaining notations are shared with \autoref{tab:forecasting summary}.}
\label{tab: time series imputation}
\centering
\begin{adjustbox}{width=2\columnwidth,center}
\begin{tabular}{lcccccccccc}
\toprule
\thead{\textbf{Approach}} & \thead{\textbf{Year}} & \thead{\textbf{Venue}} & \thead{\textbf{Task}} & \thead{\textbf{Type}} & \thead{\textbf{Spatial} \\ \textbf{Module}} & \thead{\textbf{Temporal} \\ \textbf{Module}} & \thead{\textbf{Inductiveness}} & \thead{\textbf{Input} \\ \textbf{Graph}} & \thead{\textbf{Learned} \\ \textbf{Relations}} & \thead{\textbf{Graph} \\ \textbf{Heuristics}} \\ \midrule
\rowcolor{lightgray!20}IGNNK~\cite{wu2021inductive} & 2021 & AAAI & Out-of-sample & Deterministic & Spatial GNN & - & Yes & R & - & SP, PC \\
GACN~\cite{ye2021spatial} & 2021 & ICANN & In-sample & Deterministic & Spatial GNN & T-C & No & R & - & PC \\
\rowcolor{lightgray!20}SATCN~\cite{wu2021spatial} & 2021 & arXiv & Out-of-sample & Deterministic & Spatial GNN & T-C & Yes & R & - & SP \\
GRIN~\cite{cini2022filling} & 2022 & ICLR & Both & Deterministic & Spatial GNN & T-R & Yes & R & - & SP \\
\rowcolor{lightgray!20}SPIN~\cite{mariscalearning} & 2022 & NIPS & In-sample & Deterministic & Spatial GNN & T-A & No & R & - & SP \\
FUNS~\cite{roth2022forecasting} & 2022 & ICDMW & Out-of-sample & Deterministic & Spatial GNN & T-R & Yes & R & - & - \\
\rowcolor{lightgray!20}AGRN~\cite{chen2023adaptive} & 2022 & ICONIP & In-sample & Deterministic & Spatial GNN & T-R & No & NR & S & - \\
MATCN~\cite{wu2022multi} & 2022 & IEEE IoT-J & In-sample & Deterministic & Spatial GNN & T-A & No & R & - & - \\
\rowcolor{lightgray!20}PriSTI~\cite{liu2023pristi} & 2023 & arXiv & In-sample & Probabilistic & Spatial GNN & T-A & No & R & - & SP \\
DGCRIN~\cite{kong2023dynamic} & 2023 & KBS & In-sample & Deterministic & Spatial GNN & T-R & No & NR & D & - \\
\rowcolor{lightgray!20}GARNN~\cite{shen2023bidirectional} & 2023 & Neurocomputing & In-sample & Deterministic & Spatial GNN & T-R & No & R & - & PC \\ 
MDGCN~\cite{liang2022memory} & 2023 & Transp. Res. Part C & In-sample & Deterministic & Spatial GNN & T-R & No & R & - & SP, PS \\ 
\rowcolor{lightgray!20}\revision{INCREASE~\cite{zheng2023increase}} & \revision{2023} & \revision{WWW} & \revision{Out-of-sample} & \revision{Deterministic} & \revision{Spatial GNN} & \revision{T-R} & \revision{Yes} & \revision{R} & \revision{-} & \revision{SP, PC, FD} \\
\revision{Yun et al.~\cite{yun2023imputation}} & \revision{2023} & \revision{OpenReview} & \revision{In-sample} & \revision{Probabilistic} & \revision{Spatial GNN} & \revision{-} & \revision{No} & \revision{R} & \revision{-} & \revision{-} \\
\bottomrule
\end{tabular}
\end{adjustbox}
\vspace{-0.3cm}
\end{table*}

Time series imputation, a crucial task in numerous real-world applications, involves estimating missing values within one or more data point sequences. Traditional time series imputation approaches have relied on statistical methodologies, such as mean imputation, spline interpolation~\cite{moritz2017imputets}, and regression models~\cite{saad2020machine}. However, these methods often struggle to capture complex temporal dependencies and non-linear relationships within the data. While some deep neural network-based works, such as \cite{che2018recurrent} and \cite{miao2021generative}, have mitigated these limitations, they do not explicitly consider inter-variable dependencies. The recent emergence of GNNs offers new possibilities for time series imputation by better capturing intricate spatial and temporal dependencies. From a task perspective, GNN-based time series imputation can be broadly categorized into two types: \textit{in-sample imputation} and \textit{out-of-sample imputation}. The former involves filling in missing values within the given time series data, while the latter predicts missing values in disjoint sequences~\cite{cini2022filling}. From a methodological perspective, GNNs for time series imputation can be categorized into \textit{deterministic} and \textit{probabilistic} approaches. Deterministic imputation provides a single best estimate for missing values, while probabilistic imputation accounts for uncertainty by offering a distribution of possible values. In \shortautoref{tab: time series imputation}, we summarize most of the related works on GNN for time series imputation to date, offering a comprehensive overview of the field and its current state of development.

\subsection{In-sample Imputation}
The majority of existing GNN-based methods primarily focus on in-sample time series data imputation. For instance, GACN~\cite{ye2021spatial} proposes to model spatial-temporal dependencies in time series data by interleaving GAT~\cite{velivckovic2017graph} and temporal convolution layers in its encoder. It then imputes the missing data by combining GAT and temporal deconvolution layers that map latent states back to original feature spaces. GRIN~\cite{cini2022filling} introduces the graph recurrent imputation network, where each unidirectional module consists of one spatial-temporal encoder and two different imputation executors. The spatial-temporal encoder adopted in this work combines MPNN~\cite{gilmer2017neural} and GRU~\cite{chung2014empirical}. After generating the latent time series representations, the first-stage imputation fills missing values with one-step-ahead predicted values, which are then refined by a final one-layer MPNN before passing to the second-stage imputation for further processing. Similar works using bidirectional recurrent architectures include AGRN~\cite{chen2023adaptive}, DGCRIN~\cite{kong2023dynamic}, GARNN~\cite{shen2023bidirectional}, and MDGCN~\cite{liang2022memory}, where the main differences lie in intermediate processes. \revision{Recently, a few research studies have explored probabilistic in-sample time series imputation, such as PriSTI~\cite{liu2023pristi} and \cite{yun2023imputation}, where the imputation has been regarded as a generation task.} Refer to \emph{Appendix D.1} for more details.

\subsection{Out-of-sample Imputation}
To date, only a few GNN-based methods fall into this category. Among these works, IGNNK~\cite{wu2021inductive} proposes an inductive GNN kriging model to recover signals for unobserved time series, such as a new variable in a multivariate time series. In IGNNK, the training process involves masked subgraph sampling and signal reconstruction with the diffusion graph convolution network presented in \cite{li2018diffusion}. Another similar work is SATCN~\cite{wu2021spatial}, and the primary difference between these two works lies in the underlying GNN architectures. \revision{INCREASE~\cite{zheng2023increase}, on the other hand, further considers heterogeneous spatial and diverse temporal relations among the locations, lifting the performance of inductive spatio-temporal kriging.} Notably, GRIN~\cite{cini2022filling} can handle both in-sample and out-of-sample imputations, as well as \cite{roth2022forecasting}.
More discussion is in \emph{Appendix D.2}.
\section{Practical Applications and Resources}\label{sec:application}

Graph neural networks have been applied to a broad range of disciplines related to time series analysis. We categorize the mainstream applications of GNN4TS into \revision{seven} areas: smart transportation, on-demand services, environment \& sustainable energy, internet-of-things, \revision{physical systems,} healthcare, and fraud detection. \revision{Three prominent areas in this section are discussed. A detailed expansion can be found in \emph{Appendix E}.}
\revision{In addition, we summarize the common benchmark datasets and the open-sourced implementation of representative models in \emph{Appendix F}.}
\\

\noindent\textbf{Smart Transportation.} The domain of transportation has been significantly transformed with the advent of GNNs, with typical applications spanning from traffic prediction to flight delay prediction. For example, by leveraging advanced algorithms and data analytics related to spatial-temporal GNNs, traffic conditions can be accurately predicted~\cite{tang2022domain, li2023traffic}, thereby facilitating efficient route planning and congestion management. Another important application is traffic data imputation, which is crucial for maintaining the integrity of traffic databases and ensuring the accuracy of traffic analysis and prediction models~\cite{wu2022multi, kong2023dynamic}. There is also related research on autonomous driving~\cite{tang2023trajectory} and flight delay prediction~\cite{cai2021deep, guo2020sgdan}. The use of advanced technologies like GNNs in these applications underscores their transformative impact on smart transportation, emphasizing its critical role in evolving transportation systems. \\

\noindent\textbf{Environment \& Sustainable Energy.} In this sector, GNNs have been instrumental in wind speed and power prediction, capturing the complex spatial-temporal dynamics of wind patterns to provide accurate predictions that aid in the efficient management of wind energy resources~\cite{wu2022promoting, he2022robust}. Similarly, in solar energy, GNNs have been used for solar irradiance and photovoltaic (PV) power prediction, modeling the intricate relationships between various factors influencing solar energy generation to provide accurate predictions~\cite{jiao2021graph, zhang2022optimal}. Furthermore, GNNs have been employed for air pollution prediction~\cite{oliveira2023spatiotemporal} and weather forecasting~\cite{keisler2022forecasting} that are crucial for various services in agriculture, energy, and transportation. \\

\noindent\textbf{\revision{Physical systems.}}
\revision{
Systems of interacting objects are found in numerous scientific fields, including the simulation of n-body systems \cite{battaglia2016interaction}, particle physics~\cite{shi2023towards}, modeling of human motion dynamics \cite{liu2023grapha}, and prediction of molecular dynamics \cite{wu2023equivariant}. In graph-based deep learning, the objects are represented as nodes of a graph and GNNs have proven effective in modeling their complex interactions.
Despite the challenges posed by physical constraints, promising performance has been achieved thanks to the incorporation of inductive biases from known physical laws \cite{sanchez2019hamiltonian, liu2023segno} and architectures that maintain the symmetries of the underlying system, such as equivariance to rotations and translations \cite{satorras2021equivariant, brandstetter2021geometric, wu2023equivariant}.} \\

\noindent\textbf{Other Applications.} The application of GNNs for time series analysis has also been extended to various other fields, such as finance~\cite{wang2022review,song2023towards}, on-demand services~\cite{geng2019spatiotemporal}, fraud detection~\cite{noorshams2020ties}, \revision{manufacturing}~\cite{yao2023novel}, and recommender systems~\cite{wu2019session}. As research in this area continues to evolve, it is anticipated that the application of GNN4TS will continue to expand, opening up new possibilities for data-driven decision making and system optimization.
\section{Future Directions}\label{sec:prospect}
The future of GNNs for time series analysis holds immense promise, driven by several key directions and challenges that need to be addressed to unlock their full potential. More detailed discussion is in \emph{Appendix G}. \\

\noindent\textbf{Pre-training, Transfer Learning, and Large Models.} Pre-training, transfer learning, and large models are emerging as potent strategies to bolster the performance of GNNs in time series analysis~\cite{shao2022pre, wang2023building}, especially when data is sparse or diverse. These techniques hinge on utilizing learned representations from one or more domains to enhance performance in other related domains~\cite{zhang2023selfsupervised}. Notable examples include Panagopoulos et al.'s meta-learning for COVID-19 prediction in data-limited settings~\cite{panagopoulos2021transfer}, and Shao et al.'s pre-training framework for spatial-temporal GNNs~\cite{shao2022pre}. 
The exploration of GNN pre-training and transferability for time series tasks is growing, especially with the advent of generative AI and large models capable of addressing diverse tasks~\cite{jin2023large}. Challenges include limited time series data for large-scale pre-training, ensuring wide coverage and transferability, and designing strategies to capture complex spatial-temporal dependencies~\cite{zhao2023survey}. \\

\noindent\textbf{Robustness.} Robustness of GNNs refers to their ability to handle data perturbations and distribution shifts, especially those engineered by adversaries~\cite{zhang2022trustworthy}. This is crucial for time series from dynamic systems, as operational failures can compromise the entire system's integrity~\cite{dong2023graph}. For example, inadequate handling of noise or data corruption in smart city applications can disrupt traffic management, and in healthcare, it can cause missed critical treatment periods. While GNNs perform well in many applications, enhancing their robustness and developing effective failure management strategies are essential. \\

\noindent\textbf{Privacy Enhancing.} As GNNs become integral to various sectors, privacy protection is increasingly important. GNNs' capability to learn and reconstruct relationships within complex systems necessitates safeguarding the privacy of both individual entities (nodes) and their relationships (edges) in time series data. The interpretability of GNNs, while useful for identifying vulnerabilities, can also expose sensitive information~\cite{xu2021explainability}. Therefore, maintaining robust privacy defenses while capitalizing on the benefits of GNNs for time series analysis requires a delicate balance, one that calls for constant vigilance and continual innovation. \\

\noindent\textbf{Scalability.} While GNNs are effective in analyzing complex time series data, their adaptation to vast time-dependent data volumes presents challenges like memory constraints during computations. Traditional GNNs use sampling strategies such as node-wise~\cite{hamilton2017inductive}, layer-wise~\cite{chen2018fastgcn}, and graph-wise~\cite{chiang2019cluster} to mitigate these issues, but preserving temporal dependencies is complex. Enhancing scalability for real-time GNN applications, especially on edge devices with limited computing power, is crucial. This intersection of scalability and efficient inference is a significant research area with potential for major advancements.
\section{Conclusions}\label{sec:conclusion}
This comprehensive survey bridges the knowledge gap in graph neural networks for time series analysis (GNN4TS) by reviewing recent advancements and offering a unified taxonomy to categorize existing works. It covers a wide range of analytical tasks, including forecasting, classification, anomaly detection, and imputation, providing a detailed understanding of current progress. We also delve into the intricacies of spatial and temporal dependencies modeling and overall model architecture, offering a fine-grained classification of individual studies. Highlighting the expanding applications of GNN4TS, we demonstrate its versatility and potential for future growth. This survey serves as a valuable resource for practitioners and experts, proposing future research directions to guide and inspire further work in GNN4TS.

\section*{Acknowledgments}
This research was partly funded by the CSIRO – National Science Foundation (US) AI Research Collaboration Program and Swiss National Science Foundation under grant 204061.
S. Pan was supported in part by the Australian Research Council (ARC) under grants FT210100097 and DP240101547.

\appendices

\section{GNNs for Time Series Forecasting}\label{appx:forecasting}

Time series forecasting aims to predict future time series values based on historical observations. In recent years, deep learning-based approaches have demonstrated considerable success in forecasting time series by capturing nonlinear temporal and spatial patterns more effectively than the linear counterpart~\cite{jin2022multivariate}. Techniques such as recurrent neural networks (RNNs), convolutional neural networks (CNNs), and attention-based neural networks have been employed. However, many of these approaches, such as LSTNet~\cite{lai2018modeling} and TPA-LSTM~\cite{shih2019temporal}, overlook and implicitly model the rich underlying dynamic spatial correlations between time series. Recently, graph neural network (GNN)-based methods have shown great potential in explicitly and effectively modeling spatial and temporal dependencies in multivariate time series data, leading to enhanced forecasting performance.

GNN-based forecasting models can be categorized and examined from multiple perspectives. In terms of forecasting tasks, while many models focus on \textit{multi-step forecasting} (i.e., predicting multiple consecutive steps ahead based on historical observations), a minority also discuss \textit{single-step forecasting} (i.e., predicting the next or one arbitrary step ahead). From a methodological standpoint, these models can be dissected from three aspects: (1) modeling spatial (i.e., inter-variable) dependencies, (2) modeling inter-temporal dependencies, and (3) the fusion of spatial and temporal modules for time series forecasting. A summary of representative works is given in Tab. 2.

\subsection{Modeling Inter-variable Dependencies}
Spatial dependencies, or inter-variable relationships, play a pivotal role in affecting a model's forecasting capability~\cite{jin2023powerful}. When presented with time series data and corresponding graph structures that delineate the strength of interconnections between time series, current studies typically employ (1) \textit{spectral GNNs}, (2) \textit{spatial GNNs}, or (3) a \textit{hybrid} of both to model these spatial dependencies. At a high level, these methods all draw upon the principles of graph signal processing (as detailed in Def. 5 and subsequent discussion). Considering input variables $\mathbf{X}_{t}$ and $\mathbf{A}_{t}$ at a given time $t$, the goal here is to devise an effective GNN-based model, termed $\textsc{Spatial}(\cdot)$, to adeptly capture salient patterns from different time series at time $t$. This can be expressed as $\Hat{\mathbf{X}}_{t} = \textsc{Spatial}(\mathbf{X}_{t}, \mathbf{A}_{t})$, where $\Hat{\mathbf{X}}_{t}$ collects all time series representations at time $t$ with spatial dependencies embedded.

\textbf{Spectral GNN-based Approaches.} Early GNN-based forecasting models predominantly utilized ChebConv~\cite{defferrard2016convolutional} to approximate graph convolution with Chebyshev polynomials, thereby modeling inter-variable dependencies. For instance, STGCN~\cite{yu2018spatio} intersects temporal convolution~\cite{lea2017temporal} and ChebConv layers to capture both spatial and temporal patterns. StemGNN~\cite{cao2020spectral} further proposes spectral-temporal graph neural networks to extract rich time series patterns by leveraging ChebConv and frequency-domain convolution neural networks. Other relevant research has largely followed suit, employing ChebConv to model spatial time series dependencies, while introducing innovative modifications. These include attention mechanisms~\cite{guo2019attention, huang2020lsgcn}, multi-graph construction~\cite{geng2019spatiotemporal, yu2022regularized}, and combinations of the two~\cite{lan2022dstagnn}. Recently, building upon StemGNN, Jin et al.~\cite{jin2023powerful} have theoretically demonstrated the benefits of using spectral GNNs to model different signed time series relations, such as strongly positive and negative correlated variables within a multivariate time series. They also observed that any orthonormal family of polynomials could achieve comparable expressive power for such tasks, albeit with varying convergence rates and empirical performances.

\textbf{Spatial GNN-based Approaches.} Inspired by the recent success of spatial GNNs~\cite{wu2020comprehensive}, another line of research has been modeling inter-variable dependencies using message passing~\cite{gilmer2017neural} or graph diffusion~\cite{atwood2016diffusion, gasteiger2019diffusion}. 
From the graph perspective, these methods are certain simplifications compared to those based on spectral GNNs, where strong local homophilies are emphasised~\cite{jin2023powerful, zheng2023finding}. Early methods such as DCRNN~\cite{lidiffusion} and Graph WaveNet~\cite{wu2019graph} incorporated graph diffusion layers into GRU~\cite{chung2014empirical} or temporal convolution to model time series data. Subsequent works including GTS~\cite{shang2021discrete} and ST-GDN~\cite{zhang2021traffic} also applied graph diffusion. 
In contrast, STGCN$(1^{st})$ (a second version of STGCN~\cite{yu2018spatio}) and ST-MetaNet~\cite{pan2019urban} modelled spatial dependencies with GCN~\cite{kipf2016semi} and GAT~\cite{velivckovic2017graph} to aggregate information from adjacent time series. Related works, such as MRA-BGCN~\cite{chen2020multi}, STGNN*\cite{wang2020traffic}, GMAN\cite{zheng2020gman}, and AGCRN~\cite{bai2020adaptive}, proposed variants to model inter-variable relations based on message passing. 
To enhance learning capabilities, STSGCN~\cite{song2020spatial} proposed spatial-temporal synchronous graph convolution, extending GCN to model spatial and temporal dependencies on localized spatial-temporal graphs. STFGNN~\cite{li2021spatial} constructed spatial-temporal fusion graphs based on dynamic time wrapping (DTW) before applying graph and temporal convolutions. GSS models \cite{zambon2023graph} rely on state-space representations with attributed graphs as inputs, states and outputs, and where the states' topology is learned from data. Z-GCNETs~\cite{chen2021z} enhanced existing methods with salient time-conditioned topological information, specifically zigzag persistence images. METRO~\cite{cui2021metro} introduced multi-scale temporal graphs to characterize dynamic spatial and temporal interactions in time series data, together with the single-scale graph update and cross-scale graph fusion modules to unify the modeling of spatial-temporal dependencies. 
Another line of improvements incorporates graph propagation, allowing for the mixing of neighborhood information from different hops to learn high-order relations and substructures in the graph. Example are MTGNN~\cite{wu2020connecting}, SGP~\cite{cini2023scalable} and HiGP \cite{cini2023graphbased}. 
In particular, SGP exploits reservoir computing and multi-hop spatial processing to precompute spatio-temporal representations yielding effective, yet scalable predictive models, while HiGP exploits learned cluster assignments to regularize predictions and propagate messages hierarchically among nodes.
Follow-up works such as MTGODE~\cite{jin2022multivariate} and TPGNN~\cite{liu2022multivariate} proposed continuous graph propagation and graph propagation based on temporal polynomial coefficients. Other similar works include STGODE~\cite{fang2021spatial} and STG-NCDE~\cite{choi2022graph}.
Distinct from GAT-based methods, approaches based on graph transformer~\cite{ying2021transformers} can capture long-range spatial dependencies due to their global receptive field, making them a separate branch of enhanced methods. Examples include STAR~\cite{yu2020spatio}, ASTTN~\cite{feng2022adaptive}, and ASTTGN~\cite{huang2022adaptive}.
\revision{Alternatively, methods like GPT-ST~\cite{li2024gpt} leverage hypergraphs~\cite{feng2019hypergraph}, which can more effectively capture spatial dependencies across various scales from a global perspective.} At last, we mention NGAR~\cite{zambon2019autoregressive} and GEN~\cite{paassen2021graph} as works that can predict the graph topology alongside node-level signals.

Other potential modeling approaches, such as hyperbolic graph neural networks (HGNNs), exhibit great potential and have achieved state-of-the-art performance in graph learning~\cite{liu2019hyperbolic, chami2019hyperbolic, xu2024scalable}. One important characteristic of hyperbolic space is its exponential expansion, which can be viewed as a smoother version of trees and effectively abstracts hierarchical structures~\cite{krioukov2010hyperbolic}. Consequently, data with implicit hierarchical structures can be naturally represented in hyperbolic geometry, as opposed to being directly embedded in Euclidean space, which would introduce significant structural inductive biases and excessive distortion. This characteristic provides potential for modeling spatio-temporal graphs in hyperbolic space~\cite{mostafa2022hyperbolic}, and several works have proposed adapting hyperbolic GNNs for temporal graph learning~\cite{yang2021discrete, yang2022hyperbolic, bai2023hgwavenet, xu2024scalable,li2024dhgat}.

\textbf{Hybrid Approaches.} Some hybrid models also exist, integrating both spectral and spatial GNNs. For instance, SLCNN~\cite{zhang2020spatio} employs ChebConv~\cite{defferrard2016convolutional} and localized message passing as global and local convolutions to capture spatial relations at multiple granularities. Conversely, Auto-STGNN~\cite{wang2023auto} integrates neural architecture search to identify high-performance GNN-based forecasting models. In this approach, various GNN instantiations, such as ChebConv, GCN~\cite{kipf2016semi}, and STSGCN~\cite{song2020spatial}, can be simultaneously implemented in different spatial-temporal blocks.

\subsection{Modeling Inter-temporal Dependencies}
The modeling of temporal dependencies within time series represents another important element in various GNN-based forecasting methods. These dependencies (i.e., temporal patterns) are capable of being modeled in the \textit{time} or/and \textit{frequency} domains. A summary of representative methods, along with their temporal module classifications, is presented in Tab. 2. Given a univariate time series $\mathbf{X}_n$ with length $T$, the primary goal here is to learn an effective temporal model, referred to as $\textsc{Temporal}(\cdot)$. This model is expected to accurately capture the dependencies between data points within $\mathbf{X}_n$, such that $\Hat{\mathbf{X}}_n = \textsc{Temporal}(\mathbf{X}_n)$, where $\Hat{\mathbf{X}}_n$ symbolizes the representation of time series $\mathbf{X}_n$. In the construction of $\textsc{Temporal}(\cdot)$, both the time and frequency domains can be exploited within \textit{convolutional} and \textit{attentive} mechanisms. \textit{Recurrent} models can also be employed for modeling in the time domain specifically. Additionally, \textit{hybrid} models exist in both domains, integrating different methodologies such as attention and convolution neural networks.

\textbf{Recurrent Models.}  Several early methodologies rely on recurrent models for understanding inter-temporal dependencies in the time domain. For instance, DCRNN~\cite{lidiffusion} integrates graph diffusion with gated recurrent units (GRU)~\cite{chung2014empirical} to model the spatial-temporal dependencies in traffic forecasting. ST-MetaNet~\cite{pan2019urban} incorporates two types of GRU to encode historical observations and capture diverse temporal correlations that are tied to geographical information. Inspired by \cite{lidiffusion}, MRA-BGCN~\cite{chen2020multi} combines the proposed multi-range attention-based bicomponent graph convolution with GRU. This model is designed to better capture spatial-temporal relations by modeling both node and edge interaction patterns. On a different note, AGCRN~\cite{bai2020adaptive} merges GRU with a factorized variant of GCN~\cite{kipf2016semi} and a graph structure learning module. Some studies, such as GTS~\cite{shang2021discrete} and RGSL~\cite{yu2022regularized}, share similar designs but primarily emphasize different graph structure learning mechanisms. Recently, echo state networks (ESN)~\cite{lukosevicius2009reservoir} -- a type of RNN with sparse and randomized connectivity producing rich dynamics -- have been employed to design scalable models without compromising the performance~\cite{cini2023scalable,micheli2022discretetime}. Lastly, Graph Kalman Filters \cite{alippi2023graph} introduce a feedback loop in GSS models \cite{zambon2023graph} to improve the accuracy of state estimates and the predictions when actual system outputs are acquired.

\textbf{Convolution Models.} Convolutional neural networks (CNNs), on the other hand, provide a more efficient perspective for modeling inter-temporal dependencies, with the bulk of existing studies in the time domain. An instance of this is STGCN~\cite{yu2018spatio}, which introduces temporal gated convolution that integrates 1-D convolution with gated linear units (GLU) to facilitate tractable model training. Works that adopt a similar approach include DGCNN~\cite{diao2019dynamic}, SLCNN~\cite{zhang2020spatio}, and LSGCN~\cite{huang2020lsgcn}. Building on these foundations, Graph WaveNet~\cite{wu2019graph} incorporated dilated causal convolution, which notably expands the receptive field with only a minimal increase in model layers. STGODE~\cite{fang2021spatial} and STFGNN~\cite{li2021spatial} have produced similar designs in capturing temporal dependencies. MTGNN\cite{wu2020connecting} also uses these underpinning concepts, but it enhances temporal convolution by utilizing multiple kernel sizes. Further expanding on this, MTGODE\cite{jin2022multivariate} adopts a neural ordinary differential equation\cite{chen2018neural} to generalize this modeling process. There are some other studies, such as Z-GCNETs\cite{chen2021z}, that directly apply canonical convolution to capture temporal patterns within the time domain, albeit with other focuses. An alternative strand of methodologies, including StemGNN\cite{cao2020spectral} and TGC\cite{jin2023powerful}, focuses on modeling temporal clues in the frequency domain. StemGNN applies gated convolution to filter the frequency elements generated by the discrete Fourier transform of the input time series. In contrast, TGC convolves frequency components individually across various dimensions to craft more expressive temporal frequency-domain models.

\textbf{Attention Models.} Recently, a growing number of methodologies are turning towards attention mechanisms, such as the self-attention used in the transformer model\cite{vaswani2017attention}, to embed temporal correlations. For instance, GMAN~\cite{zheng2020gman} attentively aggregates historical information by considering both spatial and temporal features. ST-GRAT~\cite{park2020st} mirrors the transformer's architecture, employing multi-head self-attention layers within its encoder to embed historical observations in conjunction with its proposed spatial attention mechanism. STAR~\cite{yu2020spatio}, TPGNN~\cite{liu2022multivariate}, and STEP~\cite{shao2022pre} similarly employ transformer layers to model the temporal dependencies within each univariate time series. There are also variations on this approach, like the multi-scale self-attention network proposed by ST-GDN~\cite{zhang2021traffic}, aiming to model inter-temporal dependencies with higher precision. 

\textbf{Hybrid Models.} Hybrid models also find application in modeling inter-temporal dependencies. For example, ASTGCN\cite{guo2019attention}, HGCN\cite{guo2021hierarchical}, and DSTAGNN\cite{lan2022dstagnn} concurrently employ temporal attention and convolution in learning temporal correlations. STGNN*\cite{wang2020traffic} amalgamates both GRU and Transformer to capture local and global temporal dependencies. Auto-STGCN\cite{wang2023auto}, on the other hand, potentially facilitates more diverse combinations when searching for high-performance neural architectures. In the frequency domain, the nonlinear variant of TGC~\cite{jin2023powerful} is currently the only hybrid model proposing to capture temporal relations through the combination of spectral attention and convolution models.

\subsection{Forecasting Architectural Fusion}
Given the spatial and temporal modules discussed, denoted as $\textsc{Spatial}(\cdot)$ and $\textsc{Temporal}(\cdot)$, four categories of neural architectural fusion have been identified as effective means to capture spatial-temporal dependencies within time series data: (1) \textit{discrete factorized}, (2) \textit{discrete coupled}, (3) \textit{continuous factorized}, and (4) \textit{continuous coupled}. In discrete factorized models, spatial and temporal dependencies are usually learned and processed independently. This approach may involve stacking and interleaving spatial and temporal modules within a model building block\cite{yu2018spatio, wu2019graph, wu2020connecting}. Discrete coupled models, on the other hand, explicitly or implicitly incorporate spatial and temporal modules into a singular process when modeling spatial-temporal dependencies, such as in \cite{li2018diffusion}, \cite{chen2020multi}, and \cite{song2020spatial}. Different from discrete models, some methods abstract the underlying modeling processes with neural differential equations, which we categorize as continuous models. Specifically, continuous factorized models involve distinct processes, either partially or entirely continuous (e.g., \cite{fang2021spatial}), to model spatial and temporal dependencies. In contrast, continuous coupled models employ a single continuous process to accomplish this task, such as \cite{jin2022multivariate} and \cite{choi2022graph}.

\textbf{Discrete Architectures.} Numerous existing GNN-based time series forecasting methods are models processing discrete data. For instance, factorized approaches like STGCN~\cite{yu2018spatio} employ a sandwich structure of graph and temporal gated convolution layers as its fundamental building block, facilitating the modeling of inter-variable and inter-temporal relations. Subsequent works, such as DGCNN~\cite{diao2019dynamic}, LSGCN~\cite{huang2020lsgcn}, STHGCN~\cite{sawhney2020spatiotemporal}, and HGCN~\cite{guo2021hierarchical}, retain this model architecture while introducing enhancements such as dynamic graph structure estimation~\cite{diao2019dynamic}, hypergraph convolution~\cite{guo2021hierarchical}, and hierarchical graph generation~\cite{guo2021hierarchical}. A multitude of other studies adhere to similar principles, stacking diverse spatial and temporal modules in their core building blocks. For example, ST-MetaNet~\cite{pan2019urban} interweaves RNN cells and GAT~\cite{velivckovic2017graph} to model evolving traffic information. Comparable works include ST-MGCN~\cite{geng2019spatiotemporal}, DSATNET~\cite{tang2022domain}, and EGL~\cite{ye2022learning}. In contrast, ASTGCN~\cite{guo2019attention}, DSTAGNN~\cite{lan2022dstagnn}, and GraphSleepNet~\cite{jia2020graphsleepnet} are constructed upon spatial-temporal attention and convolution modules, with the latter module comprising of stacking ChebConv~\cite{defferrard2016convolutional} and the convolution in the temporal dimension. Graph WaveNet~\cite{wu2019graph}, SLCNN~\cite{zhang2020spatio}, StemGNN~\cite{cao2020spectral}, MTGNN~\cite{wu2020connecting}, STFGNN~\cite{li2021spatial}, and TGC~\cite{jin2023powerful} share a similar model architecture without the attention mechanism. There are also alternative designs within the realm of discrete factorized forecasting models. For instance, STAR~\cite{yu2020spatio} integrates the proposed spatial and temporal transformers, while ST-GDN~\cite{zhang2021traffic} initially performs attention-based temporal hierarchical modeling before applying various graph domain transformations. TPGNN~\cite{liu2022multivariate} employs temporal attention and the proposed temporal polynomial graph module to more effectively capture time-evolving patterns in time series data. MTHetGNN~\cite{wang2022mthetgnn} stacks the proposed temporal, relational, and heterogeneous graph embedding modules to jointly capture spatial-temporal patterns in time series data. CausalGNN~\cite{wang2022causalgnn} models multivariate time series with causal modeling and attention-based dynamic GNN modules. Auto-STGCN~\cite{wang2023auto} explores high-performance discrete combinations of different spatial and temporal modules.

In the realm of discrete coupled models, early works such as DCRNN~\cite{lidiffusion} and Cirstea et al.\cite{cirstea2019graph} straightforwardly incorporate graph diffusion or attention models into RNN cells. This approach models spatial-temporal dependencies in historical observations for forecasting. Subsequent works, including ST-UNet\cite{yu2019st}, MRA-BGCN~\cite{chen2020multi}, STGNN*\cite{wang2020traffic}, AGCRN\cite{bai2020adaptive}, RGSL~\cite{yu2022regularized}, and MegaCRN~\cite{jiang2022spatio}, are based on similar concepts but with varying formulations of graph convolutional recurrent units. Some studies integrate spatial and temporal convolution or attention operations into a single module. For instance, GMAN~\cite{zheng2020gman} proposes a spatial-temporal attention block that integrates the spatial and temporal attention mechanisms in a gated manner. Z-GCNETs~\cite{chen2021z} initially learns time-aware topological features that persist over time (i.e., zigzag persistence representations), then applies spatial and temporal graph convolutions to capture salient patterns in time series data. TAMP-S2GCNETS~\cite{chen2022tamp} is slightly more complex, modeling the spatial-temporal dependencies in time series by coupling two types of GCN layers, dynamic Euler-Poincare surface representation learning (DEPSRL) modules, and CNNs. Another line of research direction models spatial-temporal dependencies by convolving on specially crafted graph structures (e.g., STSGCN~\cite{song2020spatial}), performing graph convolutions in a sliding window manner (e.g., STSGCN~\cite{song2020spatial} and STG2Seq~\cite{bai2019stg2seq}), or utilizing the concept of temporal message passing (e.g., METRO~\cite{cui2021metro} and ASTTN~\cite{feng2022adaptive}).

\textbf{Continuous Architectures.} To date, only a handful of existing methods fall into the category of continuous models. For factorized methods, STGODE~\cite{fang2021spatial} proposes to depict the graph propagation as a continuous process with a neural ordinary differential equation (NODE)~\cite{chen2018neural}. This approach allows for the effective characterization of long-range spatial-temporal dependencies in conjunction with dilated convolutions along the time axis. For coupled methods, MTGODE~\cite{jin2022multivariate} generalizes both spatial and temporal modeling processes found in most related works into a single unified process that integrates two NODEs. STG-NCDE~\cite{choi2022graph} shares a similar idea but operates under the framework of neural controlled differential equations (NCDEs)~\cite{kidger2020neural}. Similarly, a recent work, TGNN4I~\cite{oskarsson2023temporal}, integrates GRU~\cite{chung2014empirical} and MPNN~\cite{gilmer2017neural} as the ODE function to model continuous-time latent dynamics.

\section{GNNs for Time Series Anomaly Detection}\label{appx:anomaly detection}
Time series anomaly detection aims to identify data observations that do not conform with the nominal regime of the data-generating process~\cite{hawkins1980identification}. We define anomaly as any such data point, and use the term normal data otherwise; we note however that different terminologies, like novelty and outlier, are used almost interchangeably to anomaly in the literature \cite{pimentel2014review}.
Deviations from the nominal conditions could take the form of a single observation (point) or a series of observations (subsequence)~\cite{darban2022deep}. However, unlike normal time series data, anomalies are difficult to characterize for two main reasons. Firstly, they are typically associated with rare events, so collecting and labeling them is often a daunting task. Secondly, establishing the full range of potential anomalous events is generally impossible, spoiling the effectiveness of supervised learning techniques. Consequently, unsupervised detection techniques have been widely explored as a practical solution to real-world problems.

Traditionally, methods~\cite{knox1998algorithms} such as distance-based~\cite{keogh2005hot,breunig2000lof,hahsler2016clustering}, and distributional techniques~\cite{ting2021isolation} have been widely used for detecting irregularities in time series data. The former family uses distance measures to quantify the discrepancy of observations from representative data points, while the latter looks at points of low likelihood to identify anomalies.

The advent of deep learning has sparked significant advancements, drawing lessons from earlier methods. Early research in this area proposed recurrent models with reconstruction \cite{park2018multimodal} and forecasting\cite{hundman2018detecting} strategies respectively to improve anomaly detection in multivariate time series data. The forecasting and reconstruction strategies rely on forecast and reconstruction errors as discrepancy measures between anticipated and real signals. 
These strategies rely on the fact that, if a model trained on normal data fails to forecast or reconstruct some data, then it is more likely that such data is associated with an anomaly. However, recurrent models \cite{su2019robust} are found to lack explicit modeling of pairwise interdependence among variable pairs, limiting their effectiveness in detecting complex anomalies \cite{zhao2020multivariate, xu2021anomaly}. Recently, GNNs have shown promising potential to address this gap by effectively capturing temporal and spatial dependencies among variable pairs \cite{deng2021graph,han2022learning,chen2021learning}.

\subsection{General \revision{Approach} for Anomaly Detection}
Treating anomaly detection as an unsupervised task relies on models to learn a general concept of what normality is for a given dataset~\cite{jin2021anemone, zheng2021generative}. To achieve this, deep learning architectures deploy a bifurcated modular framework, constituted by a backbone module and a scoring module \cite{garg2021evaluation}. 
Firstly, a backbone model, $\textsc{Backbone}(\cdot)$, is trained to fit given training data, assumed to be nominal, or to contain very few anomalies.
Secondly, a scoring module, $\textsc{Scorer} (\mathbf{X}, \mathbf{\hat{X}})$, produces a score used to identify the presence of anomalies by comparing the output $\mathbf{\hat{X}}=\textsc{Backbone}(\mathbf{X})$ of the backbone module with the observed time series data $\mathbf{X}$. The score is intended as a measure of the discrepancy between the expected signals under normal and anomalous circumstances. When there is a high discrepancy score, it is more likely that an anomaly event has occurred. 
Furthermore, it is also important for a model to diagnose anomaly events by pinpointing the responsible variables. Consequently, a scoring function typically computes the discrepancy for each individual channel first, before consolidating these discrepancies across all channels into a single anomaly value.

To provide a simple illustration of the entire process, the backbone can be a GNN forecaster that makes a one-step-ahead forecast for the scorer. The scorer then computes the anomaly score as the sum of the absolute forecast error for each channel variable, represented as $\sum_i^N |x_t^i - \hat{x}_t^i|$ across $N$ channel variables. Since the final score is computed based on the summation of channel errors, an operator can determine the root cause variables by computing the contribution of each variable to the summed error.

Advancements in the anomaly detection and diagnosis field have led to the proposal of more comprehensive backbone and scoring modules~\cite{garg2021evaluation,darban2022deep}, primarily driven by the adoption of GNN methodologies ~\cite{zhao2020multivariate,deng2021graph,han2022learning,ho2023graph}.
 
\subsection{Discrepancy Frameworks for Anomaly Detection}
Most of anomaly detection methods follow the same backbone-scorer architecture. However, the way the backbone module is trained to learn data structure from nominal data and the implementation of the scoring module differentiate these methods into three categories: (1) \emph{reconstruction}, (2) \emph{forecast}, and (3) \emph{relational discrepancy} frameworks.

\textbf{Reconstruction Discrepancy.} Reconstruction discrepancy frameworks rely on the assumption that reconstructed error should be low during normal periods, but high during anomalous periods. From a high-level perspective, they are fundamentally designed to replicate their inputs as outputs like autoencoders~\cite{goodfellow2016deep}. However, it is assumed the backbone is sufficiently expressive to model and reconstruct well the training data distribution, but not out-of-sample data.
Therefore, a reconstruction learning framework often incorporates certain constraints and regularization terms, e.g., to enforce a low-dimensional embedded code~\cite{ranzato2007sparse} or applying variational objectives~\cite{kingma2019introduction}.

Once the data structure has been effectively learned, the backbone model should be able to approximate the input during non-anomalous periods, as this input would closely resemble the normal training data. In contrast, during an anomalous event, the backbone model is expected to struggle with reconstructing the input, given that the input patterns deviate from the norm and is situated outside the manifold. Using the reconstructed outputs from the backbone module, a discrepancy score is computed by the $\textsc{Scorer}(\cdot)$ to determine whether an anomalous event has occurred. Although deep reconstruction models generally follow these principles for detecting anomalies, a key distinction between GNNs and other architectural types rests in the backbone reconstructor, $\textsc{Backbone}(\cdot)$, which is characterized by its spatiotemporal GNN implementation.

MTAD-GAT~\cite{zhao2020multivariate} utilizes a variational objective~\cite{kingma2019introduction} to train the backbone reconstructor module. During inference, the reconstructor module will provide the likelihood of observing each input channel signal to the scorer. The scorer then summarizes the likelihoods into a single reconstruction discrepancy as an anomaly score. With the availability of reconstruction probability for each channel variable, MTAD-GAT can diagnose the anomaly scorer by computing the contribution of each variable to the discrepancy score. While MTAD-GAT shares the same variational objective as LSTM-VAE~\cite{park2018multimodal}, MTAD-GAT differs by employing graph attention network as a spatial-temporal encoder to learn inter-variable and inter-temporal dependencies. Empirically, it is shown to outperform LSTM on the same VAE objective\footnote{While MTAD-GAT also optimizes their network using forecasting objective, its ablation shows that using reconstruction alone on GNN can outperform its LSTM counterpart.}. Interestingly, MTAD-GAT also shows that attention scores in the graph attention network reflect substantial differences between normal and anomalous periods.

GNNs require knowledge of graph structures that is often not readily available for time series anomaly detection data~\cite{su2019robust,li2021multivariate}. To solve this issue, MTAD-GAT plainly assumes a fully-connected graph between the spatial variables in a multivariate time series. This assumption may not necessarily hold true in real-world scenarios and can potentially create unnecessary noise that weakens the ability to learn the underlying structure of normal data. 

In response, VGCRN~\cite{chen2022deep} first assigns learnable embeddings to the channel variates of a time series data. VGCRN then generates the channel-wise similarity matrix as the learned graph structure by computing the dot product between the embeddings. Under the same graph structural learning framework, FuSAGNet~\cite{han2022learning} proposes to learn a static and sparse directed graph by taking only the top-$k$ neighbors for each node in the similarity matrix. FuSAGNet however differs in the reconstruction framework by learning a sparse embedded code~\cite{ranzato2007sparse} rather than optimizing for variational objectives. 

In the category of reconstruction-based methods, a research direction focused on graph-level embeddings to represent the input graph data as vectors to enable the application of well-established and sophisticated detection methods designed for multivariate time series.
The works by Zambon et al.~\cite{zambon2018concept,zambon2019change} laid down the general framework which has been instantiated in different nuances.
Some papers design low-dimensional embedding methods trained so that the distance between any two vector representations best preserves the graph distance between their respective input graphs \cite{zambon2017detecting, zambon2018anomaly}.
Conversely, GIF \cite{zambon2022graph} employs a high-dimension graph-level embedding method based on the idea of random Fourier feature to discover anomalous observations.
CCM-CDT~\cite{grattarola2019change} is a graph autoencoder with Riemannian manifolds as latent spaces. The autoencoder, trained adversarially to reconstruct the input, produces a sequence of manifold points where statistical tests are performed to detect anomalies and changes in the data distribution.

\textbf{Forecast Discrepancy.} Forecast discrepancy frameworks rely on the assumption that forecast error should be low during normal periods, but high during anomalous periods. Here, the backbone module is substituted with a GNN forecaster that is trained to predict a one-step-ahead forecast. During model deployment, the forecaster makes a one-step-ahead prediction, and the forecast values are given to the scorer. The scorer compares the forecast against the real observed signals to compute forecast discrepancies such as absolute error~\cite{deng2021graph} or mean-squared error~\cite{chen2021learning}. Importantly, it is generally assumed that a forecasting-based model will exhibit erratic behavior during anomaly periods when the input data deviates from the normal patterns, resulting in a significant forecasting discrepancy.

A seminal work in applying forecasting-based GNN to detect anomalies in time series data is GDN~\cite{deng2021graph}. The forecaster of GDN consists of two main parts: first, a graph structure module that learns an underlying graph structure, and second, a graph attention network that encodes the input series representation. The graph structure module computes the graph adjacency matrix for the graph attention network on the learned graph to obtain an expressive representation for making a one-step-ahead forecast. Finally, the scorer computes the forecast discrepancy as the maximum absolute forecast error among the channel variables to indicate whether an anomaly event has occurred. 

Interestingly,  GDN illustrates that an anomaly event may be manifested in a single variable, which acts as a symptom, while the underlying cause may be traced to a separate, root-cause variable. Hence, GDN proposed to utilize the learned relationships between variables for diagnosing the root causes of such events rather than relying solely on the individual contributions of each variable to the discrepancy score for diagnosing the root cause of anomaly events. This is accomplished by identifying the symptomatic variable that results in the maximum absolute error, followed by pinpointing its neighbour variables. The ability of GDN to discern these associations underscores the potential of GNNs in offering a more holistic solution to anomaly detection and diagnosis through the automated learning of inter-variable relationships.

Within the context of statistical methods, the AZ-whiteness test~\cite{zambon2022aztest} operates on the prediction residuals obtained by forecasting models. Assuming that a forecasting model is sufficiently good in modeling the nominal data-generating process, the statistical test is able to identify unexpected correlations in the data that indicate shifts in the data distribution. The AZ-test is also able to distinguish between serial correlation, i.e., along the temporal dimension, and spatial correlation observed between the different graph nodes.
In a similar fashion, the AZ-analysis of residuals~\cite{zambon2023where} expands the set of analytical tools to identify anomalous nodes and time steps, thus providing a finer inspection and diagnosis.

\revision{While the forecast discrepancy frameworks can be similar to reconstruction discrepancy frameworks, both rely on fundamentally different principles and thus implementation strategies. The reconstruction discrepancy framework projects the current timestamp's input onto a learned latent space and attempts to reconstruct it, identifying anomalies when reconstruction fails, as this indicates a significant deviation from the learned normal patterns. This method directly uses current input data during both training and inference, centering on the model's ability to reconstruct well the training data distribution but not out-of-sample data. Conversely, the forecast discrepancy framework does not directly use the current timestamp data when making a forecast. Instead, it relies on historical data to predict current values and identifies anomalies by comparing these predictions with actual current observations, flagging significant discrepancies. Thus, reconstruction focuses on accurately replicating current inputs, while forecasting emphasizes predicting future data points and detecting anomalies through prediction errors. Advanced GNN forecast discrepancy framework, GST-Pro~\cite{zheng2024graph}, can even predict anomalies without using actual observations in the scoring function, allowing for the prediction of anomalies at future timestamps before they occur.}

\textbf{Relational Discrepancy.} Relational discrepancy frameworks rely on the assumption that the relationship between variables should exhibit significant shifts from normal to anomalous periods.
This direction has been alluded to in the MTAD-GAT work, where it was observed that attention weights in node neighborhoods tend to deviate substantially from normal patterns during anomaly periods. Consequently, the logical evolution of using spatiotemporal GNN involves leveraging its capability to learn graph structures for both anomaly detection and diagnosis. In this context, the backbone module serves as a graph learning module that constructs the hidden evolving relationship between variables. The scorer, on the other hand, is a function that evaluates changes in these relationships and assigns an anomaly or discrepancy score accordingly.

GReLeN~\cite{zhang2022grelen} was the first method to leverage on learned dynamic graphs to detect anomalies from the perspective of relational discrepancy. To achieve this, the reconstruction module of GReLeN learns to dynamically construct graph structures that adapt at every time point based on the input time series data. The constructed graph structures serve as the inputs for a scorer that computes the total changes in the in-degree and out-degree values for channel nodes. GReLeN discovered that by focusing on sudden changes in structural relationships, or \textit{relational discrepancy}, at each time point, they could construct a robust metric for the detection of anomalous events.

On the other hand, DyGraphAD adopts a forecasting approach to compute relational discrepancy~\cite{chen2023DyGraphAD}. The method begins by dividing a multivariate series into subsequences and converting these subsequences into a series of dynamically evolving graphs. To construct a graph for each subsequence, DyGraphAD employs the DTW distance between channel variables based on their respective values in the subsequence. Following this preprocessing step, the DTW distance graphs are treated as ground truth or target and the network is trained to predict one-step-ahead graph structures. The scorer of DyGraphAD computes the forecast error in the graph structure as the relational discrepancy for anomaly detection. 

\textbf{Hybrid and Other Discrepancies.} Each type of discrepancy-based framework often possesses unique advantages for detecting and diagnosing various kinds of anomalous events. As demonstrated in GDN\cite{deng2021graph}, the relational discrepancy framework can uncover spatial anomalies that are concealed within the relational patterns between different channels. In contrast, the forecast discrepancy framework may be particularly adept at identifying temporal anomalies such as sudden spikes or seasonal inconsistencies. Therefore, a comprehensive solution would involve harnessing the full potential of spatial-temporal GNNs by computing a hybridized measure that combines multiple discrepancies as indicators for anomaly detection. For instance, MTAD-GAT\cite{zhao2020multivariate} and FuSAGNet\cite{han2022learning} employ both reconstruction and forecast discrepancy frameworks, while DyGraphAD\cite{chen2023DyGraphAD} utilizes a combination of forecast and relational discrepancy frameworks to enhance the detection of anomalous events. In these cases, the scoring function should be designed to encapsulate the combination of either reconstruction, forecast or relational discrepancy. In general, the anomaly score can be represented as $ S_{t} = \|\mathbf{X}_{t} -\mathbf{\hat{X}}_{t}\|_2^2 + \| \mathbf{A}_{t} - \mathbf{\hat{A}}_{t} \|_F^2$, which captures the reconstruction and relational discrepancies, respectively. Here, $\mathbf{X}_{t}$ and $\mathbf{A}_{t}$  represent the target signals and inter-variable relations, while $\mathbf{\hat{X}}_{t}$ and $\mathbf{\hat{A}}_{t}$ denote the predicted signal and inter-variable relations.

Apart from learning the underlying structures of normal training data, another approach to detecting time series anomalies involves incorporating prior knowledge of how a time series might behave during anomalous events. With this aim in mind, GraphSAD \cite{chen2023time} considers six distinct types of anomalies, including spike and dip, resizing, warping, noise injection, left-side right, and upside down, to create pseudo labels on the training data. By doing so, unsupervised anomaly detection tasks can be transformed into a standard classification task, with the class discrepancy serving as an anomaly indicator.

\section{GNNs for Time Series Classification}\label{appx:classification}
Time series classification task seeks to assign a categorical label to a given time series based on its underlying patterns or characteristics. As outlined in a recent survey~\cite{foumani2023deep}, early literature in time series classification primarily focused on distance-based approaches for assigning class labels to time series~\cite{lines2015time,tan2020fastee,herrmann2021amercing}, and ensembling methods such as hierarchical vote collective of transformation-based ensembles (HIVE-COTE)~\cite{lines2018time,middlehurst2021hive}. However, despite their state-of-the-art performances, the scalability of both approaches remains limited for high-dimensional or large datasets~\cite{dempster2020rocket,tan2022multirocket}. 

To address these limitations, researchers have begun to explore the potential of deep learning techniques for enhancing the performance and scalability of time series classification methods. Deep learning, with its ability to learn complex patterns and hierarchies of features, has shown promise in its applicability to time series classification problems, especially for datasets with substantial training labels~\cite{zhang2020tapnet,zerveas2021transformer}. For a comprehensive discussion on deep learning-based time series classification, we direct readers to the latest survey by Foumani et al.~\cite{foumani2023deep}. 

One particularly intriguing development in this area not covered by the aforementioned survey~\cite{foumani2023deep} is the application of GNN to time series classification tasks. By transforming time series data into graph representations, one can leverage the powerful capabilities of GNNs to capture both local and global patterns. Furthermore, GNNs are capable of mapping the intricate relationships among different time series data samples within a particular dataset.

In the following sections, we provide a fresh GNN perspective on the univariate and multivariate time series classification problem.

\subsection{Univariate Time Series Classification}
Inherent in the study of time series classification lies a distinct differentiation from other time series analyses; rather than capturing patterns within time series data, the essence of time series classification resides in discerning differentiating patterns that help separate time series data samples based on their class labels. 

For example, in the healthcare sector, time series data in the form of heart rate readings can be utilized for health status classification. A healthy individual may present a steady and rhythmic heart rate pattern, whereas a patient with cardiovascular disease may exhibit patterns indicating irregular rhythms or elevated average heart rates. Unlike forecasting future points or detecting real-time anomalies, the classification task aims to distinguish these divergent patterns \textit{across series}, thereby enabling health status classification based on these discerned patterns.

In the following, we delve into two novel graph-based approaches to univariate time series classification, namely \revision{Series-as-Graph} and \revision{Series-as-Node}.

\textbf{\revision{Series-as-Graph.}} The \textit{\revision{Series-as-Graph}} approach transforms a univariate time series into a graph to identify unique patterns that enable accurate classification using a GNN. In this manner, each series is treated as a graph, and the graph will be the input for a GNN to make classification outputs. 

Firstly, each series is broken down into subsequences as nodes, and the nodes are connected with edges illustrating their relationships. Following this transformation, a GNN is applied to make graph classification. This procedure is represented in the upper block of Fig. 4d. Fundamentally, it seeks to model inter-temporal dependencies under a GNN framework to identify differentiating patterns across series samples into their respective classes. 

The graph classification perspective of \revision{Series-as-Graph} was first proposed by the Time2Graph technique~~\cite{cheng2020time2graph}. This approach was later developed further with the incorporation of GNN, as Time2Graph+~\cite{cheng2021time2graph+}. The Time2Graph+ modeling process can be described as a two-step process: first, a time series is transformed into a shapelet graph, and second, a GNN is utilized to model the relations between shapelets. To construct a shapelet graph, the Time2Graph algorithm partitions each time series into successive segments. It then employs data mining techniques to assign representative shapelets to the subsequences. These shapelets serve as graph nodes. Edges between nodes are formed based on the conditional probability of a shapelet occurring after another within a time series. Consequently, each time series is transformed into a graph where shapelets form nodes and transition probabilities create edges. After graph construction, Time2Graph+ utilizes graph attention networks along with a pooling operation to derive the global representation of the time series. This representation is then fed into a classifier to assign class labels to the time series. \revision{Similarly, EC-GCN~\cite{diao2023ec} constructs graphs from encrypted traffic for classification. The Series-as-Graph approach can also be extended to multivariate time-series classification task~\cite{younis2024mts2graph}.}

While we use \revision{Series-as-Graph} to describe the formulation of a time series classification task as a graph classification task, this should not be confused with the Series2Graph \cite{boniol13series2graph} approach, which is utilized for the anomaly detection tasks. The strategy of recasting a series as a graph is not unique to time-series classification, as demonstrated by Series2Graph. The usage of this strategy across various tasks and domains highlights its inherent adaptability and underscores its potential worth exploring further by researchers.

\textbf{\revision{Series-as-Node.}} As capturing differentiating class patterns across different series data samples is important, leveraging relationships across the different series data samples in a given dataset can be beneficial for classifying a time series. To achieve this, one can take the \textit{\revision{Series-as-Node}} approach, where each series sample is seen as a separate node. These series nodes are connected with edges that represent the relationships between them, creating a large graph that provides a complete view of the entire dataset. 

The \revision{Series-as-Node} was originally proposed by SimTSC~\cite{zha2022towards} approach. With SimTSC, series nodes are connected using edges, which are defined by their pairwise DTW distance, to construct a graph. During the modeling process, a primary network is initially employed to encode each time series into a feature vector, thus creating a node representation. Subsequently, a standard GNN operation is implemented to derive expressive node representations, capturing the similarities between the series. These node representations are then inputted into a classifier, which assigns class labels to each time series node in the dataset. LB-SimTSC~\cite{xi2023lb} extends on SimTSC to improve the DTW preprocessing efficiency by employing the widely-used lower bound for DTW, known as LB\_Keogh~\cite{keogh2005exact}. This allows for a time complexity of $O(L)$ rather than $O(L^2)$, dramatically reducing computation time.

The \revision{Series-as-Node} process essentially formulates a time series classification task as a node classification task. As illustrated in the lower block of Fig. 4d, the \revision{Series-as-Node} perspective aims to leverage the relationships across different series samples for accurate time series node classification~\cite{xi2023lb}. It is also an attempt to marry classical distance-based approaches with advanced GNN techniques. While not explicitly depicted in the figure, it is important to note that the same concept can be applied to classify multivariate time series by modifying the backbone network that transforms each series into a node~\cite{zha2022towards}.

\subsection{Multivariate Time Series Classification}
In essence, multivariate time series classification maintains fundamental similarities with its univariate counterpart, however, it introduces an additional layer of complexity: the necessity to capture intricate inter-variable dependencies. 

For example, instead of solely considering heart rate, patient data often incorporate time series from a multitude of health sensors, including blood pressure sensors, blood glucose monitors, pulse oximeters, and many others. Each of these sensors provides a unique time series that reflects a particular aspect of the health of a patient. By considering these time series together in a multivariate analysis, we can capture more complex and interrelated health patterns that would not be apparent from any single time series alone.

Analogously, each node in an electroencephalogram (EEG) represents electrical activity from a distinct brain region. Given the interconnectedness of brain regions, analyzing a single node in isolation may not fully capture the comprehensive neural dynamics~\cite{tang2022selfsupervised}. By employing multivariate time series analysis, we can understand the relationships between different nodes, thereby offering a more holistic view of brain activity. This approach facilitates the differentiation of intricate patterns that can classify patients with and without specific neurological conditions.

In both examples, the relationships between the variables, or inter-variable dependencies, can be naturally thought of as a network graph. Hence, they ideally suit the capabilities of GNNs as illustrated in forecasting Sec. 4. As such, spatial-temporal GNNs, exemplified by those utilized in forecasting tasks~\cite{wu2020connecting}, are conveniently adaptable for multivariate time series classification tasks. This adaptation can be achieved through the replacement of the final layer with a classification component. The unique design of these STGNN architectures enables the capture of both inter-temporal and inter-variable dependencies. The primary aim here is to effectively distill the complexity of high-dimensional series data into a more comprehensible, yet equally expressive, representation that enables differentiation of time series into their representative classes~\cite{duan2022multivariate,liu2023todynet}.

The proficiency of spatial-temporal GNNs in decoding the complexities of multivariate time series is demonstrably showcased in the Raindrop architecture~\cite{zhang2022graphguided}. To classify irregularly sampled data where subsets of variables have missing values at certain timestamps, Raindrop adaptively learns a graph structure. It then dynamically interpolates missing observations within the embedding space, based on any available recorded data. This flexible approach ensures that the data representation remains both comprehensive and accurate, despite any irregularities in the sampling. Empirical studies provide evidence that Raindrop can maintain robust, high-performance classification, even in the face of such irregularities~\cite{zhang2022graphguided}. These findings further reinforce the versatility of spatial-temporal GNNs in time series classification, highlighting their effectiveness even in scenarios characterized by missing data and irregular sampling patterns.

\section{GNNs for Time Series Imputation}\label{appx:imputation}
Time series imputation, a crucial task in numerous real-world applications, involves estimating missing values within one or more data point sequences. Traditional time series imputation approaches have relied on statistical methodologies, such as mean imputation, spline interpolation~\cite{moritz2017imputets}, and regression models~\cite{saad2020machine}. However, these methods often struggle to capture complex temporal dependencies and non-linear relationships within the data. While some deep neural network-based works, such as \cite{che2018recurrent, yoon2018gain, miao2021generative}, have mitigated these limitations, they have not explicitly considered inter-variable dependencies. The recent emergence of graph neural networks has introduced new possibilities for time series imputation. GNN-based methods better characterize intricate spatial and temporal dependencies in time series data, making them particularly suitable for real-world scenarios arising from the increasing complexity of data. From a task perspective, GNN-based time series imputation can be broadly categorized into two types: \textit{in-sample imputation} and \textit{out-of-sample imputation}. The former involves filling in missing values within the given time series data, while the latter predicts missing values in disjoint sequences~\cite{cini2022filling}. From a methodological perspective, GNN for time series imputation can be further divided into \textit{deterministic} and \textit{probabilistic imputation}. Deterministic imputation provides a single best estimate for the missing values, while probabilistic imputation accounts for the uncertainty in the imputation process and provides a distribution of possible values. In Tab. 5, we summarize most of the related works on GNN for time series imputation to date, offering a comprehensive overview of the field and its current state of development. 

\subsection{In-sample Imputation}
The majority of existing GNN-based methods primarily focus on in-sample time series data imputation. For instance, GACN~\cite{ye2021spatial} proposes to model spatial-temporal dependencies in time series data by interleaving GAT~\cite{velivckovic2017graph} and temporal convolution layers in its encoder. It then imputes the missing data by combining GAT and temporal deconvolution layers that map latent states back to original feature spaces. Similarly, SPIN~\cite{mariscalearning} first embeds historical observations and sensor-level covariates to obtain initial time series representations. These are then processed by multi-layered sparse spatial-temporal attention blocks before the final imputations are obtained with a nonlinear transformation. GRIN~\cite{cini2022filling} introduces the graph recurrent imputation network, where each unidirectional module consists of one spatial-temporal encoder and two different imputation executors. The spatial-temporal encoder adopted in this work combines MPNN~\cite{gilmer2017neural} and GRU~\cite{chung2014empirical}. After generating the latent time series representations, the first-stage imputation fills missing values with one-step-ahead predicted values, which are then refined by a final one-layer MPNN before passing to the second-stage imputation for further processing. Similar works using bidirectional recurrent architectures include AGRN~\cite{chen2023adaptive}, DGCRIN~\cite{kong2023dynamic}, GARNN~\cite{shen2023bidirectional}, and MDGCN~\cite{liang2022memory}, where the main differences lie in intermediate processes. For example, AGRN and DGCRIN propose different graph recurrent cells that integrate graph convolution and GRU to capture spatial-temporal relations, while GARNN involves the use of GAT and different LSTM~\cite{hochreiter1997long} cells to compose a graph attention recurrent cell in its model architecture. MDGCN models time series as dynamic graphs and captures spatial-temporal dependencies by stacking bidirectional LSTM and graph convolution. 
\revision{Recently, a few research studies have explored probabilistic in-sample time series imputation, such as PriSTI~\cite{liu2023pristi} and \cite{yun2023imputation}, where the imputation has been regarded as a generation task. In PriSTI, a similar architecture of denoising diffusion probabilistic models~\cite{ho2020denoising} has been adopted to effectively sample the missing data with a spatial-temporal denoising network composed of attentive MPNN and temporal attention. \cite{yun2023imputation} adopts similar concept but with GAT~\cite{velivckovic2017graph} as the backbone denoising network.}

\subsection{Out-of-sample Imputation}
To date, only a few GNN-based methods fall into the category of out-of-sample imputation. Among these works, IGNNK~\cite{wu2021inductive} proposes an inductive GNN kriging model to recover signals for unobserved time series, such as a new variable or ``virtual sensor'' in a multivariate time series. In IGNNK, the training process involves masked subgraph sampling and signal reconstruction with the diffusion graph convolution network presented in \cite{li2018diffusion}. Another similar work is SATCN~\cite{wu2021spatial}, which also focuses on performing real-time time series kriging. The primary difference between these two works lies in the underlying GNN architectures, where SATCN proposes a spatial aggregation network combined with temporal convolutions to model the underlying spatial-temporal dependencies. 
\revision{INCREASE~\cite{zheng2023increase}, on the other hand, further considers heterogeneous spatial and diverse temporal relations among the locations, lifting the performance of inductive spatio-temporal kriging.}
It is worth noting that GRIN~\cite{cini2022filling} can handle both in-sample and out-of-sample imputations, as well as a similar follow-up work~\cite{roth2022forecasting}.

\section{Practical Applications}\label{appx:application}

Graph neural networks have been applied to a broad range of disciplines related to time series analysis. We categorize the mainstream applications of GNN4TS into \revision{seven} areas: smart transportation, on-demand services, environment \& sustainable energy, internet-of-things, \revision{physical systems,} healthcare, and fraud detection.
\\

\noindent\textbf{Smart Transportation.} The domain of transportation has been significantly transformed with the advent of GNNs, with typical applications spanning from traffic prediction to flight delay prediction. Traffic prediction, specifically in terms of traffic speed and volume prediction, is a critical component of smart transportation systems. By leveraging advanced algorithms and data analytics related to spatial-temporal GNNs, traffic conditions can be accurately predicted~\cite{lidiffusion, yu2018spatio, pan2019urban, chen2020multi, lan2022dstagnn, tang2022domain, li2023traffic}, thereby facilitating efficient route planning and congestion management. Another important application is traffic data imputation, which involves the estimation of missing or incomplete traffic data. This is crucial for maintaining the integrity of traffic databases and ensuring the accuracy of traffic analysis and prediction models~\cite{liang2021dynamic, wu2022multi, kong2023dynamic, shen2023bidirectional, liang2022memory}. There is also existing research related to autonomous driving with 3D object detection and motion planners based on GNNs~\cite{tang2023trajectory, mo2023predictive, wang2023sat}, which has the potential to drastically improve road safety and traffic efficiency. Lastly, flight delay prediction is another significant application that can greatly enhance passenger experience and optimize airline operations. This is achieved through the analysis of various factors such as weather conditions, air traffic, and aircraft maintenance schedules~\cite{cai2021deep, guo2020sgdan}. In summary, smart transportation, through its diverse applications, is paving the way for a more efficient, safe, and convenient transportation system. The integration of advanced technologies, such as GNNs, in these applications underscores the transformative potential of smart transportation, highlighting its pivotal role in shaping the future of transportation. \\

\noindent\textbf{On-demand Services.} For systems providing goods or services upon request, GNNs have emerged as powerful tools for modeling time series data to accurately predict personalized real-time demands, including transportation, energy, tourism, and more. For instance, in ride-hailing services, GNNs capture the complex, temporal dynamics of ride demand across different regions, enabling accurate prediction of ride-hailing needs and thereby facilitating efficient fleet management~\cite{yao2018deep, geng2019spatiotemporal, wu2020multi, xu2022multi2, bai2019stg2seq, bai2019spatio, tang2021multi}. Similarly, in bike-sharing services, GNNs leverage the spatial-temporal patterns of bike usage to accurately predict demand, contributing to the optimization of bike distribution and maintenance schedules~\cite{kim2019graph, he2020towards, chen2021comparative, xiao2021demand, ma2022short, li2022data}. In the energy sector, GNNs model the intricate relationships between various factors influencing energy demand, providing accurate predictions that aid in the efficient management of energy resources~\cite{lin2021residential}. In the tourism industry, GNNs capture the temporal trends and spatial dependencies in tourism data, providing accurate predictions of tourism demand and contributing to the optimization of tourism services and infrastructure~\cite{zhou2023graph, zhuang2022uncertainty, zhao2022coupling, liang2023region}. There are also GNN-based works that model the complex spatial-temporal dynamics of delivery demand, accurately predicting delivery needs and facilitating efficient logistics planning and operations~\cite{wen2022graph2route}. The advent of GNN4TS has significantly improved the accuracy of demand prediction in on-demand services, enhancing their efficiency and personalization. The integration of GNNs in these applications underscores their transformative potential, highlighting their pivotal role in shaping the future of on-demand services. \\

\noindent\textbf{Environment \& Sustainable Energy.} In the sector related to environment and sustainable energy, GNNs have been instrumental in wind speed and power prediction, capturing the complex spatial-temporal dynamics of wind patterns to provide accurate predictions that aid in the efficient management of wind energy resources~\cite{khodayar2018spatio, wu2022promoting, pan2022short, fan2020m2gsnet, yu2020superposition, li2022short, he2022robust}. Similarly, in solar energy, GNNs have been used for solar irradiance and photovoltaic (PV) power prediction, modeling the intricate relationships between various factors influencing solar energy generation to provide accurate predictions~\cite{jiao2021graph, gao2022interpretable, simeunovic2021spatio, karimi2021spatiotemporal, zhang2022optimal}. In terms of system monitoring, GNNs have been applied to wind turbines and PV systems. For wind turbines, GNNs can effectively capture the temporal dynamics of turbine performance data, enabling efficient monitoring and maintenance of wind turbines~\cite{liu2023condition}. For PV systems, GNNs have been used for fault detection, leveraging the spatial dependencies in PV system data to accurately identify faults and ensure the efficient operation of PV systems~\cite{van2023cost}. Furthermore, GNNs have been employed for air pollution prediction and weather forecasting. By modeling the spatial-temporal patterns of air pollution data, GNNs can accurately predict air pollution levels, contributing to the formulation of effective air quality management strategies~\cite{zhou2021forecasting, tan2022new, oliveira2023spatiotemporal}. In weather forecasting, GNNs capture the complex, temporal dynamics of weather patterns, providing accurate forecasts that are crucial for various sectors, including agriculture, energy, and transportation~\cite{keisler2022forecasting, singh2023maximising}. \\

\noindent\textbf{Internet-of-Things (IoTs).} IoT refers to intricately linked devices that establish a network via wireless or wired communication protocols, operating in concert to fulfill shared objectives on behalf of their users \cite{madakam2015internet}. These IoT networks generate substantial quantities of high-dimensional, time series data that are increasingly complex and challenging for manual interpretation and understanding. Recent advancements have seen the application of GNNs as a powerful tool for encoding the intricate spatiotemporal relationships and dependencies inherent in IoT networks \cite{li2021multivariate,chen2021learning,dong2023graph}. GNNs leverage their ability to unravel these convoluted relationships, allowing for greater insights into the structure and behavior of these networks. This approach has garnered attention across various industrial sectors including robotics and autonomous systems \cite{lee2019joint,cai2021dignet,li2021attentional}, utility plants \cite{deng2021graph}, public services \cite{zhang2020semi}, and sports analytics \cite{li2021multiscale,stockl2021making,anzer2022detection,luo2023you}, expanding the breadth of IoT applications. With a consistent track record of state-of-the-art results\cite{dong2023graph}, GNNs have proven integral in numerous IoT applications, underpinning our understanding of these increasingly complex systems. \\

\noindent\textbf{\revision{Physical systems.}}
\revision{
Systems of interacting objects are found in numerous scientific fields, including the simulation of n-body systems \cite{battaglia2016interaction}, particle physics~\cite{shi2023towards}, modeling of human motion dynamics \cite{liu2023grapha}, and prediction of molecular dynamics \cite{wu2023equivariant}. In graph-based deep learning, the objects are represented as nodes of a graph and GNNs have proven effective in modeling their complex interactions.
Despite the challenges posed by physical constraints, promising performance has been achieved thanks to the incorporation of inductive biases from known physical laws \cite{sanchez2019hamiltonian, liu2023segno} and architectures that maintain the symmetries of the underlying system, such as equivariance to rotations and translations \cite{satorras2021equivariant, brandstetter2021geometric, wu2023equivariant}.} \\

\noindent\textbf{Healthcare.} Healthcare systems, spanning from individual medical diagnosis and treatment to broader public health considerations, present diverse challenges and opportunities that warrant the application of GNNs. In the sphere of medical diagnosis and treatment, graph structures can effectively capture the complex, temporal dynamics of diverse medical settings including electrical activity such as electronic health data\cite{li2020graphhealth,liu2020hybrid,su2020gate,li2020knowledge}, patient monitoring sensors \cite{han2019graphconvlstm,shi2020graph,zhang2022graphguided}, EEG\cite{jia2020graphsleepnet,jia2021multi,tang2022selfsupervised}, brain functional connectivity such as magnetic resonance imaging (MRI)\cite{parisot2018disease,zhang2019functional} and neuroimaging data\cite{kim2021interpretable}. Simultaneously, for public health management, GNNs have been proposed to predict health equipment useful life \cite{kong2022spatio} and forecasting ambulance demand \cite{wang2021forecasting}. More recently, GNNs have been proposed to manage epidemic disease outbreaks as temporal graphs can provide invaluable insights into disease spread, facilitating the formulation of targeted containment strategies \cite{hy2022temporal,fritz2022combining,wang2022causalgnn}. In summary, the integration of GNNs with time series data holds substantial potential for transforming healthcare, from refining medical diagnosis and treatment to strengthening population health strategies, highlighting its critical role in future healthcare research. \\

\begin{table*}[thbp]
 	\caption{\revision{A summary of selected benchmark datasets grouped by task categories.}}
	\label{tab:dataset}
	\centering
        \begin{adjustbox}{width=2\columnwidth,center}
	\begin{tabular}{l l l l l l l l }
		\toprule
		
		\multicolumn{1}{l}{\thead{\textbf{Task} \\\textbf{Category}}} & \textbf{Dataset} & \textbf{\# Samples} & \textbf{\# Nodes} & \thead{\textbf{Sampling} \\\textbf{Rate}} & \thead{\textbf{Missing} \\\textbf{Ratio}} & \textbf{Data Type} & \textbf{References} \\ 
		\bottomrule
		\multicolumn{1}{l|}{\multirow{23}{*}{\begin{tabular}[c]{@{}l@{}}Forecasting \\\& Imputation \end{tabular}}} & METR-LA \cite{lidiffusion} & 34,272 & 207 & 5 minutes & 8.109\% & Traffic Velocity &  \begin{tabular}[c]{@{}l@{}} \cite{lidiffusion,wu2019graph,chen2020multi,wu2020connecting,cao2020spectral}, \\ \cite{shang2021discrete,jin2022multivariate,shao2022pre,cini2023scalable,cini2023graphbased}
		\end{tabular}\\  
		
		\multicolumn{1}{l|}{} & PeMS-BAY \cite{lidiffusion} & 52,116 & 325 & 5 minutes & 0.003\% & Traffic Velocity  & \begin{tabular}[c]{@{}l@{}} \cite{lidiffusion,wu2019graph,chen2020multi,wu2020connecting,zheng2020gman},\\ \cite{cao2020spectral,shang2021discrete,liu2022multivariate,jin2022multivariate,shao2022pre}
		\end{tabular}\\

		\multicolumn{1}{l|}{} & PeMSD3 \cite{chen2001freeway} & 26,208 & 358 & 5 minutes & 0.672\% & Traffic Volume & \begin{tabular}[c]{@{}l@{}} \cite{song2020spatial,cao2020spectral,fang2021spatial,li2021spatial,lan2022dstagnn}, \\ \cite{choi2022graph,rao2022fogs}
		\end{tabular}\\ 

		\multicolumn{1}{l|}{} & PeMSD4 \cite{chen2001freeway} & 16,992 & 307 & 5 minutes & 3.182\% & Traffic Volume & \begin{tabular}[c]{@{}l@{}} \cite{song2020spatial,cao2020spectral,bai2020adaptive,huang2020lsgcn,chen2021z}, \\ \cite{fang2021spatial,li2021spatial,lan2022dstagnn,choi2022graph,shao2022pre}
		\end{tabular}\\ 

		\multicolumn{1}{l|}{} & PeMSD7 \cite{chen2001freeway} & 28,224 & 883 & 5 minutes & 0.452\% & Traffic Volume & \begin{tabular}[c]{@{}l@{}} \cite{yu2018spatio,song2020spatial,cao2020spectral,huang2020lsgcn,fang2021spatial}, \\ \cite{li2021spatial,lan2022dstagnn,liu2022multivariate,choi2022graph,rao2022fogs}
		\end{tabular}\\

		\multicolumn{1}{l|}{} & PeMSD8 \cite{chen2001freeway} & 17,856 & 170 & 5 minutes & 0.696\% & Traffic Volume & \begin{tabular}[c]{@{}l@{}} \cite{song2020spatial,cao2020spectral,bai2020adaptive,huang2020lsgcn,chen2021z}, \\ \cite{fang2021spatial,li2021spatial,lan2022dstagnn,choi2022graph,yu2022regularized}
		\end{tabular}\\ 

        \multicolumn{1}{l|}{} & Xiamen \cite{zheng2020gman} & 44,064 & 95 & 5 minutes & - & Traffic Volume & \begin{tabular}[c]{@{}l@{}} \cite{zheng2020gman,lau2021spatio,zheng2023increase,wang2021fine}
        \end{tabular}\\ 

        \multicolumn{1}{l|}{} & Beijing \cite{zheng2015forecasting} & 278,085 & 36 & 1 hour & - & Air Quality Index & \begin{tabular}[c]{@{}l@{}} \cite{zheng2015forecasting,zheng2023increase,zhao2020mastgn,ouyang2021spatial}, \\ \cite{han2021joint}
        \end{tabular}\\ 

        \multicolumn{1}{l|}{} & NYC-Bike \cite{yao2019revisiting} & 3.8 million & 112 & 30 minutes & - & Trip Records & \begin{tabular}[c]{@{}l@{}} \cite{li2024gpt,ye2021coupled,yao2019revisiting,lin2020preserving}, \\ \cite{pan2024urban,ye2019co}
        \end{tabular}\\ 

        \multicolumn{1}{l|}{} & NYC-Taxi \cite{yao2019revisiting} & 28.1 million & 192 & 30 minutes & - & Trip Records & \begin{tabular}[c]{@{}l@{}} \cite{li2024gpt,ye2021coupled,yao2019revisiting,lin2020preserving}, \\ \cite{pan2024urban,ye2019co}
        \end{tabular}\\ 

        \multicolumn{1}{l|}{} & NYC-Crime \cite{li2022spatial} & 31,799 - 85,899 & 256 & 1 day & - & Crime Incidents & \begin{tabular}[c]{@{}l@{}} \cite{li2022spatial,tang2023explainable}
        \end{tabular}\\ 

        \multicolumn{1}{l|}{} & AQI \cite{zheng2015forecasting} & 8,759 & 437 & 30 minutes & 25.7\% & Air Quality Index & \begin{tabular}[c]{@{}l@{}} \cite{zheng2015forecasting,cini2022filling,cini2023graphbased,chen2023group}, \\ \cite{mariscalearning,chen2023adaptive}
        \end{tabular}\\
        
        \multicolumn{1}{l|}{} & AQI-36 \cite{yi2016st} & 8,759 & 36 & 1 hour & 13.2\% & Air Quality Index & \begin{tabular}[c]{@{}l@{}} \cite{yi2016st,cini2022filling,mariscalearning,chen2023adaptive}
        \end{tabular}\\ \hline

		\multicolumn{1}{l|}{\multirow{2}{*}{\begin{tabular}[c]{@{}l@{}} Anomaly \\Detection \end{tabular}}} &  {\begin{tabular}[c]{@{}l@{}} SMD \cite{su2019robust} \end{tabular}}  & 608,342 & 38 & 1 minute & 5.84\% & Server Machine   & \begin{tabular}[c]{@{}l@{}}\cite{zhao2020multivariate,wu2021event2graph,zhang2022grelen,chen2022deep}\\ \cite{guan2022gtad,zhou2022hybrid,chen2023multivariate,zheng2023correlation}
		\end{tabular}\\ 

        \multicolumn{1}{l|}{} & SMAP \cite{hundman2018detecting} & 562,800     & 25 & 1 minute & 13.13\% & Spacecraft Telemetry  & \begin{tabular}[c]{@{}l@{}} \cite{zhao2020multivariate,chen2021learning,chen2022deep,guan2022gtad,srinivas2022hypergraph}, \\ \cite{zhou2022hybrid,chen2023multivariate}
        \end{tabular}\\

        \multicolumn{1}{l|}{} & MSL \cite{hundman2018detecting}& 132,046     & 55 & 1 minute & 10.72\% & Spacecraft Telemetry & \begin{tabular}[c]{@{}l@{}} \cite{zhao2020multivariate,chen2021learning,chen2022deep,guan2022gtad,srinivas2022hypergraph} \\ \cite{zhou2022hybrid,zhou2022hybrid,chen2023multivariate}
        \end{tabular}\\
        
        \multicolumn{1}{l|}{} & SWaT \cite{mathur2016swat}& 925,010     & 51 & 1 second &  11.97\% &  Industrial Systems & \begin{tabular}[c]{@{}l@{}} \cite{deng2021graph,chen2021learning,dai2022graphaugmented,zhang2022grelen,han2022learning} \\ \cite{srinivas2022hypergraph,zhou2022hybrid,chen2023multivariate,zheng2023correlation,zheng2024graph}
        \end{tabular}\\

        \multicolumn{1}{l|}{} & WADI \cite{ahmed2017wadi}& 1.4 million    & 127 & 1 second &  5.99\% &  Industrial Systems & \begin{tabular}[c]{@{}l@{}} \cite{deng2021graph,chen2021learning,zhang2022grelen,han2022learning}, \\ \cite{srinivas2022hypergraph,chen2023multivariate,zheng2023correlation,zheng2024graph}
        \end{tabular}\\
        
        \hline
  
		\multicolumn{1}{l|}{\multirow{2}{*}{\begin{tabular}[c]{@{}l@{}} Classification \end{tabular}}} & {\begin{tabular}[c]{@{}l@{}} UCR Archive\cite{dau2019ucr}  \end{tabular}}  & 40-24,000 & 1 & Varies  & - & Wide Applications  &        \begin{tabular}[c]{@{}l@{}}\cite{duan2022multivariate,zha2022towards,xi2023lb}\\
		\end{tabular}\\ 
		
		\multicolumn{1}{l|}{} & {\begin{tabular}[c]{@{}l@{}} UEA Repository \cite{bagnall2018uea}\end{tabular}}  & 27-50,000 & Varies & 2-1345  & - & Wide Applications &        \begin{tabular}[c]{@{}l@{}}\cite{liu2023todynet,younis2024mts2graph}\\
		\end{tabular}\\ \hline

  \toprule
	\end{tabular}
 \end{adjustbox}
\end{table*}

\noindent\textbf{Fraud Detection.} As elucidated in the four-element fraud diamond \cite{wolfe2004fraud}, the perpetration of fraud necessitates not just the presence of incentive and rationalization, but also a significant degree of capability - often attainable only via coordinated group efforts undertaken at opportune moments. This suggests that fraud is typically committed by entities possessing sufficient capability, which can be primarily achieved through collective endeavors during suitable periods. Consequently, it is rare for fraudsters to operate in isolation \cite{li2021happens,chen2022medical}. They also frequently demonstrate unusual temporal patterns in their activities, further supporting the necessity for sophisticated fraud detection measures \cite{noorshams2020ties,zhao2020early}. To this end, GNNs have been proposed to capture these complex relational and temporal dynamics inherent to fraud network activities. They have found successful applications in various domains, such as detecting frauds and anomalies in social networks \cite{noorshams2020ties, zhao2020early,huang2021recurrent}, financial networks and systems \cite{cheng2020graphfraud,li2021temporalfraud,wang2021temporal,reddy2021tegraf}, and in several other sectors \cite{chen2022medical,chu2022exploiting,jin2022towards, lu2022bright, xu2022multi,koh2022empirical}. \\

\noindent\textbf{Other Applications.} Beyond the aforementioned sectors, the application of GNNs for time series analysis has also been extended to various other fields, such as finance~\cite{wang2022review}, urban planning~\cite{xiao2021hybrid, hou2022urban}, epidemic control~\cite{kapoor2020examining, yu2023spatio, geng2022analysis, liu2023human}, recommender systems~\cite{wu2019session, su2023enhancing, ye2023sincere}, and \revision{manufacturing}~\cite{yao2023novel}. As research in this area continues to evolve, it is anticipated that the application of GNN4TS will continue to expand, opening up new possibilities for data-driven decision making and system optimization.

\revision{\section{Datasets and Implementations}\label{appx:resources}}

\begin{table*}[htbp]
	\caption{\revision{A summary of open-source implementation of representative approaches.}}
	\label{tab:implementation}
	\centering
	\begin{adjustbox}{width=2.0\columnwidth,center}
	\begin{tabular}{ l l l l}
		\toprule
		\textbf{Approach} & \textbf{Year} & \textbf{Task Category} & \textbf{Github Repository} \\ \bottomrule
		DCRNN~\cite{lidiffusion} & 2018 & Forecasting & \url{https://github.com/liyaguang/DCRNN} \\
        STGCN~\cite{yu2018spatio} & 2018 & Forecasting & \url{https://github.com/VeritasYin/STGCN_IJCAI-18} \\
        ST-MetaNet~\cite{pan2019urban} & 2019 & Forecasting & \url{https://github.com/panzheyi/ST-MetaNet} \\
        ASTGCN~\cite{guo2019attention} & 2019 & Forecasting & \url{https://github.com/guoshnBJTU/ASTGCN-2019-pytorch} \\
        Graph WaveNet~\cite{wu2019graph} & 2019 & Forecasting & \url{https://github.com/nnzhan/Graph-WaveNet} \\
        MRA-BGCN~\cite{chen2020multi} & 2020 & Forecasting & \url{https://github.com/wumingyao/MAR-BGCN_GPU_pytorch} \\
        MTGNN~\cite{wu2020connecting} & 2020 & Forecasting & \url{https://github.com/nnzhan/MTGNN} \\
        GMAN~\cite{zheng2020gman} & 2020 & Forecasting & \url{https://github.com/zhengchuanpan/GMAN} \\
        StemGNN~\cite{cao2020spectral} & 2020 & Forecasting & \url{https://github.com/microsoft/StemGNN} \\
        GTS~\cite{shang2021discrete} & 2021 & Forecasting & \url{https://github.com/chaoshangcs/GTS} \\
        Z-GCNETs~\cite{chen2021z} & 2021 & Forecasting & \url{https://github.com/Z-GCNETs/Z-GCNETs} \\
        STGODE~\cite{fang2021spatial} & 2021 & Forecasting & \url{https://github.com/square-coder/STGODE} \\
        STFGNN~\cite{li2021spatial} & 2021 & Forecasting & \url{https://github.com/MengzhangLI/STFGNN} \\
        TPGNN~\cite{liu2022multivariate} & 2022 & Forecasting & \url{https://github.com/zyplanet/TPGNN} \\
        MTGODE~\cite{jin2022multivariate} & 2022 & Forecasting & \url{https://github.com/TrustAGI-Lab/MTGODE} \\
        STG-NCDE~\cite{choi2022graph} & 2022 & Forecasting & \url{https://github.com/jeongwhanchoi/STG-NCDE} \\
        STEP~\cite{shao2022pre} & 2022 & Forecasting & \url{https://github.com/zezhishao/STEP} \\
        CaST~\cite{xia2023deciphering} & 2023 & Forecasting & \url{https://github.com/yutong-xia/CaST} \\
        GPT-ST~\cite{li2024gpt} & 2023 & Forecasting & \url{https://github.com/HKUDS/GPT-ST} \\ \hline
        MTAD-GAT \cite{zhao2020multivariate} & 2020 & Anomaly Detection & \url{https://github.com/mangushev/mtad-gat} \\
        GDN\cite{deng2021graph} & 2021 & Anomaly Detection & \url{https://github.com/d-ailin/GDN} \\
        GANF\cite{dai2022graphaugmented} & 2022 & Anomaly Detection & \url{https://github.com/EnyanDai/GANF} \\
        VGCRN\cite{chen2022deep} & 2022 & Anomaly Detection & \url{https://github.com/BoChenGroup/DVGCRN} \\
        STGAN\cite{zhou2022hybrid} & 2022 & Anomaly Detection & \url{https://github.com/dleyan/STGAN} \\
        CST-GL\cite{zheng2023correlation} & 2023 & Anomaly Detection & \url{https://github.com/huankoh/CST-GL} \\ \hline
        Time2Graph+\cite{cheng2021time2graph+} & 2021 & Classification & \url{https://github.com/petecheng/Time2GraphPlus} \\
        RainDrop\cite{zhang2022graphguided} & 2022 & Classification & \url{https://github.com/mims-harvard/Raindrop} \\
        SimTSC\cite{zha2022towards} & 2022 & Classification & \url{https://github.com/daochenzha/SimTSC} \\
        TodyNet\cite{liu2023todynet} & 2023 & Classification & \url{https://github.com/liuxz1011/TodyNet} \\
        MTS2Graph\cite{younis2024mts2graph} & 2024 & Classification & \url{https://github.com/raneeny/MTS2Graph} \\ \hline
        IGNNK~\cite{wu2021inductive} & 2021 & Imputation & \url{https://github.com/Kaimaoge/IGNNK} \\
        GRIN~\cite{cini2022filling} & 2022 & Imputation & \url{https://github.com/Graph-Machine-Learning-Group/grin} \\
        SPIN~\cite{mariscalearning} & 2022 & Imputation & \url{https://github.com/Graph-Machine-Learning-Group/spin} \\
        PriSTI~\cite{liu2023pristi} & 2023 & Imputation & \url{https://github.com/LMZZML/PriSTI} \\
        INCREASE~\cite{zheng2023increase} & 2023 & Imputation & \url{https://github.com/zhengchuanpan/INCREASE} \\
 	  \toprule
	\end{tabular}
	\end{adjustbox}
\end{table*}

\noindent \revision{We summarize the common benchmark datasets and the open-sourced implementation of representative models, which are listed in \shortautoref{tab:dataset} and \shortautoref{tab:implementation}.} \\

\noindent\revision{\textbf{Forecasting and Imputation Benchmarks.} The \emph{METR-LA}~\cite{lidiffusion} dataset is a prominent benchmark for traffic modeling and forecasting. It comprises data collected from 207 sensors on highways in Los Angeles, recording speed in miles per hour from March 1st to June 30th, 2012.}
\revision{Similarly, the \emph{PEMS-BAY}~\cite{lidiffusion} dataset is another widely utilized benchmark for traffic modeling and forecasting. Collected by the California State Transportation Agency (CalSTA), it includes data from 325 sensors in the Bay Area, recording speeds in miles per hour from January 1st to May 31st, 2017.} 

\revision{PeMS~\cite{chen2001freeway} datasets such as \emph{PeMSD3}, \emph{PeMSD4}, \emph{PeMSD7}, and \emph{PeMSD8} are derived from the Caltrans Performance Measurement System (PeMS). These datasets are collected in real-time every 30 seconds and aggregated into 5-minute intervals from the raw data. They typically contain various measurements, such as vehicle count and traffic velocity; however, the mentioned four PeMS datasets focus specifically on traffic volume.}

\revision{The \emph{Xiamen}~\cite{zheng2020gman} dataset includes five months of data from 95 traffic sensors in Xiamen, China, spanning from August 1st, 2015, to December 31st, 2015, with a sampling rate of five minutes.}
\revision{The \emph{Beijing}~\cite{zheng2015forecasting} dataset contains pollutant concentration data (PM2.5) from 36 sensors in Beijing, recorded over one year from May 1st, 2014, to April 30th, 2015.}

\revision{The \emph{NYC-Taxi}~\cite{yao2019revisiting} dataset features taxi trip records from New York City (NYC) for the period from January 1st to March 1st, 2015. The \emph{NYC-Bike}~\cite{yao2019revisiting} dataset includes bike trajectory data collected from the NYC Citi Bike system between July 1st and August 29th, 2016. The \emph{NYC-Crime}~\cite{li2022spatial} dataset records various types of crime occurrences in NYC, such as robbery and larceny, using a 1 km $\times$ 1 km spatial grid to generate an adjacency matrix.}

\revision{The \emph{AQI}~\cite{zheng2015forecasting} dataset consists of recordings of several air quality indices from 437 monitoring stations across 43 Chinese cities, with a focus on the PM2.5 pollutant. The \emph{AQI-36} is a reduced version of this dataset, containing data from only 36 sensors.}

\noindent\revision{\textbf{Anomaly Detection Benchmarks.}
For time-series anomaly detection, significant attention has been directed towards a select few real-world datasets, including SMD\cite{su2019robust}, MSL\cite{hundman2018detecting}, SMAP\cite{hundman2018detecting}, SWAT\cite{mathur2016swat} and WADI\cite{ahmed2017wadi}. These datasets have been pivotal in driving research and development within the field. However, there have been critical investigations into their limitations \cite{wu2021current,garg2021evaluation,schmidl2022anomaly}, particularly regarding how well they represent real-world scenarios\cite{audibert2022deep}. These studies highlight the need for more diverse and representative datasets to ensure that anomaly detection models can perform robustly in practical applications.}

\revision{To address these limitations, many models are also evaluated using internal datasets from proprietary investigations, providing a broader perspective on their performance. Furthermore, there have been concerted efforts to propose and develop better datasets that more comprehensively cover various types of anomaly scenarios\cite{li2021multivariate,nedelkoski2020multi}. These enhanced datasets aim to capture a wider range of anomalies and identify root-cause scenarios more effectively, thereby improving the generalizability and applicability of anomaly detection models\cite{jacob2020exathlon}.}

\revision{The push towards creating more representative datasets and refining evaluation metrics underscores the ongoing efforts to enhance the reliability and real-world relevance of time-series anomaly detection. However, this also requires researchers to thoroughly benchmark their models against baseline models on these datasets in a fair and comparative manner. Comprehensive research studies that benchmark existing models are crucial for building a solid foundation for future advancements. By addressing the gaps in current benchmark datasets and ensuring models are tested against realistic scenarios, the field can develop more robust and effective solutions for detecting anomalies in diverse environments.}\\

\noindent\revision{\textbf{Classification Benchmarks.}
Time-series classification benchmarks, particularly in the realm of univariate and multivariate datasets, have evolved and progressed substantially. The univariate time series classification benchmarks, exemplified by the UCR archive\cite{dau2019ucr}, and the multivariate time series benchmarks, represented by the UEA repository\cite{bagnall2018uea}, demonstrate considerable progress compared to other fields. This progress is attributed to dedicated efforts in carefully collecting datasets from a wide array of domains and applications, ensuring a diverse and representative sample of real-world scenarios.} \\

\revision{These standardized benchmarks have further been extensively studied through comprehensive investigations, often referred to as Bakeoff studies\cite{bagnall2017great,ruiz2021great,middlehurst2024bake}. Such studies provide a systematic framework to evaluate a wide range of methods, enabling robust comparisons and fostering methodological advancements. By rigorously assessing various approaches against a consistent set of benchmarks, researchers can identify the strengths and limitations of different models, facilitating continuous improvements in the field.}

\revision{Moreover, there have been concerted efforts to refine the metrics used for model comparison, enhancing the accuracy and relevance of performance evaluations\cite{ismail2023approach}. These improved metrics make the benchmark datasets more indicative of real-world performances, ensuring that the models developed and tested under these conditions are better aligned with practical applications. This focus on realistic and meaningful evaluation criteria underscores the critical role of these benchmarks in advancing the state-of-the-art in time-series classification.} \\

\section{Future Directions}\label{appx:prospect}
\noindent\textbf{Pre-training, Transfer Learning, and Large Models.}
Pre-training, transfer learning, and large models are emerging as potent strategies to bolster the performance of GNNs in time series analysis~\cite{shao2022pre, mallick2021transfer, wang2023building}, especially when data is sparse or diverse. These techniques hinge on utilizing learned representations from one or more domains to enhance performance in other related domains~\cite{zhang2023selfsupervised, pan2023unifying}. Recent successful examples are Panagopoulos et al.'s~\cite{panagopoulos2021transfer} model-agnostic meta-learning schema for COVID-19 spread prediction in data-limited cities, and Shao et al.'s~\cite{shao2022pre} pre-training enhanced framework for spatial-temporal GNNs. The exploration of pre-training strategies and GNN transferability for time series tasks is a burgeoning research area, especially in the current era of generative AI and large models, which showcase the potential for a single, multimodal model to address diverse tasks~\cite{jin2023large}. However, several challenges remain, including the limited availability of time series data for large-scale pre-training compared to language data for large language models (LLMs)~\cite{zhao2023survey}, to ensure wide coverage and transferability of learned knowledge, and designing effective pre-training strategies that capture complex spatial-temporal dependencies. Addressing these challenges is pivotal for the future development and application of GNN4TS. \\

\noindent\textbf{Robustness.} Robustness of GNNs refers to their ability to handle various forms of data perturbations and distribution shifts, particularly those that are deliberately engineered by adversaries \cite{zhang2022trustworthy}. This quality becomes critical when dealing with time series data generated by rapidly evolving systems. Any operational failures within GNNs can potentially precipitate adverse consequences on the integrity of the entire system \cite{moore2020iot,dong2023graph}. For instance, if a GNN fails to adequately handle noise or data corruption in a smart city application, it might disrupt essential traffic management functions. Similarly, in healthcare applications, the inability of a GNN to remain robust amidst disturbances could lead to healthcare providers missing out on critical treatment periods, potentially having serious health implications for patients. While GNNs have demonstrated superior performance across numerous applications, improving their robustness and creating effective failure management strategies remains vital. This not only enhances their reliability but also widens their potential usage across contexts. \\

\noindent\textbf{Interpretability.} The interpretability of GNNs plays an equally pivotal role in facilitating the transparent and accountable use of these sophisticated tools. This attribute sheds light on the opaque decision-making processes of GNNs, allowing users to comprehend the reasoning behind a given output or prediction. Such understanding fosters trust in the system and enables the discovery of latent patterns within the data \cite{luo2020parameterized,zhang2022trustworthy}. For example, in drug discovery~\cite{koh2023psichic,nguyen2023gpcr} and financial time series analyses~\cite{abrate2021counterfactual,yang2021financial}, interpretability may illuminate causal factors, facilitating more informed decision-making. As we strive to harness the full potential of GNN4TS, advancing their interpretability is paramount to ensuring their ethical and judicious application in increasingly complex environments. \\

\noindent\textbf{Uncertainty Quantification.} Time series data, by its nature, is often fraught with unpredictable noise and uncertainty associated with the data-generating process. The ability of a model to account for and quantify uncertainty can greatly enhance its reliability and utility\cite{richardson2012uncertainty,hullermeier2021aleatoric,koh2022far}. Uncertainty quantification provides a probabilistic measure of the confidence to the predictions made by the model and to the system state estimates \cite{zambon2023graph, alippi2023graph}, aiding in the understanding of the range and likelihood of potential outcomes\cite{abdar2021review}. This becomes particularly important when GNNs are used for decision-making processes in fields where high stakes are involved, such as financial forecasting, healthcare monitoring\cite{li2020graphhealth,li2020knowledge}, or traffic prediction in smart cities\cite{pan2019urban,guo2019attention,geng2019spatiotemporal,wen2022graph2route}. Despite progress, a gap remains in the current GNN models, which largely provide point estimates\cite{pan2019urban,guo2019attention,wu2020connecting,chen2021learning,deng2021graph,zhang2022graphguided,zha2022towards}, inadequately addressing the potential uncertainties. This underlines an essential research direction: developing sophisticated uncertainty quantification methods for GNNs to better navigate the complexities of time series data. This endeavor not only enhances the interpretability and reliability of predictions but also fosters the development of advanced models capable of learning from uncertainty. Thus, uncertainty quantification represents a pivotal element in the ongoing advancement of GNN4TS. \\

\noindent\textbf{Privacy Enhancing.}
GNNs have established themselves as invaluable tools in time series analysis, playing crucial roles in diverse, interconnected systems across various sectors \cite{yao2018deep,li2020graphhealth,noorshams2020ties,cai2021deep,wang2021temporal,mo2023predictive}. As these models gain broader adoption, particularly in fields that require the powerful data forecasting\cite{deng2021graph,tang2022selfsupervised} and reconstruction \cite{zhao2020multivariate,cini2022filling} capabilities of GNNs, the need for stringent privacy protection becomes increasingly apparent. Given the ability of GNNs to learn and reconstruct the relationships between entities within complex systems \cite{guo2021hierarchical,chen2023DyGraphAD}, it is essential to safeguard not only the privacy of individual entities (nodes), but also their relationship (edges) within the time series data \cite{zheleva2008preserving,zhang2022trustworthy}. Furthermore, the interpretability of GNNs can serve as a double-edged sword. While it can help identify and mitigate areas vulnerable to malicious attacks, it could also expose the system to new risks by revealing sensitive information \cite{xu2021explainability}. Therefore, maintaining robust privacy defenses while capitalizing on the benefits of GNN models for time series analysis requires a delicate balance, one that calls for constant vigilance and continual innovation. \\ 

\noindent\textbf{Scalability.} GNNs have emerged as powerful tools to model and analyze increasingly large and complex time series data, such as large social networks consisting of billions of users and relationships \cite{noorshams2020ties,huang2021recurrent}. However, the adaptation of GNNs to manage vast volumes of time-dependent data introduces unique challenges. Traditional GNN models often require the computation of the entire adjacency matrix and node embeddings for the graph, which can be extraordinarily memory-intensive \cite{wu2022graph}. To counter these challenges, traditional GNN methods utilize sampling strategies like node-wise \cite{hamilton2017inductive,chen2018stochastic}, layer-wise \cite{chen2018fastgcn}, and graph-wise\cite{chiang2019cluster} sampling. Yet, incorporating these methods while preserving consideration for temporal dependencies remains a complex task. Additionally, improving scalability during the inference phase to allow real-time application of GNNs for time series analysis is vital, especially in edge devices with limited computational resources. This intersection of scalability, time-series analysis, and real-time inference presents a compelling research frontier, ripe with opportunities for breakthroughs. Hence, exploring these areas can be a pivotal GNN4TS research direction. \\

\noindent\textbf{AutoML and Automation.}  Despite the notable success of GNNs in temporal analytics \cite{wu2020connecting,zhao2020multivariate,cini2022filling,zhang2022graphguided}, their empirical implementations often necessitate meticulous architecture engineering and hyperparameter tuning to accommodate varying types of graph-structured data \cite{guan2021autogl, zheng2023auto}.  A GNN architecture is typically instantiated from its model space and evaluated in each graph analysis task based on prior knowledge and iterative tuning processes \cite{wu2022graph}. Furthermore, with the plethora of architectures being proposed for different use cases \cite{cini2022filling,jiang2022graph,foumani2023deep,zha2022towards,ho2023graph}, discerning the most suitable option poses a significant challenge for end users. 

AutoML and automation in time series analysis using GNNs thus plays a pivotal role in overcoming the complexities associated with diverse model architectures. It can simplify the selection process, enhancing efficiency and scalability while fostering effective model optimization \cite{rapaport2019eegnas,li2020autost,guan2021autogl}. Furthermore, it is important to note that GNNs may not always be the optimal choice compared to other methods\cite{bagnall2017great,alsharef2022review,wang2022forecast}. Their role within the broader landscape of AutoML must therefore be thoughtfully evaluated. By encouraging reproducibility and broadening accessibility, automation democratizes the benefits of GNNs for advanced temporal analytics.

\bibliographystyle{IEEEtran}
\bibliography{IEEEabrv,reference}

\vspace{-1.1cm}
\begin{IEEEbiography}
[{\includegraphics[width=1in,height=1.25in,clip,keepaspectratio]{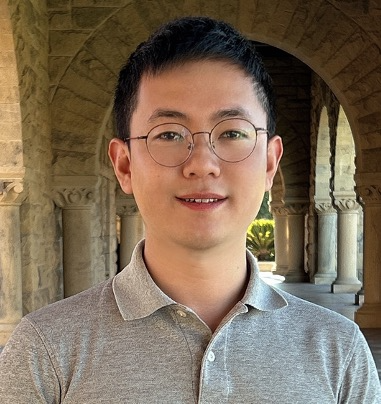}}]
{Ming Jin} is currently an Assistant Professor at the School of Information and Communication Technology, Griffith University. He obtained his Ph.D. degree from Monash University, Australia, in 2024. He has published over twenty peer-reviewed papers in top-ranked journals and conferences, including TPAMI, TKDE, NeurIPS, ICLR, ICML, etc. He serves as an Associate Editor for Journal of Neurocomputing and regularly contributes as Area Chair/PC member for major AI conferences. His research interests include time series analysis, graph neural networks, and multi-modal learning.
\vspace{-1cm}
\end{IEEEbiography}

\begin{IEEEbiography}[{\includegraphics[width=1in,height=1.25in,clip,keepaspectratio]{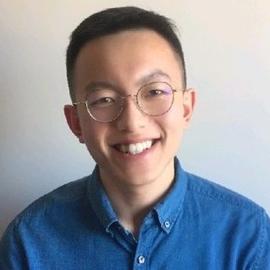}}]
{Huan Yee Koh} received the Master of Data Science and Bachelors of Commerce from Monash University, Melbourne, Australia, in 2021 and 2018, respectively. He is currently pursuing his Ph.D. degree in machine learning at Monash University, Melbourne, Australia. His research focuses on graph neural networks, time series analysis, drug discovery, data mining, machine learning. 
\end{IEEEbiography}

\begin{IEEEbiography}
[{\includegraphics[width=1in,height=1.25in,clip,keepaspectratio]{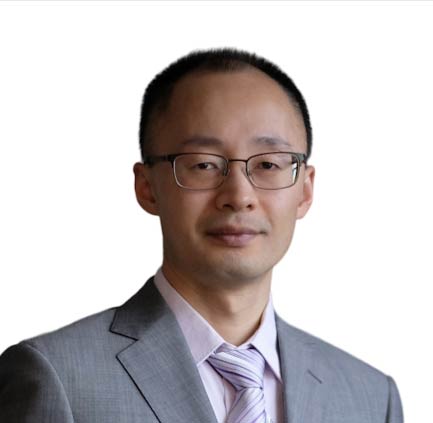}}]
{Qingsong Wen} is the Head of AI Research \& Chief Scientist at Squirrel Ai Learning. He holds a Ph.D. in Electrical and Computer Engineering from Georgia Institute of Technology. He has published over 100 top-ranked AI conference and journal papers, had multiple Oral/Spotlight Papers at NeurIPS, ICML, and ICLR, had multiple Most Influential Papers at IJCAI, received multiple IAAI Deployed Application Awards at AAAI, and won First Place of SP Grand Challenge at ICASSP. He organizes workshops on AI for Time Series and AI for Education and serves as an Associate Editor for Neurocomputing and IEEE Signal Processing Letters, and Guest Editor for Applied Energy and IEEE Internet of Things Journal. His research focuses on AI for time series, AI for education, and general machine learning.
\end{IEEEbiography}

\begin{IEEEbiography}
[{\includegraphics[width=1in,height=1.25in,clip,keepaspectratio]{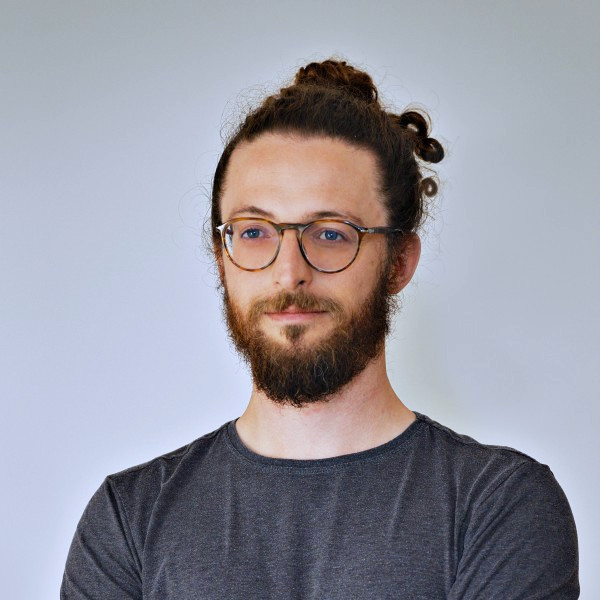}}]
{Daniele Zambon} is a postdoctoral researcher at the Swiss AI Lab IDSIA, Università della Svizzera italiana (Switzerland). He received his Ph.D.\ degree from Università della Svizzera italiana. 
He has been a visiting researcher/intern at the University of Florida (US), the University of Exeter (UK), and STMicroelectronics (Italy). 
He is a member of the IEEE CIS Task Force on Learning for Graphs and has co-organized special sessions and tutorials on deep learning and graph data. 
His main research interests include graph representation learning, time series analysis, and learning in non-stationary environments.
\end{IEEEbiography}

\begin{IEEEbiography}
[{\includegraphics[width=1in,height=1.25in,clip,keepaspectratio]{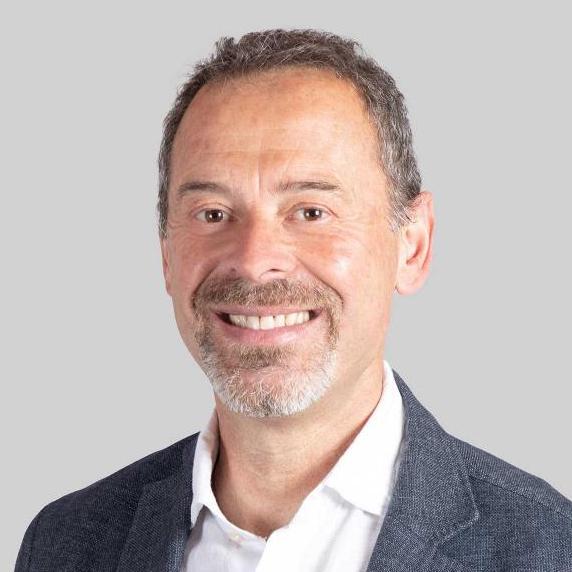}}]
{Cesare Alippi} (Fellow, IEEE) is a Professor at Università della Svizzera italiana and Politecnico di Milano, a Visiting Professor at the University of Guangzhou, and an Advisory Professor at Northwestern Polytechnic, Xi’an. He has authored/coauthored one monograph, seven edited books, and around 200 papers, and holds eight patents. His research focuses on adaptation and learning in non-stationary environments, graph learning, intelligence for embedded systems, IoT, and cyber-physical systems. He is a fellow of the European Laboratory for Learning and Intelligent Systems, serves on the IEEE Computational Intelligence Society (CIS) Administrative Committee, and is a Board of Governors Member of the International Neural Network Society. He has held various roles within IEEE CIS, including vice-president for education and awards committee chair, and has received several awards, including the International Neural Networks Society Gabor Award.
\end{IEEEbiography}

\begin{IEEEbiography}[{\includegraphics[width=1in,height=1.25in,clip,keepaspectratio]{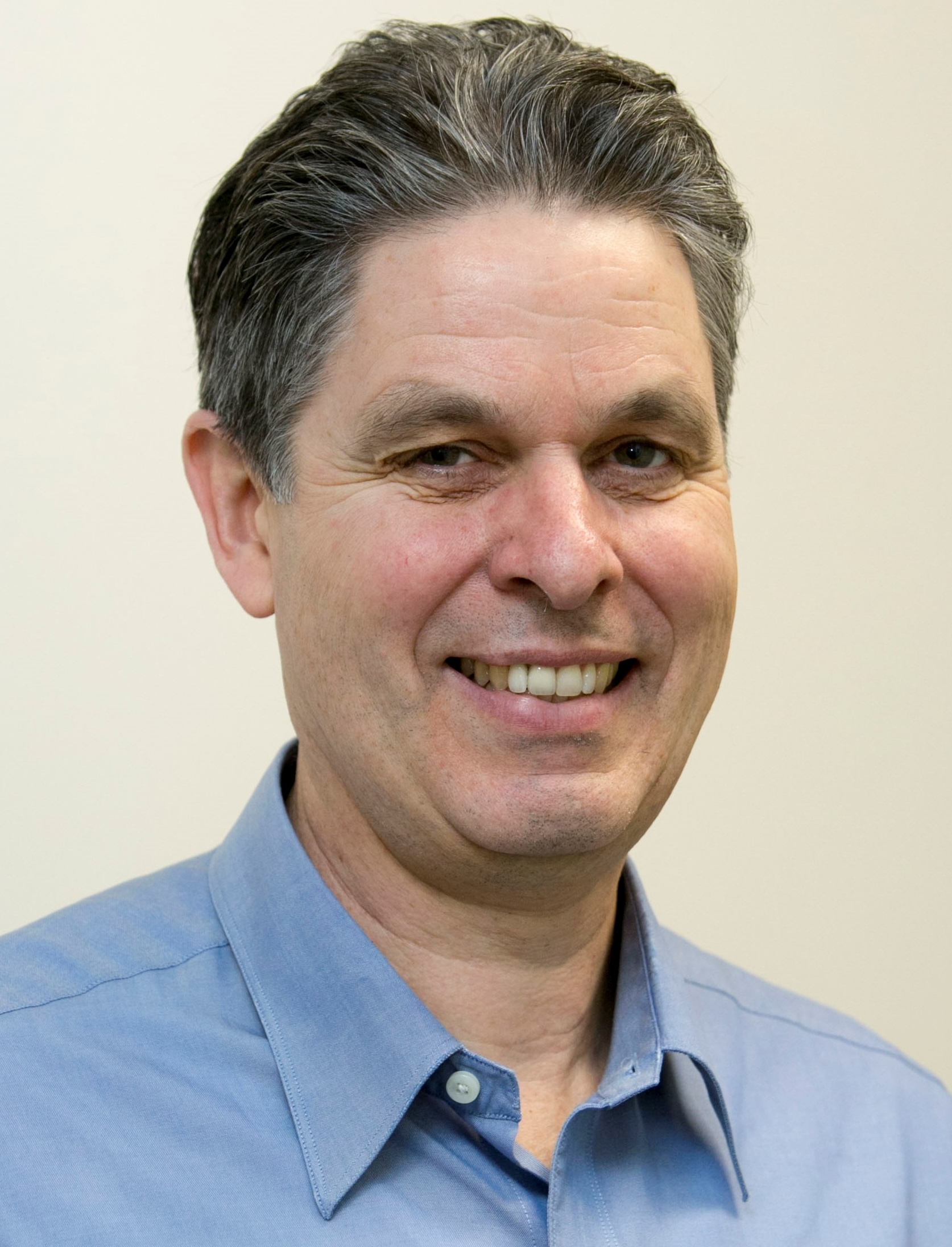}}]{Geoffrey I. Webb} (Fellow, IEEE) is a professor in the Monash University Department of Data Science and AI, Australia. He was editor in chief of the
leading Data Mining Journal, Data Mining and
Knowledge Discovery, from 2005 to 2014. He
has been program committee chair of both the
leading data mining conferences, ACM SIGKDD
and IEEE ICDM, as well as general chair of
ICDM. He is a technical advisor to the startups
BigML Inc and FROOMLE. He developed many
of the key mechanisms of support-confidence
association discovery in the 1980s. He pioneered multiple research
areas as diverse as black-box user modelling, interactive data analytics
and statistically-sound pattern discovery. His many awards include IEEE
Fellow, IEEE ICDM Ten-Year Impact Award (2023) and the inaugural Eureka Prize for Excellence in Data Science.
\end{IEEEbiography}

\begin{IEEEbiography}[{\includegraphics[width=1in,height=1.25in,clip,keepaspectratio]{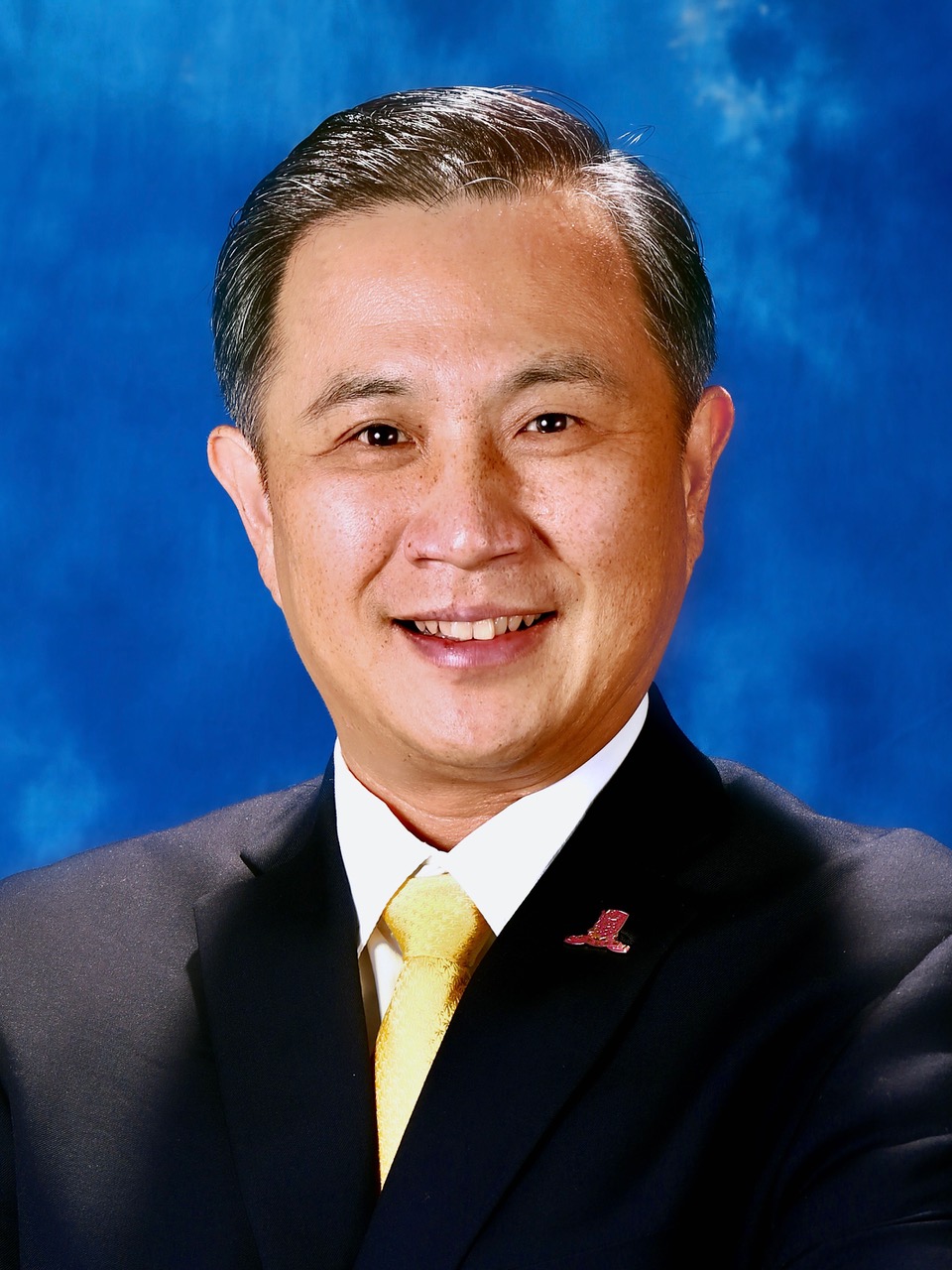}}]
{Irwin King} (Fellow, IEEE) received his B.S. and Ph.D. in computer science from Caltech and the University of Southern California, respectively. He is professor in the Department of Computer Science \& Engineering at The Chinese University of Hong Kong. His research interests include machine learning, social computing, AI, web intelligence, data mining, and multimedia information processing, with over 300 technical publications in these areas. He is an Associate Editor of the Journal of Neural Networks, an ACM distinguished member, and a fellow of the International Neural Network Society (INNS), and Hong Kong Institute of Engineers (HKIE). He has served as President of the International Neural Network Society (INNS) and as general co-chair of several major conferences. Currently he is the Vice-chair of ACM SIGWEB and WebConf Steering Committee.  He has received the ACM CIKM 2019 Test of Time Award, the ACM SIGIR 2020 Test of Time Award, and the 2020 APNNS Outstanding Achievement Award for contributions to social computing with machine learning.
\vspace{-11cm}
\end{IEEEbiography}

\begin{IEEEbiography}[{\includegraphics[width=1in,height=1.25in,clip,keepaspectratio]{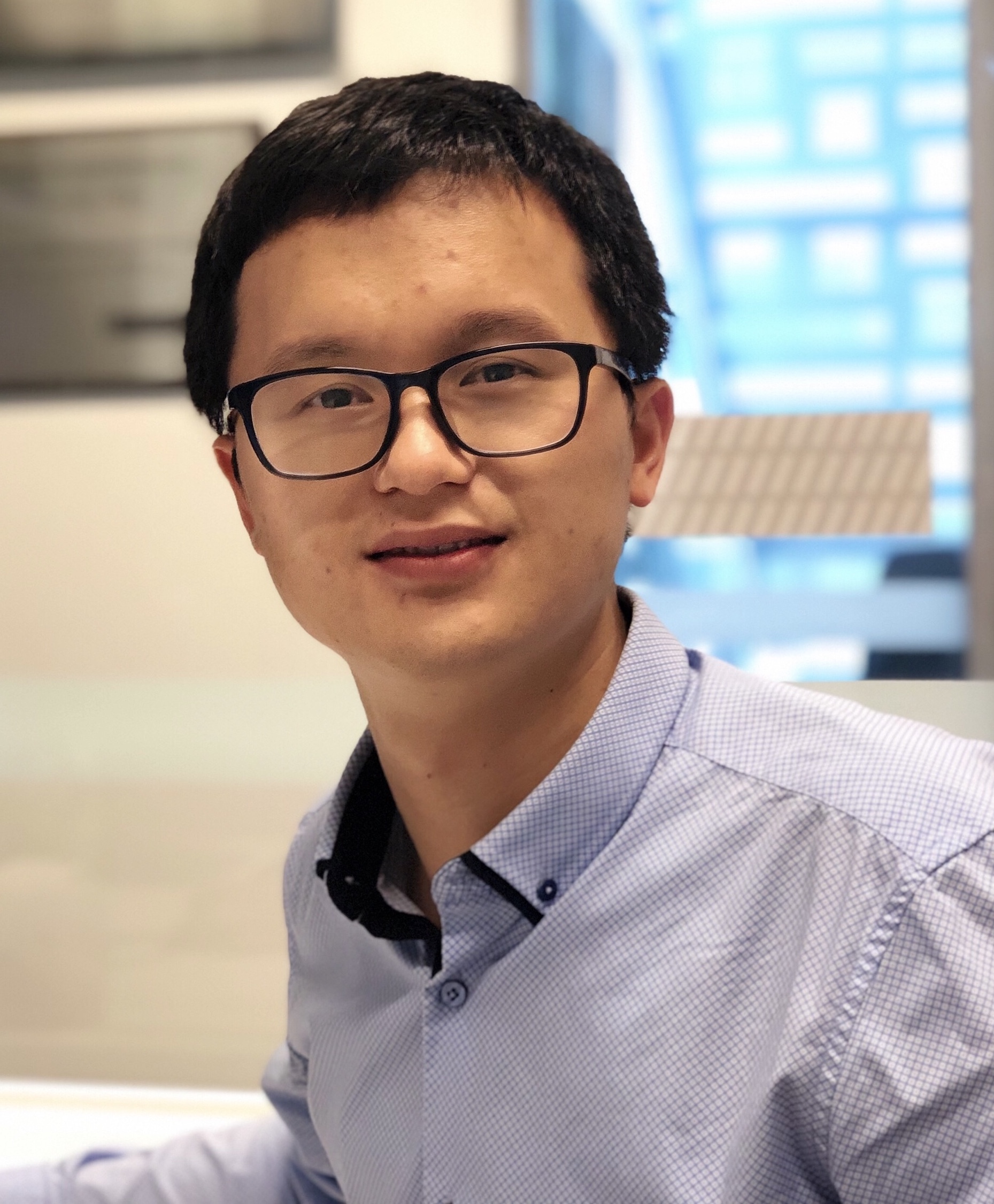}}]{Shirui Pan} received a Ph.D. in computer science from the University of Technology Sydney (UTS), Ultimo, NSW, Australia. He is a Professor with the School of Information and Communication Technology, Griffith University, Australia.
Prior to this, he was a Senior Lecturer with the Faculty of IT at Monash University. 
His research interests include data mining and machine learning. To date, Dr Pan has published over 100 research papers in top-tier journals and conferences, including TPAMI, TKDE, TNNLS, ICML, NeurIPS, and KDD.  His research received the 2024 CIS IEEE TNNLS Oustanding Paper Award and the 2020 IEEE ICDM Best Student Paper Award. He is recognized as one of the AI 2000 AAAI/IJCAI Most Influential Scholars in Australia. He is an ARC Future Fellow and a Fellow of Queensland Academy of Arts and Sciences (FQA).
\end{IEEEbiography}

\balance

\end{document}